\RequirePackage{fix-cm}
\RequirePackage{lineno}
\documentclass[twocolumn]{svjour3}          
\usepackage{graphicx}
\usepackage{epstopdf}
\usepackage{amsbsy}
\usepackage{amstext}
\usepackage{mathrsfs} 
\usepackage [round, authoryear]{natbib}    
\usepackage{float}
\usepackage{units}
\usepackage{esint}
\usepackage{subfig}
\usepackage{color}
\usepackage{multirow}

\journalname{International Journal of Computer Vision}
\begin{document}

\title{Rotational Projection Statistics for 3D Local Surface Description and Object Recognition
\thanks{This research is supported by a China Scholarship Council (CSC) scholarship and Australian Research Council grants (DE120102960, DP110102166). }
}


\author{Yulan Guo         \and
        Ferdous Sohel \and
        Mohammed Bennamoun \and
        Min Lu \and
        Jianwei Wan
}


\institute{Y. Guo \at
              College of Electronic Science and Engineering,\\
              National University of Defense Technology, \\
              Changsha, Hunan, P.R.China \\
              \email{yulan.guo@nudt.edu.cn}
           \and
           Y. Guo \and F. Sohel \and M. Bennamoun \at
           School of Computer Science and Software Engineering,\\
           The University of Western Australia, \\
           Perth, Australia
           \and
           M. Lu \and J. Wan \at
           College of Electronic Science and Engineering,\\
           National University of Defense Technology
}

\date{Received: date / Accepted: date}

\maketitle
\begin{abstract}
Recognizing 3D objects in the presence of noise, varying mesh resolution,
occlusion and clutter is a very challenging task. This paper presents
a novel method named Rotational Projection Statistics (RoPS). It has
three major modules: Local Reference Frame (LRF) definition, RoPS
feature description and 3D object recognition. We propose a novel
technique to define the LRF by calculating the scatter matrix of all
points lying on the local surface. RoPS feature descriptors are obtained
by rotationally projecting the neighboring points of a feature point
onto 2D planes and calculating a set of statistics (including low-order
central moments and entropy) of the distribution of these projected
points. Using the proposed LRF and RoPS descriptor, we present a hierarchical
3D object recognition algorithm. The performance of the proposed LRF,
RoPS descriptor and object recognition algorithm was rigorously tested
on a number of popular and publicly available datasets. Our proposed
techniques exhibited superior performance compared to existing techniques.
We also showed that our method is robust with respect to noise and
varying mesh resolution. {Our RoPS based algorithm
achieved recognition rates of 100\%, 98.9\%, 95.4\% and 96.0\% respectively
when tested on the Bologna, UWA, Queen's and Ca' Foscari Venezia Datasets.}

\end{abstract}

\section{Introduction}

Object recognition is an active research area in computer vision with
numerous applications including navigation, surveillance, automation,
biometrics, surgery and education \citep{guo2013part,Johnson1999,lei2012efficient,tombari2010unique}.
The aim of object recognition is to correctly identify the objects
that are present in a scene and recover their poses (i.e., position
and orientation) \citep{Mian2006}. Beyond object recognition from
2D images {\citep{brown2003recognising,lowe2004distinctive,mikolajczyk2004scale}},
3D object recognition has been extensively investigated during the
last two decades due to the availability of low cost scanners and
high speed computing devices \citep{mamic2002representation}. However,
recognizing objects from range images in the presence of noise, varying
mesh resolution, occlusion and clutter is still a challenging task.

Existing algorithms for 3D object recognition can broadly be classified
into two categories, i.e., global feature based and local feature
based algorithms \citep{bayramoglu2010shape,castellani2008sparse}.
The global feature based algorithms construct a set of features which
encode the geometric properties of the entire 3D object. Examples
of these algorithms include the geometric 3D moments \citep{paquet2000description},
shape distribution \citep{osada2002shape} and spherical harmonics
\citep{funkhouser2003search}. However, these algorithms require complete
3D models and are therefore sensitive to occlusion and clutter \citep{bayramoglu2010shape}.
In contrast, the local feature based algorithms define a set of features
which encode the characteristics of the local neighborhood of feature
points. The local feature based algorithms are robust to occlusion
and clutter. They are therefore even suitable to recognize partially
visible objects in a cluttered scene \citep{petrelli2011repeatability}.

A number of local feature based 3D object recognition algorithms have
been proposed in the literature, including point signature based \citep{chua1997point},
spin image based \citep{Johnson1999}, tensor based \citep{Mian2006}
and Exponential Map (EM) based \citep{bariya20123d} algorithms. Most
of these algorithms follow a paradigm that has three phases, i.e.,
feature matching, hypothesis generation and verification, and pose
refinement \citep{taati2011local}. Among these phases, feature matching
plays a critical role since it directly affects the effectiveness
and efficiency of the  two subsequent phases \citep{taati2011local}.

Descriptiveness and robustness of a feature descriptor are crucial
for accurate feature matching \citep{bariya2010scale}. The feature
descriptors should be highly descriptive to ensure an accurate and
efficient object recognition. That is because the accuracy of feature
matching directly influences the quality of the estimated transformation
which is used to align the model to the scene, as well as the computational
time required for verification and refinement \citep{taati2011local}.
Moreover, the feature descriptors should be robust to a set of nuisances,
including noise, varying mesh resolution, clutter, occlusion,{{}
holes and topology changes \citep{bronstein2010shrec,boyer2011shrec}. }

A number of local feature descriptors exist in literature (Section
\ref{sub:Local-Surface-Feature-Description}). These descriptors can
be divided into two broad categories based on whether they use a Local
Reference Frame (LRF) or not. Feature descriptors without any LRF
use a histogram or the statistics of the local geometric information
(e.g., normal, curvature) to form a feature descriptor (Section \ref{sub:Features-without-LRF}).
Examples of this category include surface signature \citep{Yamany2002},
Local Surface Patch (LSP) \citep{chen20073d} and THRIFT \citep{flint2007thrift}.
In contrast, feature descriptors with LRF encode the spatial distribution
and/or geometric information of the neighboring points with respect
to the defined LRF (Section \ref{sub:Features-with-LRF}). Examples
include spin image \citep{Johnson1999}, Intrinsic Shape Signatures
(ISS) \citep{Zhong2009} and MeshHOG \citep{zaharescu2012keypoints}.
However, most of the existing feature descriptors still suffer from
either low descriptiveness or weak robustness \citep{bariya20123d}.

In this paper we present a highly descriptive and robust feature descriptor
together with an efficient 3D object recognition algorithm. This paper
first proposes a unique, repeatable and robust LRF for both local
feature description and object recognition (Section \ref{sec:Local-Reference-Frame}).
The LRF is constructed by performing an eigenvalue decomposition on
the scatter matrix of all the points lying on the local surface together
with a sign disambiguation technique. A novel feature descriptor,
namely Rotational Projection Statistics (RoPS), is then presented
(Section \ref{sec:Local-Surface-Description}). RoPS exhibits both
high discriminative power and strong robustness to noise, varying
mesh resolution{{} and a set of deformations.} The RoPS
feature descriptor is generated by rotationally projecting the neighboring
points onto three local coordinate planes and calculating several
statistics (e.g, central moment and entropy) of the distribution matrices
of the projected points. Finally, this paper presents a novel hierarchical
3D object recognition algorithm based on the proposed LRF and RoPS
feature descriptor (Section \ref{sec:3D-Object-Recognition-Algorithm}).
Comparative experiments on four popular datasets were performed to
demonstrate the superiority of the proposed method (Section \ref{sec:Experimental-ObjectRecognition}).

The rest of this paper is organized as follows. Section \ref{sec:Related-Work}
provides a brief literature review of local surface feature descriptors
and 3D object recognition algorithms. Section \ref{sec:Local-Reference-Frame}
introduces a novel technique for LRF definition. Section \ref{sec:Local-Surface-Description}
describes our proposed RoPS method for local surface feature description.{{}
Section \ref{sec:Performance-of-RoPS} presents the evaluation results
of the RoPS descriptor on two datasets. Section \ref{sec:3D-Object-Recognition-Algorithm}
introduces a RoPS based hierarchical algorithm for 3D object recognition.
Section \ref{sec:Experimental-ObjectRecognition} presents the results
and analysis of our 3D object recognition experiments on four datasets.
Section \ref{sec:Conclusion}  concludes this paper.}

\section{Related Work \label{sec:Related-Work}}

This section presents a brief overview of the existing main methods
for local surface feature description and local feature based 3D object
recognition.

\subsection{Local Surface Feature Description \label{sub:Local-Surface-Feature-Description}}

\subsubsection{Features without LRF\label{sub:Features-without-LRF}}

\citet{stein1992structural} proposed a splash feature by recording
the relationship between the normals of the geodesic neighboring points
and the feature point. This relationship is then encoded into a 3D
vector and finally transformed into curvatures and torsion angles.
\citet{hetzel20013d} constructed a set of features by generating
histograms using depth values, surface normals, shape indices and
their combinations. Results show that the surface normal and shape
index exhibit high discrimination capabilities. \citet{Yamany2002}
introduced a surface signature by encoding the surface curvature information
into a 2D histogram. This method can be used to estimate scaling transformations
as well as recognizing objects in 3D scenes. \citet{chen20073d} proposed
a LSP feature that encodes the shape indices and normal deviations
of the neighboring points. \citet{flint2008local} introduced a THRIFT
feature by calculating a weighted histogram of the deviation angles
between the normals of the neighboring points and the feature point.
\citet{taati2007variable} considered the selection of a good local
surface feature for 3D object recognition as an optimization problem
and proposed a set of Variable-Dimensional Local Shape Descriptors
(VD-LSD). However, the process of selecting an optimized subset of
VD-LSDs for a specific object is very time consuming \citep{taati2011local}.
{\citet{kokkinosintrinsic} proposed a generalization
of 2D shape context feature \citep{belongie2002shape} to curved surfaces,
namely Intrinsic Shape Context (ISC). The ISC is a meta-descriptor
which can be applied to any photometric or geometric field defined
on a surface. }

Without LRF, most of these methods generate a feature descriptor by
accumulating certain geometric attributes (e.g., normal, curvature)
into a histogram. Since most of the 3D spatial information is discarded
during the process of histogramming, the descriptiveness of the features
without LRF is limited \citep{tombari2010unique}.

\subsubsection{Features with LRF\label{sub:Features-with-LRF}}

\citet{chua1997point} proposed a point signature by using the distances
from the neighboring points to their corresponding projections on
a fitted plane. One merit of the point signature is that no surface
derivative is required. One of its limitations relate to the fact
that the reference direction may not be unique. It is also sensitive
to mesh resolution \citep{Mian2010}. \citet{johnson1998surface}
used the surface normal as a reference axis and proposed a spin image
representation by spinning a 2D image about the normal of a feature
point and summing up the number of points falling into the bins of
that image. The spin image is one of the most cited methods. But its
descriptiveness is relatively low and it is also sensitive to mesh
resolution \citep{Zhong2009}. \citet{Frome2004} also used the normal
vector as a reference axis and generated a 3D Shape Context (3DSC)
by counting the weighted number of points falling in the neighboring
3D spherical space. However, a reference axis is not a complete reference
frame and there is an uncertainty in the rotation around the normal
\citep{petrelli2011repeatability}.

\citet{sun2001surface} introduced an LRF by using the normal of a
feature point and an arbitrarily chosen neighboring point. Based on
the LRF, they proposed a descriptor named point's fingerprint by projecting
the geodesic circles onto the tangent plane. It was reported that
their approach outperforms the 2D histogram based methods. One major
limitation of this method is that their LRF is not unique \citep{tombari2010unique}.
\citet{Mian2006} proposed a tensor representation by defining an
LRF for a pair of oriented points and encoding the intersected surface
area into a multidimensional table. This representation is robust
to noise, occlusion and clutter. However, a pair of points are required
to define an LRF, which causes a combinatorial explosion \citep{Zhong2009}.
\citet{novatnack2008scale} used the surface normal and a projected
eigenvector on the tangent plane to define an LRF. They proposed an
EM descriptor by encoding the surface normals of the neighboring points
into a 2D domain. The effectiveness of exploiting geometric scale
variability in the EM descriptor has been demonstrated. \citet{Zhong2009}
introduced an LRF by calculating the eigenvectors of the scatter matrix
of the neighboring points of a feature point, and proposed an ISS
feature by recording the point distribution in the spherical angular
space. Since the sign of the LRF is not defined unambiguously, four
feature descriptors can be generated from a single feature point.
\citet{Mian2010} proposed a keypoint detection method and used a
similar LRF to \citet{Zhong2009} for their feature description. \citet{tombari2010unique}
analyzed the strong impact of LRF on the performance of feature descriptors
and introduced a unique and unambiguous LRF by performing an eigenvalue
decomposition on the scatter matrix of the neighboring points and
using a sign disambiguation technique. Based on the proposed LRF,
they introduced a feature descriptor called Signature of Histograms
of OrienTations (SHOT). SHOT is very robust to noise, but sensitive
to mesh resolution variation. \citet{petrelli2011repeatability} proposed
a novel LRF which aimed to estimate a repeatable LRF at the border
of a range image. \citet{zaharescu2012keypoints} proposed a MeshHOG
feature by first projecting the gradient vectors onto three planes
defined by an LRF and then calculating a two-level histogram of these
vectors.

However, none of the existing LRF definition techniques is simultaneously
unique, unambiguous, and robust to noise and mesh resolution. Besides,
most of the existing feature descriptors suffer from a number of limitations,
including a low robustness and discriminating power \citep{bariya20123d}.

\subsection{3D Object Recognition}

Most of the existing algorithms for local feature based 3D object
recognition follow a three-phase paradigm including feature matching,
 hypothesis generation and verification, and pose refinement \citep{taati2011local}.

\citet{stein1992structural} used the splash features to represent
the objects and generated hypotheses by using a set of triplets of
feature correspondences. These hypotheses are then grouped into clusters
using geometric constraints. They are finally verified through a least
square calculation. \citet{chua1997point} used point signatures of
a scene to match them against those of their models. The rigid transformation
between the scene and a candidate model was then calculated using
three pairs of corresponding points. Its ability to recognize objects
in both single-object and multi-object scenes has been demonstrated.
However, verifying each triplet of feature correspondences is very
time consuming. \citet{Johnson1999} generated point correspondences
by matching the spin images of the scene with the spin images of the
models. These point correspondences are first grouped using geometric
consistency. The groups are then used to calculate rigid transformations,
which are finally be verified. This algorithm is robust to clutter
and occlusion, and capable to recognize objects in complicated real
scenes. \citet{Yamany2002} used surface signatures as feature descriptors
and adopted a similar strategy to \citet{Johnson1999} for object
recognition. \citet{Mian2006} obtained feature correspondences and
model hypothesis by matching the tensor representations of the scene
with those of the models. The hypothesis model is then transformed
to the scene and finally verified using the Iterative Closest Point
(ICP) algorithm \citep{besl1992method}. Experimental results revealed
that it is superior in terms of recognition rate and efficiency compared
to the spin image based algorithm. \citet{Mian2010} also developed
a 3D object recognition algorithm based on keypoint matching. This
algorithm can be used to recognize objects at different and unknown
scales. \citet{taati2011local} developed a 3D object recognition
algorithm based on their proposed VD-LSD feature descriptors. The
optimal VD-LSD descriptor is selected based on the geometry of the
objects and the characteristics of the range sensors. \citet{bariya20123d}
introduced a 3D object recognition algorithm based on the EM feature
descriptor and a constrained interpretation tree.

There are some algorithms in the literature which do not follow the
aforementioned three-phase paradigm. For example, \citet{Frome2004}
performed 3D object recognition using the sum of the distances between
the scene features (i.e. 3DSC) and their corresponding model features.
This algorithm is efficient. However, it is not able to segment the
recognized object from a scene, and its effectiveness on real data
has not been demonstrated. \citet{shang2010real} proposed a Potential
Well Space Embedding (PWSE) algorithm for real-time 3D object recognition
in sparse range images. It cannot however handle clutter and therefore
requires the objects to be segmented a priori from the scene.

None of the existing object recognition algorithms has explicitly
explored the use of LRF to boost the performance of the recognition.
Moreover, most of these algorithms require three pairs of feature
correspondences to establish a transformation between a model and
a scene. This not only increases the run time due to the combinatorial
explosion of the matching pairs, but also decreases the precision
of the estimated transformation (since the chance to find three correct
feature correspondences is much lower compared to finding only one
correct correspondence).

\subsection{Paper Contributions}

{This paper is an extended version of \citep{guo2013Free,guo2013rops}.
}It has three major contributions, which are summarized as follows.

i) We introduce a unique, unambiguous and robust 3D LRF using all
the points lying on the local surface rather than just the mesh vertices.
Therefore, our proposed LRF is more robust to noise and varying mesh
resolution. We also use a novel sign disambiguation technique, our
proposed LRF is therefore unique and unambiguous. This LRF offers
a solid foundation for effective and robust feature description and
object recognition.

ii) We introduce a highly descriptive and robust RoPS feature descriptor.
RoPS is generated by rotationally projecting the neighboring points
onto three coordinate planes and encoding the rich information of
the point distribution into a set of statistics. {The proposed RoPS
descriptor has been evaluated on two datasets. Experimental
results show that RoPS achieved a high power of descriptiveness. It
is shown to be robust to a number of deformations including noise,
varying mesh resolution, rotation, holes and topology changes. (see
Section \ref{sec:Performance-of-RoPS} for details) . }

iii) We introduce an efficient hierarchical 3D object recognition
algorithm based on the LRF and RoPS feature descriptor. One major
advantage of our algorithm is, a single correct feature correspondence
is sufficient for object recognition. Moreover, by integrating our
robust LRF, the proposed object recognition algorithm can work with
any of the existing feature descriptors (e.g., spin image) in the
literature.{{} Rigorous evaluations of the proposed
3D object recognition algorithm were conducted on four different popular
datasets. Experimental results show that our algorithm achieved high
recognition rates, good efficiency and strong robustness to different
nuisances. It consistently resulted in the best recognition results
on the four datasets.}

\section{Local Reference Frame\label{sec:Local-Reference-Frame}}

A unique, repeatable and robust LRF is important for both effective
and efficient feature description and 3D object recognition. Advantages
of such an LRF are many fold. First, the repeatability of an LRF directly
affects the descriptiveness and robustness of the feature descriptor,
i.e., an LRF with a low repeatability will result in a poor performance
of feature matching \citep{petrelli2011repeatability}. Second, compared
with the methods which associate multiple descriptors to a single
feature point (e.g., ISS \citep{Zhong2009}), a unique LRF can help
to improve both the precision and the efficiency of feature matching
\citep{tombari2010unique}. Third, a robust 3D LRF helps to boost
the performance of 3D object recognition.

We propose a novel LRF by fully employing the point localization information
of the local surface. The three axes for the LRF are determined by
performing an eigenvalue decomposition on the scatter matrix of all
points lying on the local surface. The sign of each axis is disambiguated
by aligning the direction to the majority of the point scatter.

\subsection{Coordinate Axis Construction}

\begin{figure}
\begin{centering}
\includegraphics[scale=0.1]{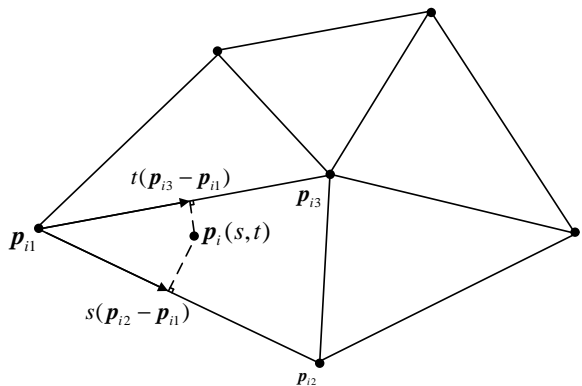}
\par\end{centering}

\caption{An illustration of a triangle mesh and a point lying on the surface.
An arbitrary point within a triangle can be represented by the triangle's
vertices. \label{fig:points-lying-on-mesh}}
\end{figure}

Given a feature point $\boldsymbol{p}$ and a support radius $r$,
the local surface mesh $S$ which contains $N$ triangles and $M$
vertices, is cropped from the range image using a sphere of radius
$r$ centered at $\boldsymbol{p}$. For the $i$th triangle with vertices
$\boldsymbol{p}_{i1}$, $\boldsymbol{p}_{i2}$ and $\boldsymbol{p}_{i3}$,
a point lying within the triangle can be represented as:

\begin{equation}
\boldsymbol{p}_{i}\left(s,t\right)=\boldsymbol{p}_{i1}+s(\boldsymbol{p}_{i2}-\boldsymbol{p}_{i1})+t\left(\boldsymbol{p}_{i3}-\boldsymbol{p}_{i1}\right),\label{eq:pi}
\end{equation}
where $0\leq s,t\leq1$, and $s+t\leq1$, as illustrated in Fig. \ref{fig:points-lying-on-mesh}.

\begin{figure*}
\begin{centering}
\subfloat[Armadillo]{\includegraphics[scale=0.3]{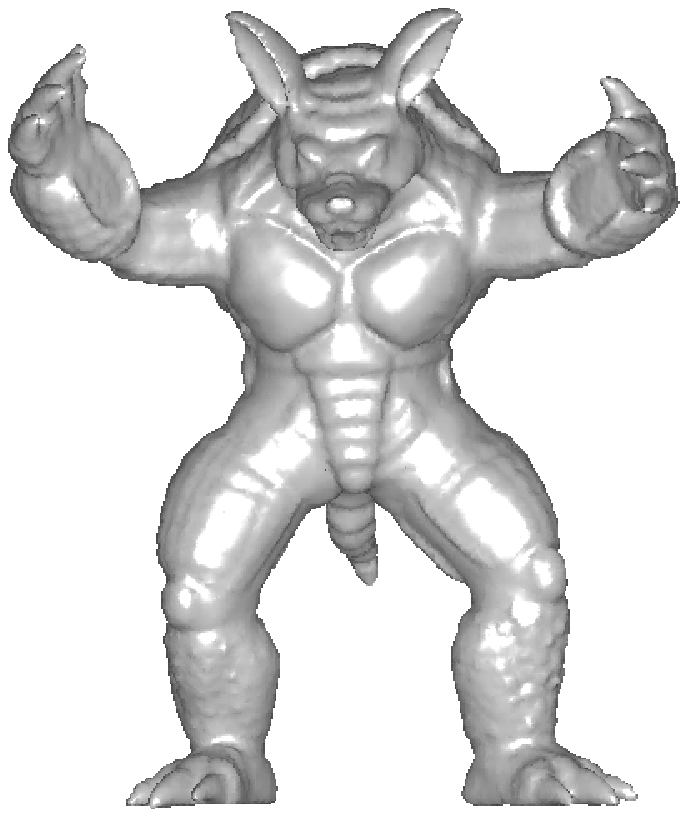}

}\,\subfloat[Asia Dragon]{\includegraphics[scale=0.3]{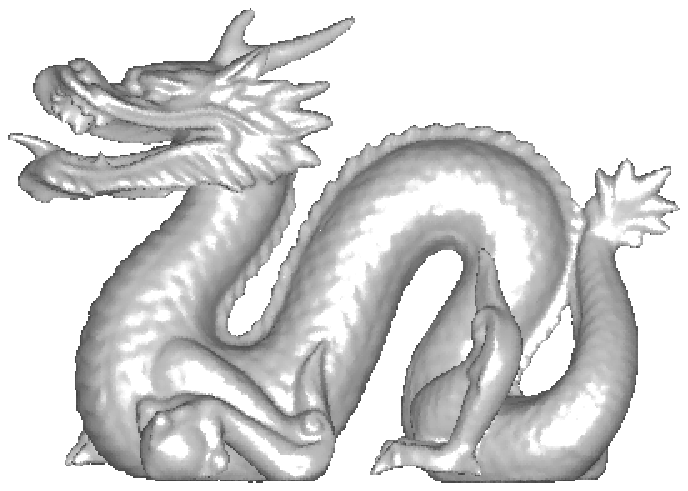}

}\,\subfloat[Bunny]{\includegraphics[scale=0.3]{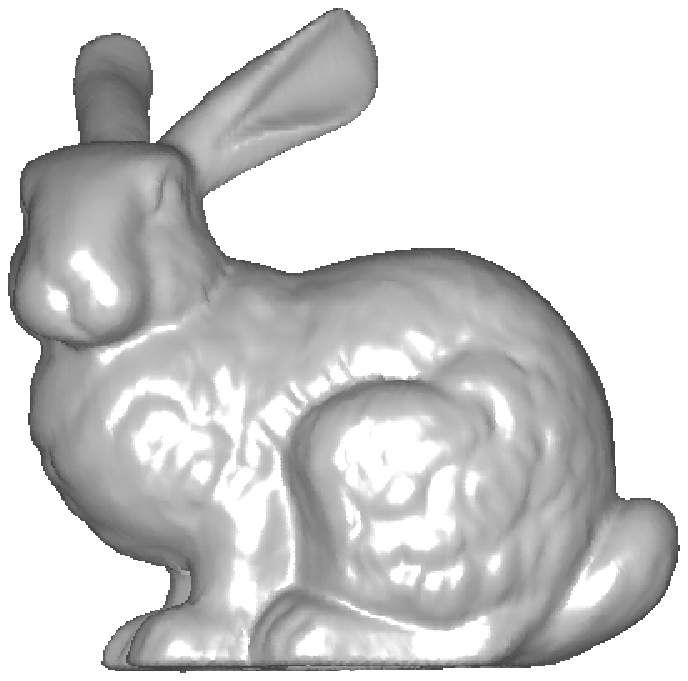}

}\,\subfloat[Dragon]{\includegraphics[scale=0.3]{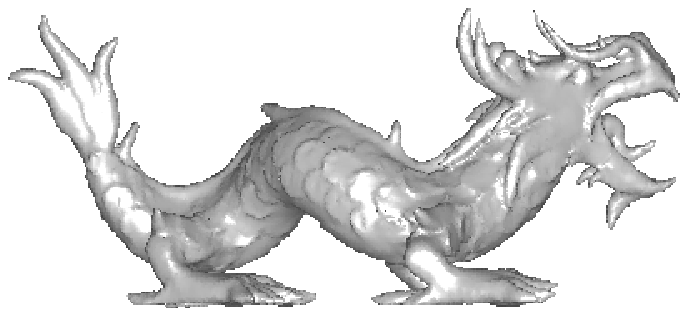}

}\,\subfloat[Happy Buddha]{\includegraphics[scale=0.3]{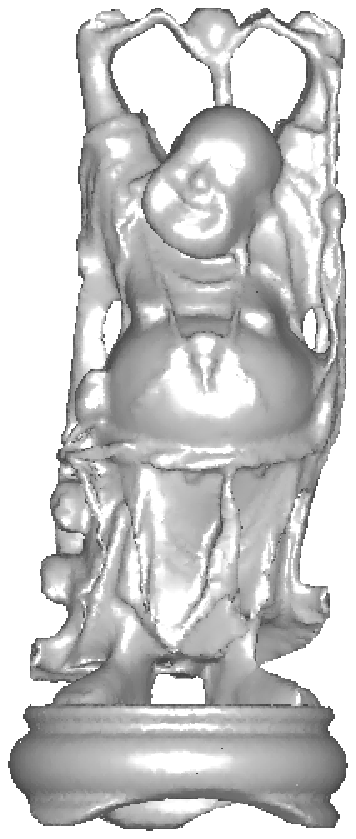}

}\,\subfloat[Thai Statue]{\includegraphics[scale=0.3]{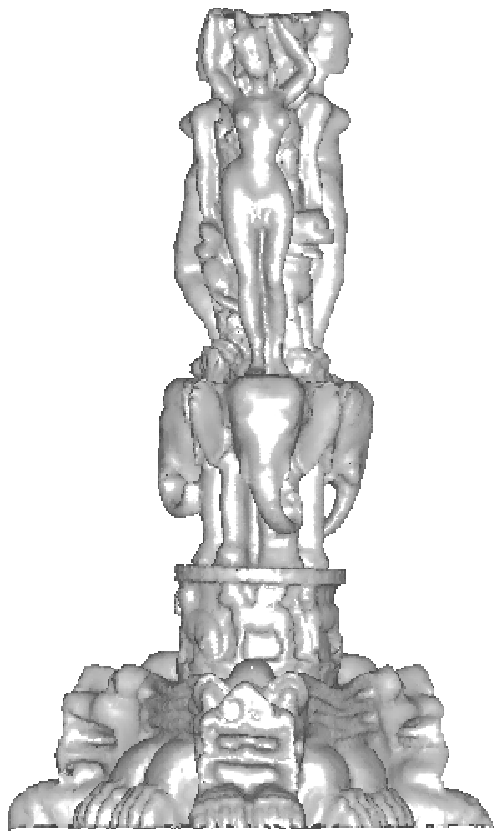}

}
\par\end{centering}

\caption{The six models of the Tuning Dataset. \label{fig:The-6-models}}
\end{figure*}

The scatter matrix $\mathbf{C}_{i}$ of all the points lying within
the $i$th triangle can be calculated as:

\begin{equation}
\mathbf{C}_{i}=\frac{\int_{0}^{1}\int_{0}^{1-s}\left(\boldsymbol{p}_{i}\left(s,t\right)-\boldsymbol{p}\right)\left(\boldsymbol{p}_{i}\left(s,t\right)-\boldsymbol{p}\right)^{\textrm{T}}dtds}{\int_{0}^{1}\int_{0}^{1-s}dtds}.
\end{equation}

Using Eq. \ref{eq:pi}, the scatter matrix $\mathbf{C}_{i}$ be can
expressed as:

\begin{eqnarray}
\mathbf{C}_{i} & = & \frac{1}{12}\sum_{j=1}^{3}\sum_{k=1}^{3}\left(\boldsymbol{p}_{ij}-\boldsymbol{p}\right)\left(\boldsymbol{p}_{ik}-\boldsymbol{p}\right)^{\textrm{T}}\nonumber \\
 &  & +\frac{1}{12}\sum_{j=1}^{3}\left(\boldsymbol{p}_{ij}-\boldsymbol{p}\right)\left(\boldsymbol{p}_{ij}-\boldsymbol{p}\right)^{\textrm{T}}.
\end{eqnarray}

The overall scatter matrix $\mathbf{C}$ of the local surface S is
calculated as the weighted sum of the scatter matrices of all the
triangles, that is:

\begin{equation}
\mathbf{C}=\sum_{i=1}^{N}w_{i1}w_{i2}\mathbf{C}_{i},\label{eq:overall scatter matrix}
\end{equation}
where $N$ is the number of triangles in the local surface $S$. Here,
$w_{i1}$ is the ratio between the area of the $i$th triangle and
the total area of the local surface $S$, that is:

\begin{equation}
w_{i1}=\frac{\left|\left(\boldsymbol{p}_{i2}-\boldsymbol{p}_{i1}\right)\times\left(\boldsymbol{p}_{i3}-\boldsymbol{p}_{i1}\right)\right|}{\sum_{i=1}^{N}\left|\left(\boldsymbol{p}_{i2}-\boldsymbol{p}_{i1}\right)\times\left(\boldsymbol{p}_{i3}-\boldsymbol{p}_{i1}\right)\right|},\label{eq:weight1}
\end{equation}
where $\times$ denotes the cross product.

$w_{i2}$ is a weight that is related to the distance from the feature
point to the centroid of the $i$th triangle, that is:

\begin{equation}
w_{i2}=\left(r-\left|\boldsymbol{p}-\frac{\boldsymbol{p}_{i1}+\boldsymbol{p}_{i2}+\boldsymbol{p}_{i3}}{3}\right|\right)^{2}.\label{eq:weight2}
\end{equation}

Note that, the first weight $w_{i1}$ is expected to improve the robustness
of LRF to varying mesh resolutions, since a compensation with respect
to the triangle area is incorporated through this weighting. The second
weight $w_{i2}$ is expected to improve the robustness of LRF to occlusion
and clutter, since distant points will contribute less to the overall
scatter matrix.

We then perform an eigenvalue decomposition on the overall scatter
matrix $\mathbf{C}$, that is:

\begin{equation}
\mathbf{C}\mathbf{V}=\mathbf{EV},
\end{equation}
where $\mathbf{E}$ is a diagonal matrix of the eigenvalues $\left\{ \lambda_{1},\lambda_{2},\lambda_{3}\right\} $
of the matrix $\mathbf{C}$, and $\mathbf{V}$ contains three orthogonal
eigenvectors $\left\{ \boldsymbol{v}_{1},\boldsymbol{v}_{2},\boldsymbol{v}_{3}\right\} $
that are in the order of decreasing magnitude of their associated
eigenvalues. The three eigenvectors offer a basis for LRF definition.
However, the signs of these vectors are numerical accidents and are
not repeatable between different trials even on the same surface \citep{bro2008resolving,tombari2010unique}.
We therefore propose a novel sign disambiguation technique which is
described in the next subsection.

It is worth noting that, although some existing techniques also use
the idea of eigenvalue decomposition to construct the LRF (e.g., \citep{Mian2010,tombari2010unique,Zhong2009}),
they calculate the scatter matrix using just the mesh vertices. Instead,
our technique employs all the points in the local surface and, is
therefore more robust compared to exiting techniques (as demonstrated
in Section \ref{sub:Robustness-of-LRF}).

\subsection{Sign Disambiguation}

In order to eliminate the sign ambiguity of the LRF, each eigenvector
should point in the major direction of the scatter vectors (which
start from the feature point and point in the direction of the points
lying on the local surface). Therefore, the sign of each eigenvector
is determined from the sign of the inner product of the eigenvector
and the scatter vectors. Specifically, the unambiguous vector $\widetilde{\boldsymbol{v}_{1}}$
is defined as:

\begin{equation}
\widetilde{\boldsymbol{v}_{1}}=\boldsymbol{v}_{1}\cdot\textrm{sign}\left(h\right),
\end{equation}
where $\mathrm{sign\left(\cdot\right)}$ denotes the signum function
that extracts the sign of a real number, and $h$ is calculated as:

\begin{eqnarray}
h & = & \sum_{i=1}^{N}w_{i1}w_{i2}\left(\int_{0}^{1}\int_{0}^{1-s}\left(\boldsymbol{p}_{i}\left(s,t\right)-\boldsymbol{p}\right)\boldsymbol{v}_{1}dtds\right)\nonumber \\
 & = & \sum_{i=1}^{N}w_{i1}w_{i2}\left(\frac{1}{6}\sum_{j=1}^{3}\left(\boldsymbol{p}_{ij}-\boldsymbol{p}\right)\boldsymbol{v}_{1}\right).\label{eq:signum function}
\end{eqnarray}

Similarly, the unambiguous vector $\widetilde{\boldsymbol{v}_{3}}$
is defined as:

\begin{equation}
\widetilde{\boldsymbol{v}_{3}}=\boldsymbol{v}_{3}\cdot\textrm{sign}\left(\sum_{i=1}^{N}w_{i1}w_{i2}\left(\frac{1}{6}\sum_{j=1}^{3}\left(\boldsymbol{p}_{ij}-\boldsymbol{p}\right)\boldsymbol{v}_{3}\right)\right).\label{eq:vector v3}
\end{equation}

Given two unambiguous vectors $\widetilde{\boldsymbol{v}_{1}}$ and
$\widetilde{\boldsymbol{v}_{3}}$, $\widetilde{\boldsymbol{v}_{2}}$
is defined as $\widetilde{\boldsymbol{v}_{3}}\times\widetilde{\boldsymbol{v}_{1}}$.
Therefore, a unique and unambiguous 3D LRF for feature point $\boldsymbol{p}$
is finally defined. Here, $\boldsymbol{p}$ is the origin, and $\widetilde{\boldsymbol{v}_{1}}$,
$\widetilde{\boldsymbol{v}_{2}}$ and $\widetilde{\boldsymbol{v}_{3}}$
are the $x$, $y$ and $z$ axes respectively. With this LRF, a unique,
pose invariant and highly discriminative local feature descriptor
can now be generated.

\subsection{Performance of the Proposed LRF \label{sub:Robustness-of-LRF}}

To evaluate the repeatability and robustness of our proposed LRF,
we calculated the LRF errors between the corresponding points in the
scenes and models. The six models (i.e., ``Armadillo'', ``Asia
Dragon'', ``Bunny'', ``Dragon'', ``Happy Buddha'' and ``Thai
Statue'') used in this experiment were taken from the Stanford 3D
Scanning Repository \citep{curless1996volumetric}. They are shown
in Fig. \ref{fig:The-6-models}. The six scenes were created by resampling
the models down to $\nicefrac{1}{2}$ of their original mesh resolution
and then adding Gaussian noise with a standard deviation of 0.1 mesh
resolution (mr) to the data. We refer to this dataset as the ``Tuning
Dataset'' in the rest of this paper.

We randomly selected 1000 points in each model and we refer to these
points as feature points. We then obtained the corresponding points
in the scene by searching the points with the smallest distances to
the feature points in the model. For each point pair $\left(\boldsymbol{p}_{Si},\boldsymbol{p}_{Mi}\right)$,
we calculated the LRFs for both points, denoted as $\mathbf{L}_{Si}$
and $\mathbf{L}_{Mi}$, respectively. Using the similar criterion
as in \citep{mian2006novel}, the error between two LRFs of the $i$th
point pair can be calculated by:

\begin{equation}
\epsilon_{i}=\arccos\left(\frac{\textrm{trace}\left(\mathbf{L}_{Si}\mathbf{L}_{Mi}^{-1}\right)-1}{2}\right)\frac{180}{\pi},
\end{equation}
where $\epsilon_{i}$ represents the amount of rotation error between
two LRFs and is zero in the case of no error.

\begin{figure}[t]
\begin{centering}
\includegraphics[scale=0.54]{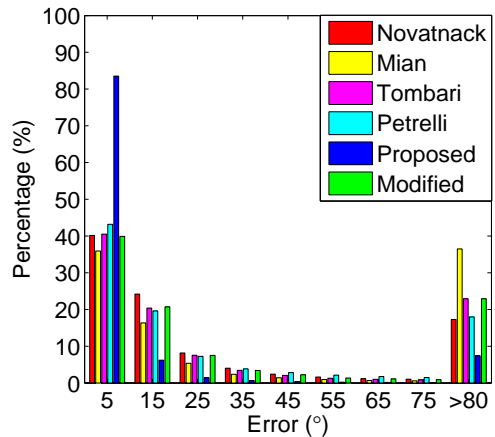}
\par\end{centering}

\centering{}\caption{Histogram of the LRF errors for the six scenes and models of the Tuning
Dataset. Our proposed technique outperformed the existing techniques
by a large margin. (Figure best seen in color.) \label{fig:Performance-LRF} }
\end{figure}

Our proposed LRF technique was tested on the Tuning Dataset with comparison
to several existing techniques, e.g., proposed by \citet{novatnack2008scale},
\citet{Mian2010}, \citet{tombari2010unique}, and \citet{petrelli2011repeatability}.{{}
We tested each LRF technique five times by randomly selecting 1000
different point pairs each time. The overall LRF errors of each technique
are shown in  Fig. \ref{fig:Performance-LRF} as a histogram. Ideally,
all of the LRF errors should lie around the zero value (in the first
bin of the histogram). }It is clear that our proposed technique performed
best, with 83.5\% of the point pairs having LRF errors less than 10
degrees. Whereas the second best one (i.e., proposed by \citet{petrelli2011repeatability})
secured only 43.2\% of the point pairs with LRF errors less than 10
degrees. Other techniques only had around 40\% point pairs with LRF
errors less than 10 degrees. These results clearly indicate that our
proposed LRF is more repeatable and more robust than the state-of-the-art
in the presence of noise and mesh resolution variation.

{In order to further assess the influence of a weighting
strategy, we used a distance weight $w_{i3}=r-\left|\boldsymbol{p}-\frac{\boldsymbol{p}_{i1}+\boldsymbol{p}_{i2}+\boldsymbol{p}_{i3}}{3}\right|$
(following the approach of \citep{tombari2010unique}) to replace
the weights $w_{i1}$ and $w_{i2}$ in Equations \ref{eq:overall scatter matrix},
\ref{eq:signum function} and \ref{eq:vector v3}, resulting in a
modified LRF. The histogram of LRF errors of the modified technique
is shown in Fig. \ref{fig:Performance-LRF}. The performance of the
modified LRF decreased significantly compared to the original proposed
LRF. This observation reveals that the weighting strategy using both
quadratic distance weight $w_{i2}$ and area weight $w_{i1}$ produced
more robust results compared to those using only a linear distance
weight $w_{i3}$.}

{Fig. \ref{fig:Performance-LRF} shows that part of
the LRF errors of each technique are larger than 80 degrees. This
is mainly due to the presence of local symmetrical surfaces (e.g.,
flat or spherical surfaces) in the scenes. For a local symmetrical
surface, there is an inherent sign ambiguity of its LRF because the
distribution of points is almost the same in all directions. In order
to deal with this case, we adopt a feature point selection technique
which uses the ratio of eigenvalues to avoid local symmetrical surfaces
(see Section \ref{sub:Candidate-Model-Generation}).}

Once an LRF is determined, the next step is to define a local surface
descriptor. In the next section, we propose a novel RoPS descriptor.

\section{Local Surface Description \label{sec:Local-Surface-Description}}

A local surface descriptor needs to be invariant to rotation and robust
to noise, varying mesh resolution, occlusion, clutter\textcolor{black}{{}
and other nuisances. }In this section, we propose a novel local surface
feature descriptor namely RoPS by performing local surface rotation,
neighboring points projection and statistics calculation.

\subsection{RoPS Feature Descriptor}

\begin{figure*}
\begin{centering}
\includegraphics[scale=0.75]{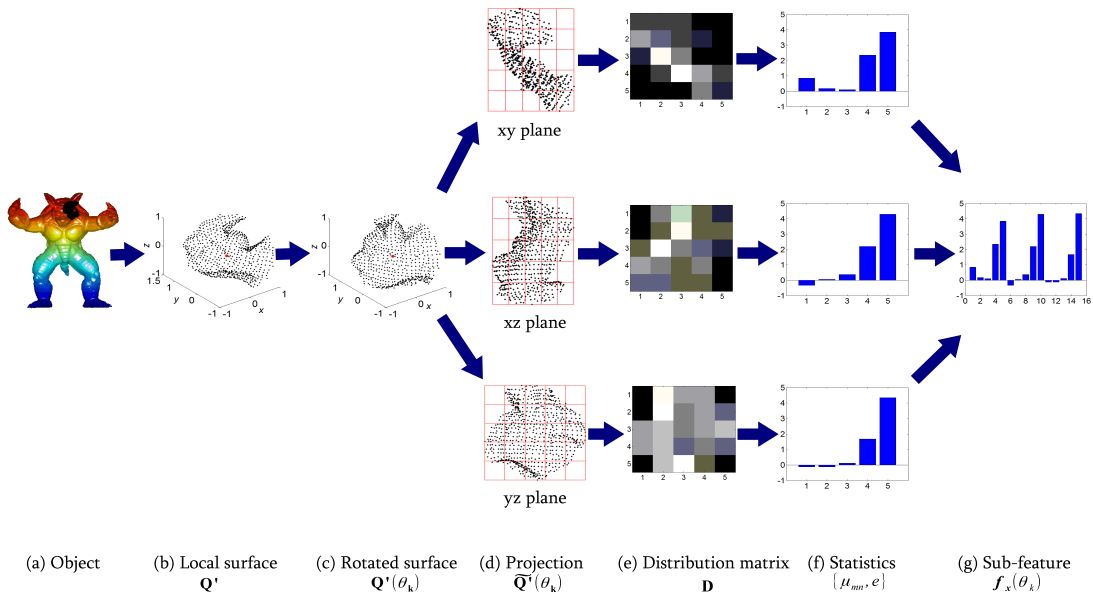}
\par\end{centering}

\caption{An illustration of the generation of a RoPS feature descriptor for
one rotation. (a) The Armadillo model and the local surface around
a feature point. (b) The local surface is cropped and transformed
in the LRF. (c) The local surface is rotated around a coordinate axis.
(d) The neighboring points are projected onto three 2D planes. (e)
A distribution matrix is obtained for each plane by partitioning the
2D plane into bins and counting up the number of points falling into
each bin. The dark color indicates a large number. (f) Each distribution
matrix is then encoded into several statistics. (g) The statistics
from three distribution matrices are concatenated to form a sub-feature
descriptor for one rotation. (Figure best seen in color.) \label{fig:Illustration-of-RoPS}}
\end{figure*}

An illustrative example of the overall RoPS method is given in Fig.
\ref{fig:Illustration-of-RoPS}. From a range image/model, a local
surface is selected for a feature point $\boldsymbol{p}$ given a
support radius $r$. Figures \ref{fig:Illustration-of-RoPS}(a) and
(b) respectively show a model and a local surface. We already have
defined the LRF for $\boldsymbol{p}$ and the vertices of the triangles
in the local surface $S$ constitute a pointcloud $\mathbf{Q}=\left\{ \boldsymbol{q}_{1},\boldsymbol{q}_{2},\ldots,\boldsymbol{q}_{M}\right\} $.
The pointcloud $\mathbf{Q}=\left\{ \boldsymbol{q}_{1},\boldsymbol{q}_{2},\ldots,\boldsymbol{q}_{M}\right\} $
is then transformed with respect to the LRF in order to achieve rotation
invariance, resulting in a transformed pointcloud $\mathbf{Q}'=\left\{ \boldsymbol{q}_{1}',\boldsymbol{q}_{2}',\ldots,\boldsymbol{q}_{M}'\right\} $.
We then follow a number of steps which are described as follows.

First, the pointcloud is rotated around the $x$ axis by an angle
$\theta_{k}$, resulting in a rotated pointcloud $\mathbf{Q}'\left(\theta_{k}\right)$,
as shown in Fig. \ref{fig:Illustration-of-RoPS}(c). This pointcloud
$\mathbf{Q}'\left(\theta_{k}\right)$ is then projected onto three
coordinate planes (i.e., the $xy$, $xz$ and $yz$ planes) to obtain
three projected pointclouds $\widetilde{\mathbf{Q}'}_{i}\left(\theta_{k}\right),i=1,2,3$.
Note that, the projection offers a means to describe the 3D local
surface in a concise and efficient manner. That is because 2D projections
clearly preserve a certain amount of unique 3D geometric information
of the local surface from that particular viewpoint.

Next, for each projected pointcloud $\widetilde{\mathbf{Q}'}_{i}\left(\theta_{k}\right)$,
a 2D bounding rectangle is obtained, which is subsequently divided
into $L\times L$ bins, as shown in Fig. \ref{fig:Illustration-of-RoPS}(d).
The number of points falling into each bin is then counted to yield
an $L\times L$ matrix $\mathbf{D}$, as shown in Fig. \ref{fig:Illustration-of-RoPS}(e).
We refer to the matrix $\mathbf{D}$ as a ``distribution matrix''
since it represents the 2D distribution of the neighboring points.
The distribution matrix $\mathbf{D}$ is further normalized such that
the sum of all bins is equal to one in order to achieve invariance
to variations in mesh resolution.

The information in the distribution matrix $\mathbf{D}$ is further
condensed in order to achieve computational and storage efficiency.
In this paper, a set of statistics is extracted from the distribution
matrix $\mathbf{D}$, including central moments \citep{demi2000first,hu1962visual}
and Shannon entropy \citep{shannon1948mathematical}. The central
moments are utilized for their mathematical simplicity and rich descriptiveness
\citep{hu1962visual}, while Shannon entropy is selected for its strong
power to measure the information contained in a probability distribution
\citep{shannon1948mathematical}.

The central moment $\mu_{mn}$ of order $m+n$ of matrix $\mathbf{D}$
is defined as:

\begin{equation}
\mu_{mn}=\sum_{i=1}^{L}\sum_{j=1}^{L}\left(i-\bar{i}\right)^{m}\left(j-\bar{j}\right)^{n}\mathbf{D}\left(i,j\right),
\end{equation}
where
\begin{equation}
\bar{i}=\sum_{i=1}^{L}\sum_{j=1}^{L}i\mathbf{D}\left(i,j\right),
\end{equation}
and

\begin{equation}
\bar{j}=\sum_{i=1}^{L}\sum_{j=1}^{L}j\mathbf{D}\left(i,j\right).
\end{equation}

The Shannon entropy $e$ is calculated as:

\begin{equation}
e=-\sum_{i=1}^{L}\sum_{j=1}^{L}\mathbf{D}\left(i,j\right)\log\left(\mathbf{D}\left(i,j\right)\right).\label{eq:pde-1}
\end{equation}

Theoretically, a complete set of central moments can be used to uniquely
describe the information contained in a matrix \citep{hu1962visual}.
However in practice, only a small subset of the central moments can
sufficiently represent the distribution matrix $\mathbf{D}$. These
selected central moments together with the Shannon entropy are then
used to form a statistics vector, as shown in Fig. \ref{fig:Illustration-of-RoPS}(f).
The three statistics vectors from the $xy$, $xz$ and $yz$ planes
are then concatenated to form a sub-feature $\boldsymbol{f}_{x}\left(\theta_{k}\right)$.
Note that $\boldsymbol{f}_{x}\left(\theta_{k}\right)$ denotes the
total statistics for the $k$th rotation around the $x$ axis, as
shown in Fig. \ref{fig:Illustration-of-RoPS}(g).

In order to encode the ``complete'' information of the local surface,
the pointcloud $\mathbf{Q}'$ is rotated around the $x$ axis by a
set of angles $\left\{ \theta_{k}\right\} ,k=1,2,\ldots,T$, resulting
in a set of sub-features $\left\{ \boldsymbol{f}_{x}\left(\theta_{k}\right)\right\} ,k=1,2,\ldots,T$.
Further, $\mathbf{Q}'$ is rotated by a set of angles around the $y$
axis and a set of sub-features $\left\{ \boldsymbol{f}_{y}\left(\theta_{k}\right)\right\} ,k=1,2,\ldots,T$
is calculated. Finally, $\mathbf{Q}'$ is rotated by a set of angles
around the $z$ axis and a set of sub-features $\left\{ \boldsymbol{f}_{z}\left(\theta_{k}\right)\right\} ,k=1,2,\ldots,T$
is calculated. The overall feature descriptor is then generated by
concatenating the sub-features of all the rotations into a vector,
that is:

\begin{equation}
\boldsymbol{f}=\left\{ \boldsymbol{f}_{x}\left(\theta_{k}\right),\boldsymbol{f}_{y}\left(\theta_{k}\right),\boldsymbol{f}_{z}\left(\theta_{k}\right)\right\} ,k=1,2,\ldots,T.
\end{equation}

{It is expected that the RoPS descriptor would be
highly discriminative (as demonstrated in Section \ref{sec:Performance-of-RoPS})
since it encodes the geometric information of a local surface from
a set of viewpoints. Note that, some existing view-based methods can
be found in the literature, such as \citep{yamauchi2006towards},
\citep{ohbuchi2008salient} and \citep{Atmosukarto2010}. However,
these methods are based on global features and originate from the
3D shape retrieval area. They are, however, not suitable for 3D object
recognition due to their sensitivity to occlusion and clutter.}

{Other related methods, however, include the spin
image \citep{Johnson1999} and snapshot \citep{malassiotis2007snapshots}
descriptors. A spin image is generated by projecting a local surface
onto a 2D plane using a cylindrical parametrization. Similarly, a
snapshot is obtained by rendering a local surface from the viewpoint
which is perpendicular to the surface. Our RoPS differs from these
methods in several aspects. First, RoPS represents a local surface
from a set of viewpoints rather than just one view (as in the case
of spin image and snapshot). Second, RoPS is associated with a unique
and unambiguous LRF, and it is invariant to rotation. In contrast,
spin image discards cylindrical angular information and snapshot is
prone to rotation. Third, RoPS is more compact than spin image and
snapshot since RoPS further encodes 2D matrices with a set of statistics.
The typical lengths of RoPS, spin image and snapshot are 135, 225
and 1600, respectively (see Table \ref{tab:Tuned-parameter-values},
\citep{Johnson1999} and \citep{malassiotis2007snapshots}).}

\subsection{RoPS Generation Parameters \label{sub:RoPS-Generation-Parameters}}

The RoPS feature descriptor has four parameters: i) the combination
of statistics, ii) the number of partition bins $L$, iii) the number
of rotations $T$ around each coordinate axis, and iv) the support
radius $r$. The performance of RoPS descriptor against different
settings of these parameters was tested on the Tuning Dataset using
the criterion of Recall vs 1-Precision Curve (RP Curve).

RP Curve is one of the most popular criteria used for the assessment
of a feature descriptor \citep{flint2008local,hou2010efficient,ke2004pca,mikolajczyk2005performance}.
It is calculated as follows: given a scene, a model and the ground
truth transformation, a scene feature is matched against all model
features to find the closest feature. If the ratio between the smallest
distance and the second smallest one is less than a threshold, then
the scene feature  and the closest model feature  are considered a
match. Further, a match is considered a true positive only if the
distance between the physical locations of the two features is sufficiently
small, otherwise it is considered a false positive. Therefore, recall
is defined as:

\begin{equation}
\textrm{recall}=\frac{\textrm{the number of true positives}}{\textrm{total number of positives}}.
\end{equation}

1-precision is defined as:

\begin{equation}
\textrm{1-precision}=\frac{\textrm{the number of false positives}}{\textrm{total number of matches}}.
\end{equation}

By varying the threshold, a RP Curve can be generated. Ideally, a
RP Curve would fall in the top left corner of the plot, which means
that the feature obtains both high recall and precision.

\begin{figure*}
\begin{centering}
\subfloat[]{\includegraphics[scale=0.75]{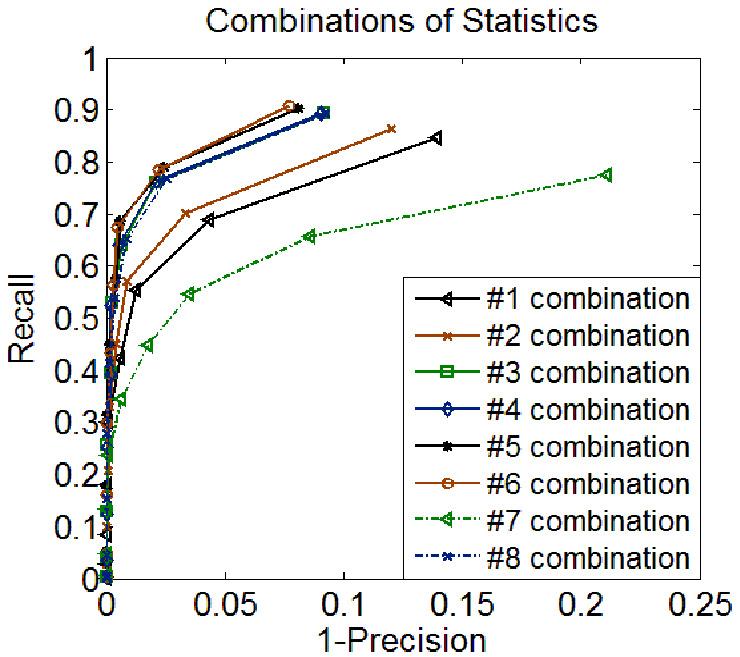}

}\,\subfloat[]{\includegraphics[scale=0.75]{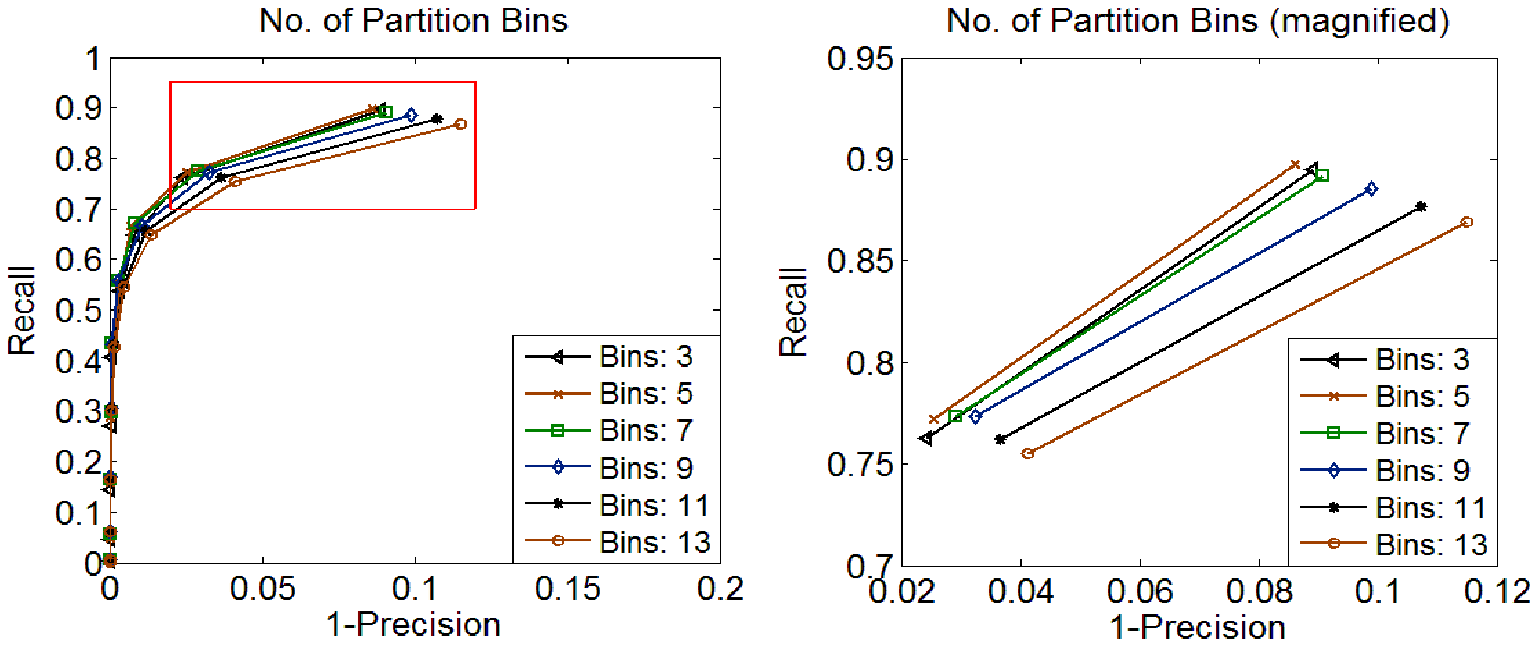}

}
\par\end{centering}

\begin{centering}
\subfloat[]{\includegraphics[scale=0.75]{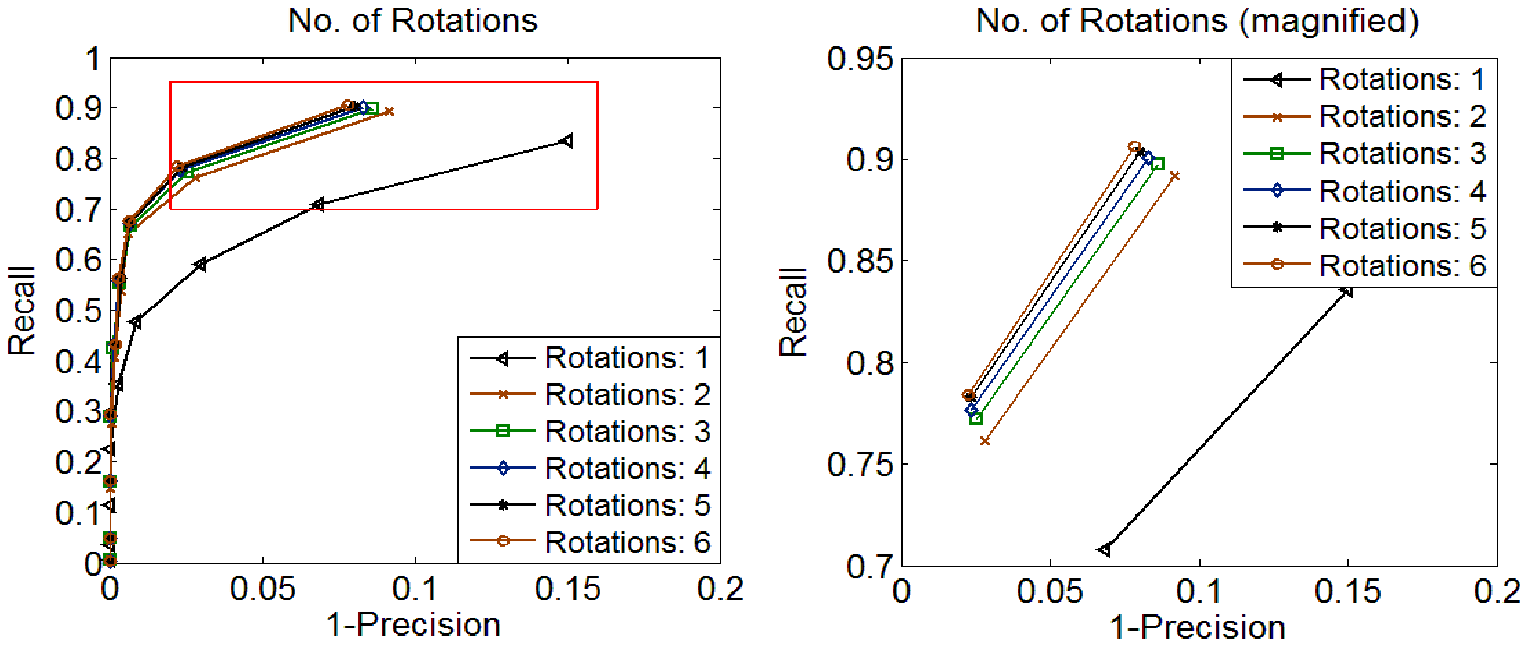}

}\,\subfloat[]{\includegraphics[scale=0.75]{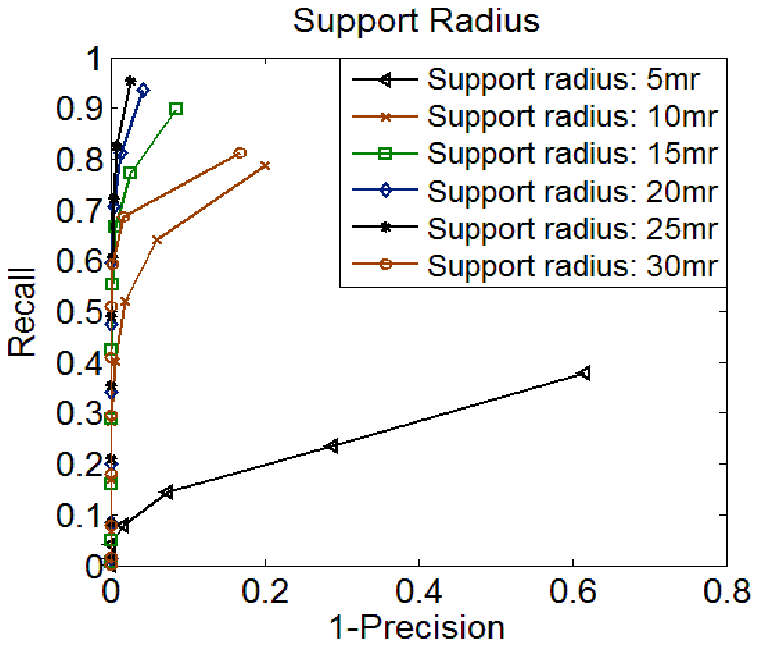}

}
\par\end{centering}

\caption{Effect of the RoPS generation parameters. (a) Different combinations
of statistics. (b) The number of partition bins $L$. There is a twin
plot in (b), where the right plot is a magnified version of the region
indicated by the rectangle in the left plot. (c) The number of rotations
$T$. There is a twin plot in (c), where the right plot is a magnified
version of the region indicated by the rectangle in the left plot.
(d) The support radius $r$. (We chose the No.6 combination of the
statistics and set $L=5$, $T=3$ and $r=15$mr in this paper as a
tradeoff between effectiveness and efficiency. \textcolor{black}{Figure
best seen in color.}) \label{fig:Effect-of-RoPS}}
\end{figure*}

\subsubsection{The Combination of Statistics }

The selection of the subset of statistics plays an important role
in the generation of a RoPS feature descriptor. It determines not
only the capability for encapsulating the information in a distribution
matrix but also the size of a feature vector. We considered eight
combinations of statistics (a number of low-order moments and entropy),
as listed in Table \ref{tab:combination-of-statistics}, and tested
the performance for each combination in the terms of RP Curve. The
other three parameters were set constant as $L=5$, $T=3$ and $r=15$mr.
It is worth noting that the zeroth-order central moment $\mu_{00}$
and the first-order central moments $\mu_{01}$ and $\mu_{10}$ were
excluded from the combinations of the statistics. Because these moments
are constant (i.e., $\mu_{00}=1$, $\mu_{01}=0$ and $\mu_{10}=0$)
and therefore contain no information of the local surface. Our experimental
results are shown in Fig. \ref{fig:Effect-of-RoPS}(a).

\begin{table}
\caption{Different combinations of the statistics.\label{tab:combination-of-statistics}}

\centering{}%
\begin{tabular}{cc}
\hline
{\scriptsize No.} & {\scriptsize Combination of the statistics}\tabularnewline
\hline
{\scriptsize 1} & {\scriptsize $\mu_{02},\mu_{11},\mu_{20}$}\tabularnewline
{\scriptsize 2} & {\scriptsize $\mu_{02},\mu_{11},\mu_{20}$,$\mu_{03},\mu_{12},\mu_{21},\mu_{30}$}\tabularnewline
{\scriptsize 3} & {\scriptsize $\mu_{02},\mu_{11},\mu_{20}$,$\mu_{03},\mu_{12},\mu_{21},\mu_{30}$,$\mu_{04},\mu_{13},\mu_{22},\mu_{31},\mu_{40}$}\tabularnewline
{\scriptsize 4} & {\scriptsize $\mu_{02},\mu_{11},\mu_{20}$,$\mu_{03},\mu_{12},\mu_{21},\mu_{30}$,$\mu_{04},\mu_{13},\mu_{22},\mu_{31},\mu_{40},e$}\tabularnewline
{\scriptsize 5} & {\scriptsize $\mu_{11},\mu_{21},\mu_{12},\mu_{22}$}\tabularnewline
{\scriptsize 6} & {\scriptsize $\mu_{11},\mu_{21},\mu_{12},\mu_{22},e$}\tabularnewline
{\scriptsize 7} & {\scriptsize $\mu_{11},\mu_{21},\mu_{12},\mu_{22},\mu_{31},\mu_{13}$}\tabularnewline
{\scriptsize 8} & {\scriptsize $\mu_{11},\mu_{21},\mu_{12},\mu_{22},\mu_{31},\mu_{13},e$}\tabularnewline
\hline
\end{tabular}
\end{table}

It is clear that the No.6 combination achieved the best performance,
followed by the No.5 combination. While the No.3, No.4 and No.8 combinations
obtained comparable performance, with recall being a little lower
than the No.6 combination. The superior performance of the No.6 combination
is due to the facts that, first, the low-order moments $\mu_{11},\mu_{21},\mu_{12},\mu_{22}$
and entropy $e$ contain the most meaningful and significant information
of the distribution matrix. Consequently, the descriptiveness of these
statistics is sufficiently high. Second, the low-order moments are
more robust to noise and varying mesh resolution compared to the high-order
moments. Beyond the high precision and recall, the size of the No.6
combination is also small, which means that the calculation and matching
of feature descriptors can be performed efficiently. Therefore, the
No.6 combination, i.e., $\left\{ \mu_{11},\mu_{21},\mu_{12},\mu_{22},e\right\} $,
was selected to represent the information in a distribution matrix
and to form the RoPS descriptor.

\subsubsection{The Number of Partition Bins}

The number of partition bins $L$ is another important parameter in
the RoPS generation. It determines both the descriptiveness and robustness
of a descriptor. That is, a dense partition of the projected points
offers more details about the point distribution, it however increases
the sensitivity to noise and varying mesh resolution. We tested the
performance of RoPS descriptor on the Tuning Dataset with respect
to a number of partition bin, while the two other parameters were
set to $T=3$ and $r=15$mr. The experimental results are shown in
Fig. \ref{fig:Effect-of-RoPS}(b) as a twin plot, where the right
plot is a magnified version of the region indicated by the rectangle
in the left plot.

The plot shows that the performance of RoPS descriptor improved as
the number of partition bins increased from 3 to 5. This is because
more details about the point distribution were encoded into the feature
descriptor. However, for a number of partition bins larger than 5,
the performance degraded as the number of partition bins increased.
This is due to the reason that a dense partition makes the distribution
matrix more susceptible to the variation of spatial position of the
neighboring points. It can therefore be inferred that 5 is the most
suitable number of partitions as a tradeoff between the descriptiveness
and the robustness to noise and varying mesh resolution. We therefore
used $L=5$ in this paper.

\subsubsection{The Numbers of Rotations}

The number of rotations $T$ determines the ``completeness'' when
describing the local surface using a RoPS feature descriptor. That
is, increasing the number of rotations means that more information
of the local surface are encoded into the overall feature descriptor.
We tested the performance of the RoPS feature descriptor with respect
to a varying number of rotations while keeping the other parameters
constant (i.e., $r=15$mr). The results are given in Fig. \ref{fig:Effect-of-RoPS}(c)
as a twin plot, where the right plot is a magnified version of the
region indicated by the rectangle in the left plot.

It was found that as the number of rotations increased, the descriptiveness
of the RoPS increased, resulting in an improvement of the matching
performance (which confirmed our assumption). Specifically, the performance
of the RoPS descriptor improved significantly as the number of rotations
increased from 1 to 2, as shown in the left plot of Fig. \ref{fig:Effect-of-RoPS}(c).
The performance then improved slightly as the number of rotations
increased from 2 to 6, as indicated in the magnified version shown
in the right plot of Fig. \ref{fig:Effect-of-RoPS}(c). In fact, there
was no notable difference between the performance with respect to
the number of rotations of 3 and 6. That is because almost all the
information of the local surface is encoded in the feature descriptor
by rotating the neighboring points 3 times around each axis. Therefore,
increasing the number of rotations any further will not necessarily
add any significant information to the feature descriptor. Moreover,
increasing the number of rotations will cost more computational and
memory resources. We therefore, set the number of rotations to be
3 in this paper.

\subsubsection{The Support Radius}

The support radius $r$ determines the amount of surface that is encoded
by the RoPS feature descriptor. The value of $r$ can be chosen depending
on how local the feature should be, and a tradeoff lies between the
feature's descriptiveness and robustness to occlusion. That is, a
large support radius enables the RoPS descriptor to encapsulate more
information of the object and therefore provides more descriptiveness.
On the other hand, a large support radius increases the sensitivity
to occlusion and clutter. We tested the performance of the RoPS feature
descriptor with respect to varying support radius while keeping the
other parameters fixed. The results are given in Fig. \ref{fig:Effect-of-RoPS}(d).

\begin{figure}
\begin{centering}
\includegraphics[scale=0.4]{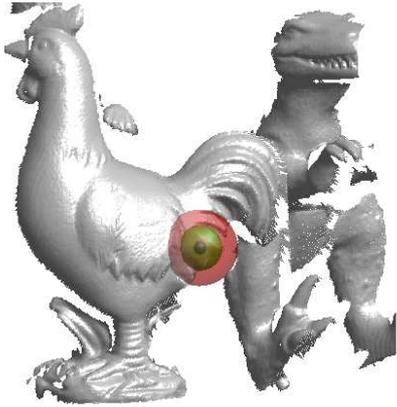}
\par\end{centering}

\caption{An illustration of the descriptor's robustness to occlusion and clutter
with respect to varying support radius. The red, green and blue spheres
respectively represent the support regions with radius of 25 mr, 15mr
and 5mr for a feature point. (Figure best seen in color.) \label{fig:OcclusionVsRadius}}
\end{figure}

The results show that the recall and precision performance of the
RoPS feature descriptor improved steadily as the support radius increased
from 5mr (mr = mesh resolution) to 25mr. Specifically, there was a
significant improvement of the matching performance as the support
radius increased from 5mr to 10mr, this is because a radius of 5mr
is too small to contain sufficient discriminating information of the
underlying surface. The RoPS feature descriptor achieved good results
with a support radius of 15mr, achieving a high precision of about
0.9 and a high recall of about 0.9.\textcolor{black}{{} Although the
performance of RoPS feature descriptor further improved slightly as
the support radius was increased to 25mr, the performance deteriorated
sharply when the support radius was set to 30mr. }We choose to set
the support radius to 15mr in the paper to maintain a strong robustness
to occlusion and clutter. An illustration is shown in Fig. \ref{fig:OcclusionVsRadius}.
The range image contains two objects in the presence of occlusion
and clutter, and a feature point is selected near the tail of the
chicken. The red, green and blue spheres, respectively represent the
support regions with radius of 25 mr, 15mr and 5mr for the feature
point. As the radius increases from 5mr to 25 mr, points on the surface
within the support region are more likely to be missing due to occlusion,
and points from other objects (e.g., T-rex on the right) are more
likely to be included in the support region due to clutter. Therefore,
the resulting feature descriptor is more likely to be affected by
occlusion and clutter.

Note that, several adaptive-scale keypoint detection methods have
been proposed for the purpose of determining the support radius based
on the inherent scale of a feature point \citep{tombari2012performance}.
However, we simply adopt a fixed support radius since our focus is
on feature description and object recognition rather than keypoint
detection. Moreover, our proposed RoPS descriptor has been demonstrated
to achieve an even better performance compared to the methods w\textcolor{black}{ith
}adaptive-scale\textcolor{black}{{} k}eypoint detection (e.g., EM matching
and keypoint matching), as analyzed in Section \ref{sec:Experimental-ObjectRecognition}.

\section{Performance of the RoPS Descriptor \label{sec:Performance-of-RoPS}}

{The descriptiveness and robustness of our proposed
RoPS feature descriptor was first evaluated on the Bologna Dataset
\citep{tombari2010unique} with respect to different levels of noise,
varying mesh resolution and their combinations. It was also evaluated
on the PHOTOMESH Dataset \citep{zaharescu2012keypoints} with respect
to 13 transformations. In these experiments, the RoPS was compared
to several state-of-the-art feature descriptors. }

\subsection{Performance on The Bologna Dataset}

\subsubsection{Dataset and Parameter Setting \label{sub:Dataset-and-Parameter}}

\begin{figure}
\begin{centering}
\includegraphics[scale=0.4]{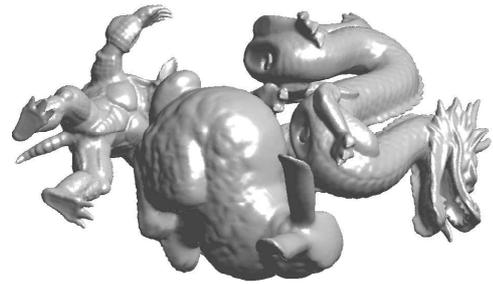}
\par\end{centering}

\caption{A scene on the Bologna Dataset. \label{fig:illustration-noise-Bologna}}
\end{figure}

The Bologna Dataset used in this paper comprises six models and 45
scenes. The six models (i.e., ``Armadillo'', ``Asia Dragon'',
``Bunny'', ``Dragon'', ``Happy Buddha'' and ``Thai Statue'')
were taken from the Stanford 3D Scanning Repository. They are shown
in Fig. \ref{fig:The-6-models}. Each scene was synthetically generated
by randomly rotating and translating three to five models in order
to create clutter and pose variances. As a result, the ground truth
rotations and translations between each model and its instances in
the scenes were known a priori during the process of construction.
An example scene is shown in Fig. \ref{fig:illustration-noise-Bologna}.

The performance of each feature descriptor was assessed using the
criterion of RP Curve (as detailed in Section \ref{sub:RoPS-Generation-Parameters}).
We compared our RoPS feature descriptor with five state-of-the-art
feature descriptors, including spin image \citep{Johnson1999}, normal
histogram (NormHist) \citep{hetzel20013d}, LSP \citep{chen20073d},
THRIFT \citep{flint2007thrift} and SHOT \citep{tombari2010unique}.
The support radius $r$ for all methods was set to be 15mr as a compromise
between the descriptiveness and the robustness to occlusion. The parameters
for generating all these feature descriptors were tuned by optimizing
the performance in terms of RP Curve on the Tuning Dataset. The tuned
parameter settings for all feature descriptors are presented in Table
\ref{tab:Tuned-parameter-values}.

\begin{table}[H]
\caption{Tuned parameter settings for six feature descriptors. \label{tab:Tuned-parameter-values}}

\centering{}%
\begin{tabular}{cccc}
\hline
 & Support Radius  & Dimensionality & Length\tabularnewline
\hline
Spin image & 15mr & 15{*}15 & 225\tabularnewline
NormHist & 15mr & 15{*}15 & 225\tabularnewline
LSP & 15mr & 15{*}15 & 225\tabularnewline
THRIFT & 15mr & 32{*}1 & 32\tabularnewline
SHOT & 15mr & 8{*}2{*}2{*}10 & 320\tabularnewline
RoPS & 15mr & 3{*}3{*}3{*}5 & 135\tabularnewline
\hline
\end{tabular}
\end{table}

In order to avoid the impact of the keypoint detection method on feature's
descriptiveness, we randomly selected 1000 feature points from each
model, and extracted their corresponding points from the scene. We
then employed the methods listed in Table \ref{tab:Tuned-parameter-values}
to extract feature descriptors for these feature points. Finally,
we calculated a RP Curve for each feature descriptor to evaluate the
performance.

\subsubsection{Robustness to Noise}

\begin{figure*}
\begin{centering}
\subfloat[Noise free]{\includegraphics[scale=0.5]{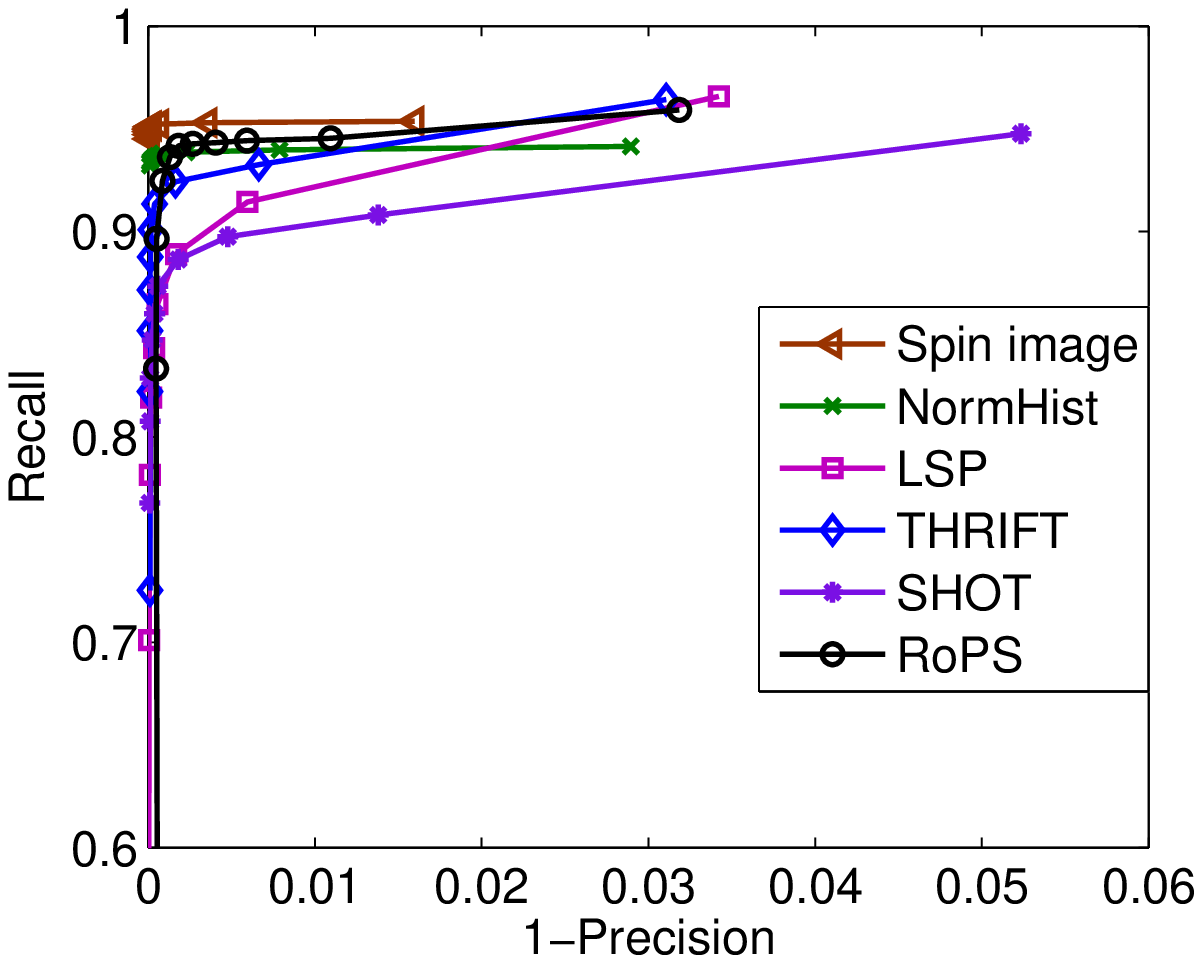}

}\,\subfloat[Noise with a standard deviation of 0.1mr]{\includegraphics[scale=0.5]{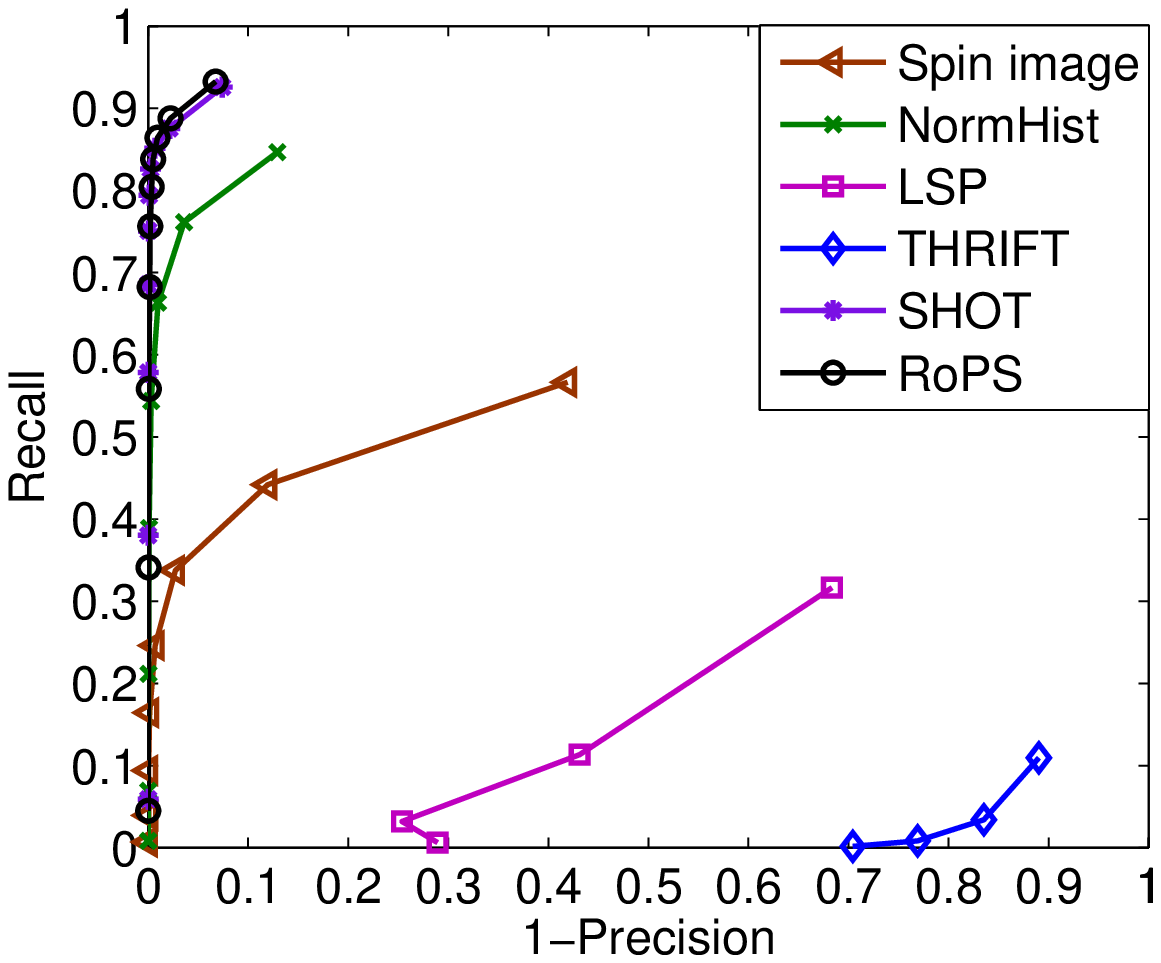}

}
\par\end{centering}

\begin{centering}
\subfloat[Noise with a standard deviation of 0.2mr]{\includegraphics[scale=0.5]{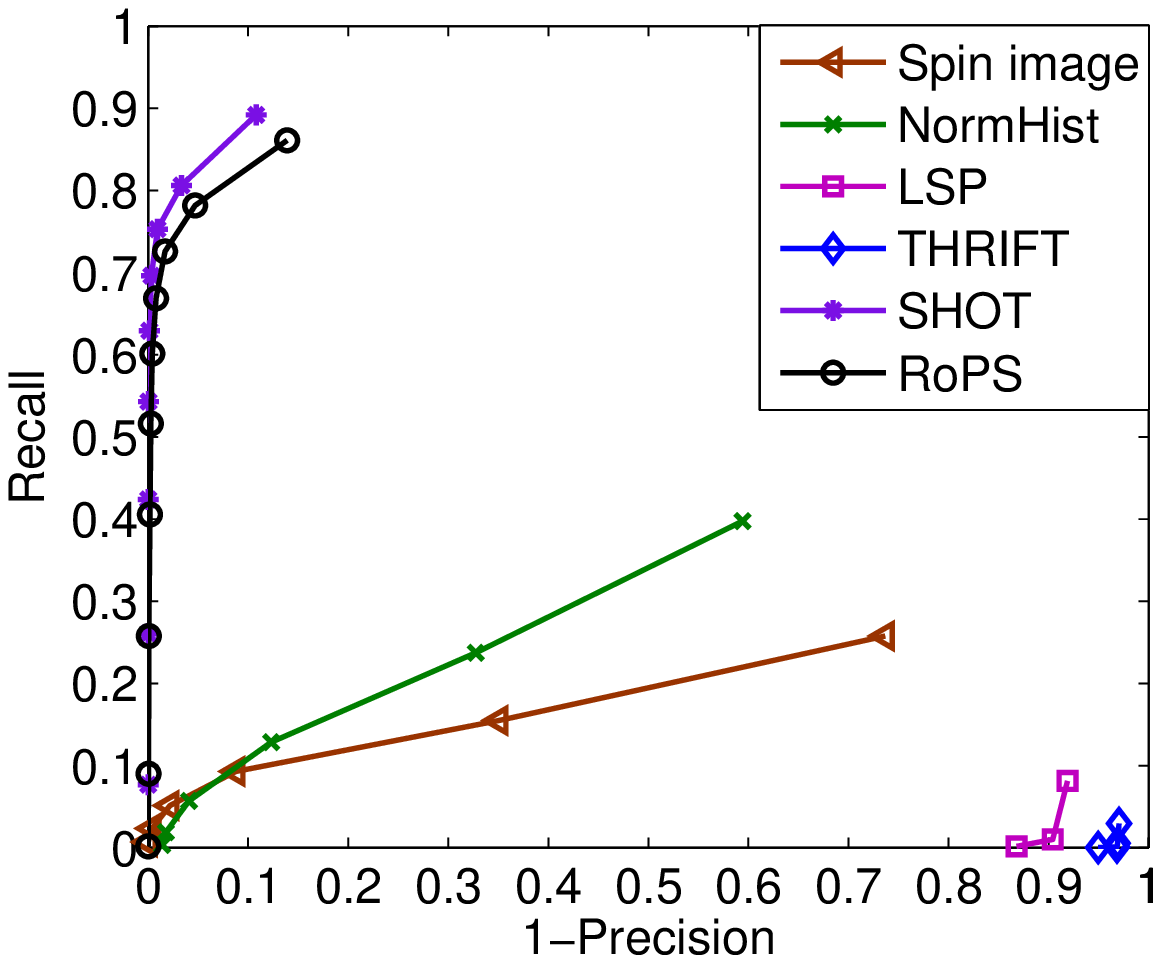}

}\,\subfloat[Noise with a standard deviation of 0.3mr]{\includegraphics[scale=0.5]{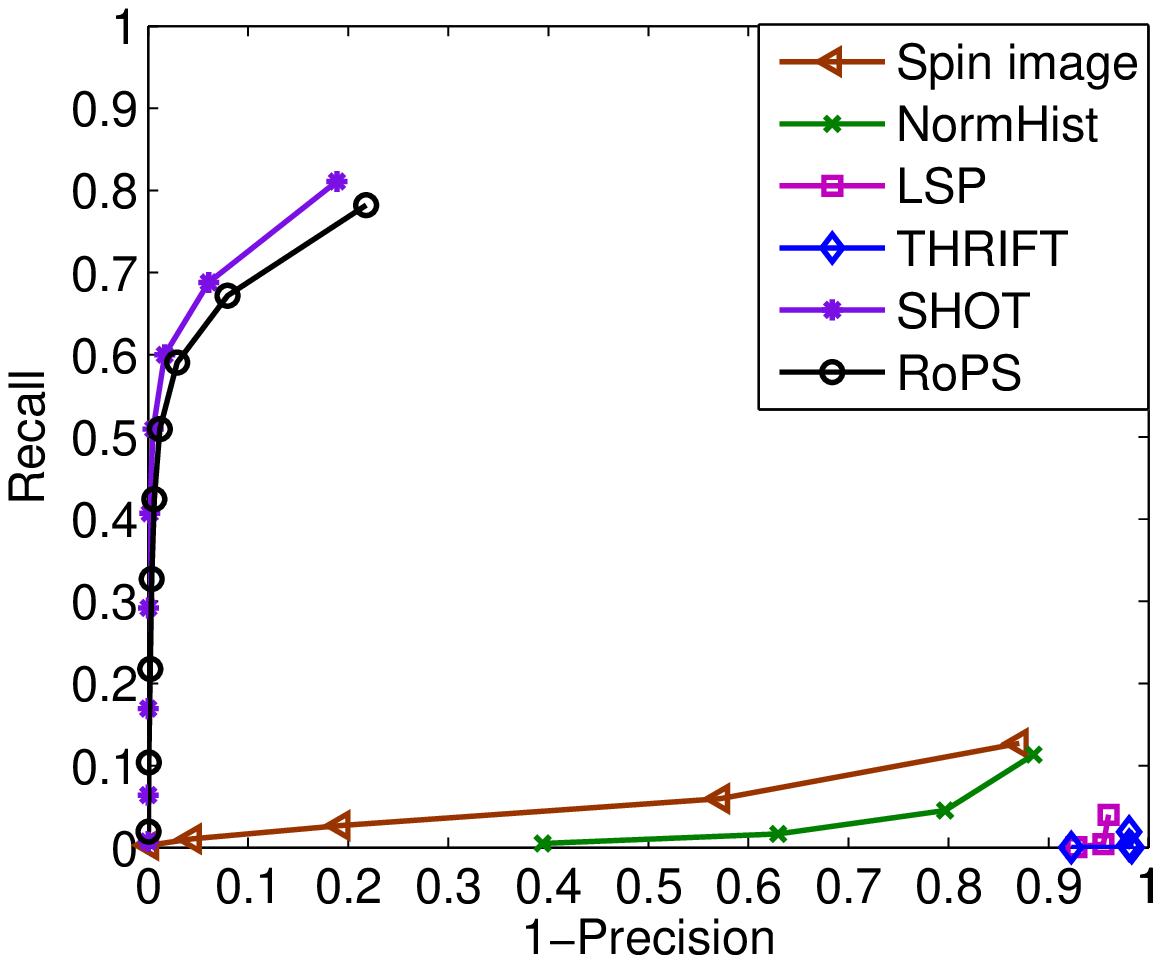}

}
\par\end{centering}

\begin{centering}
\subfloat[Noise with a standard deviation of 0.4mr]{\includegraphics[scale=0.5]{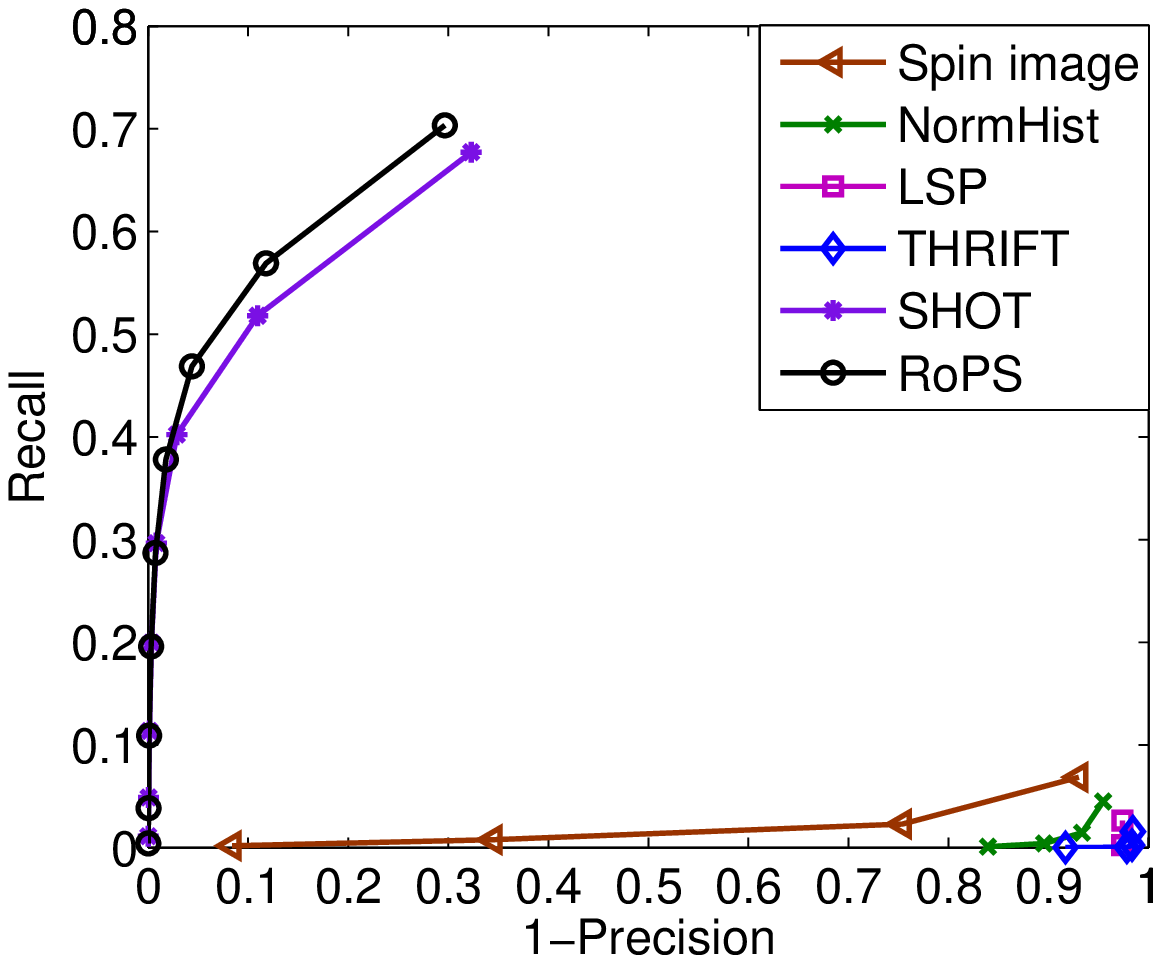}

}\,\subfloat[ Noise with a standard deviation of 0.5mr]{\includegraphics[scale=0.5]{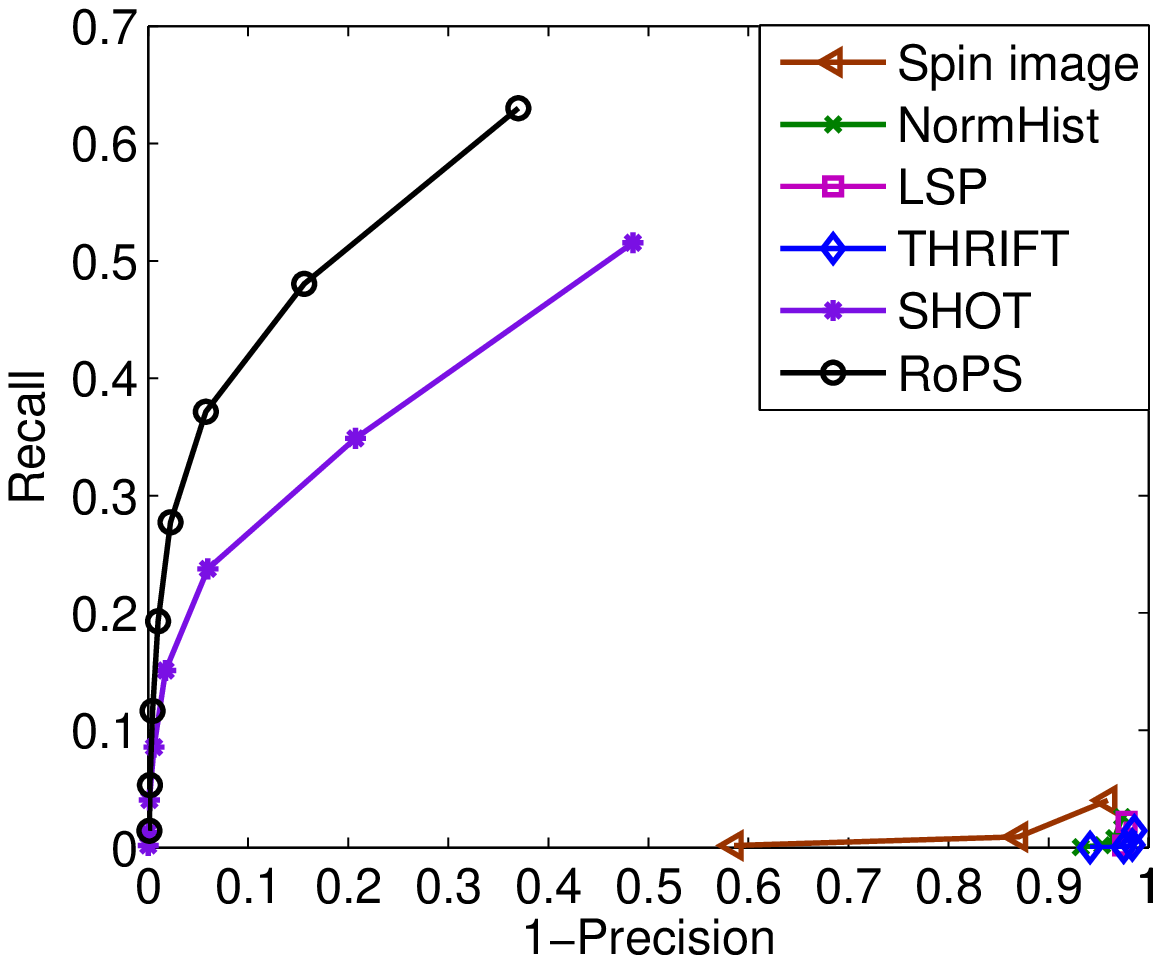}

}
\par\end{centering}

\caption{Recall vs 1-Precision curves in the presence of noise. (Figure best seen in color.) \label{fig:RP-Curve-Noise}}
\end{figure*}

\begin{figure*}
\begin{centering}
\subfloat[$\nicefrac{1}{2}$ mesh decimation]{\includegraphics[scale=0.5]{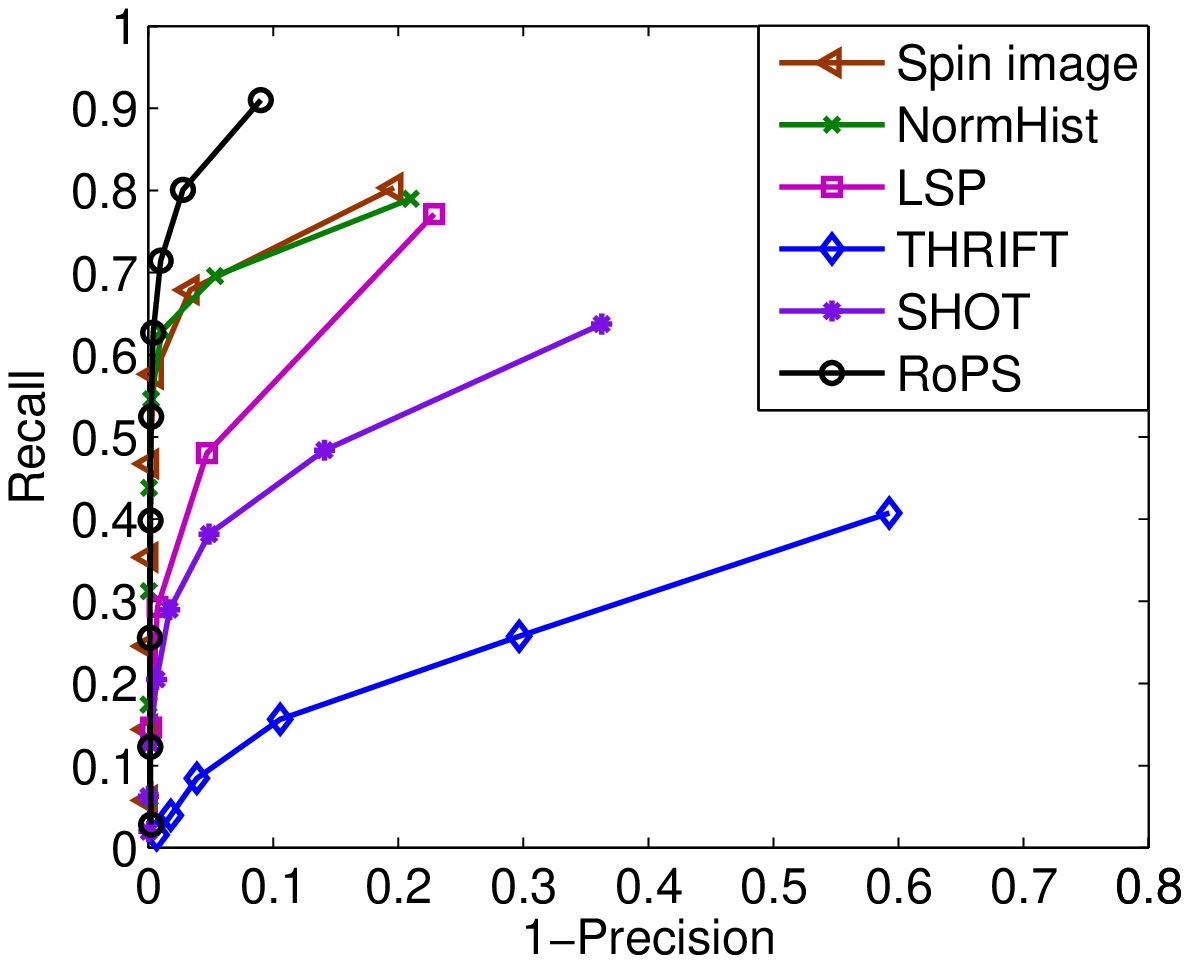}

}\,\subfloat[$\nicefrac{1}{4}$ mesh decimation]{\includegraphics[scale=0.5]{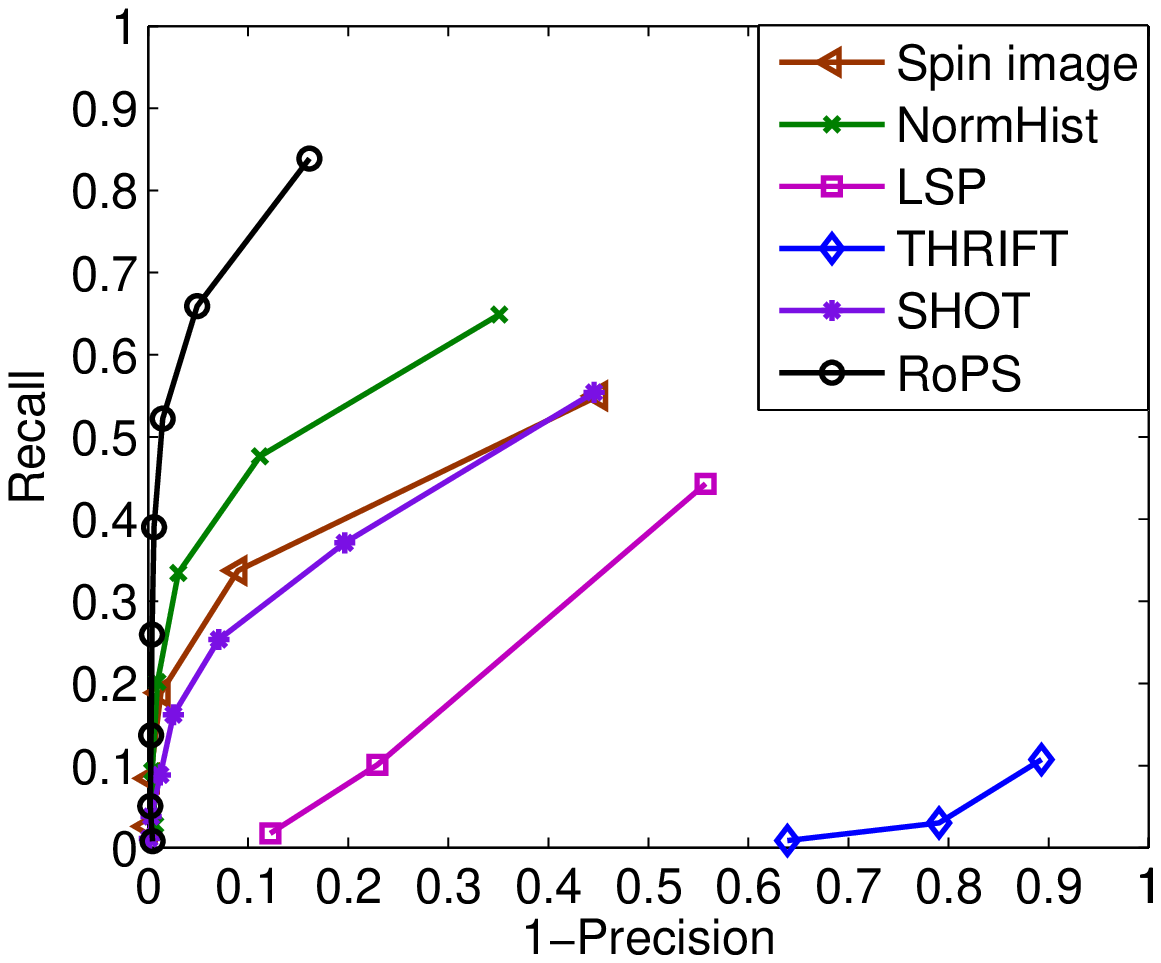}

}
\par\end{centering}

\begin{centering}
\subfloat[$\nicefrac{1}{8}$ mesh decimation]{\includegraphics[scale=0.5]{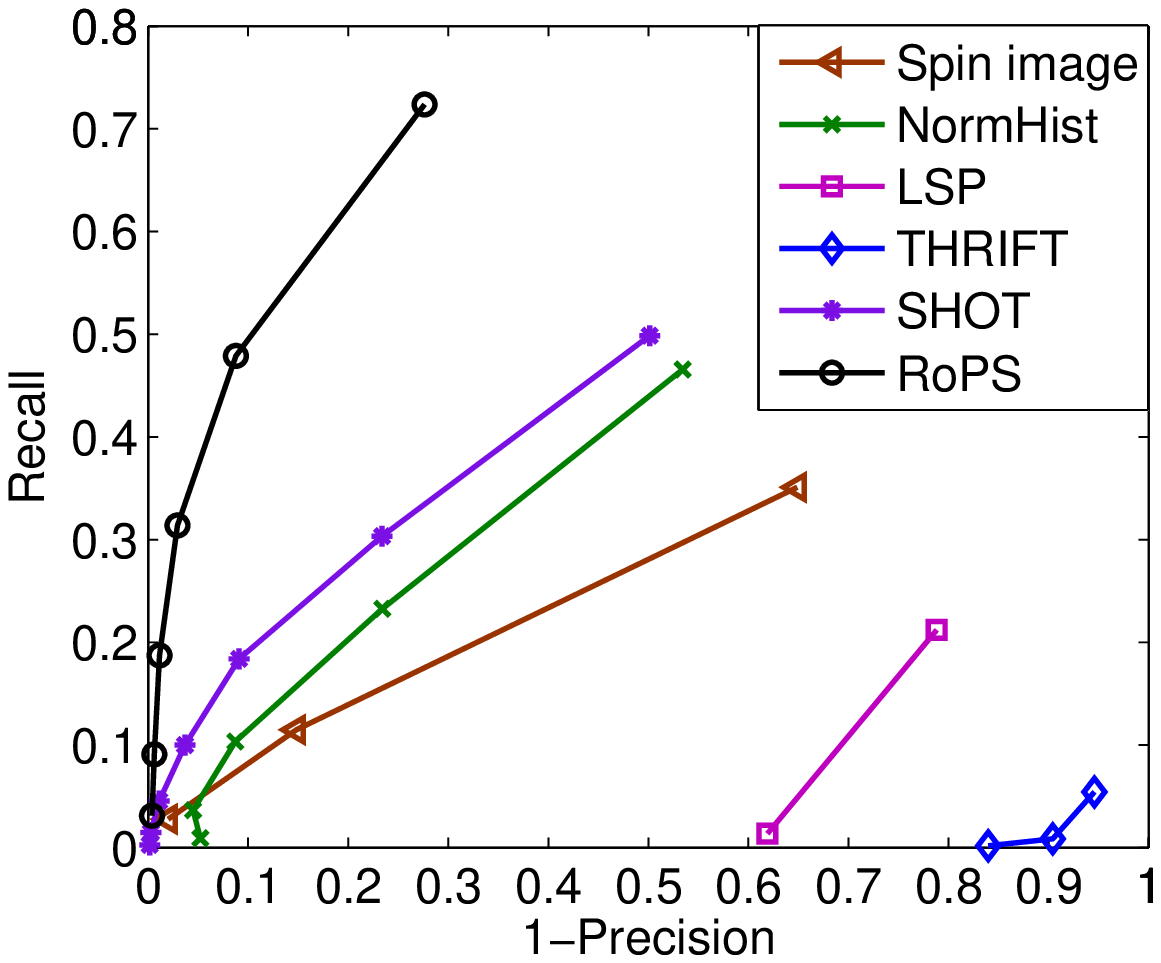}

}\,\subfloat[$\nicefrac{1}{2}$ mesh decimation and 0.1mr Gaussian noise]{\includegraphics[scale=0.5]{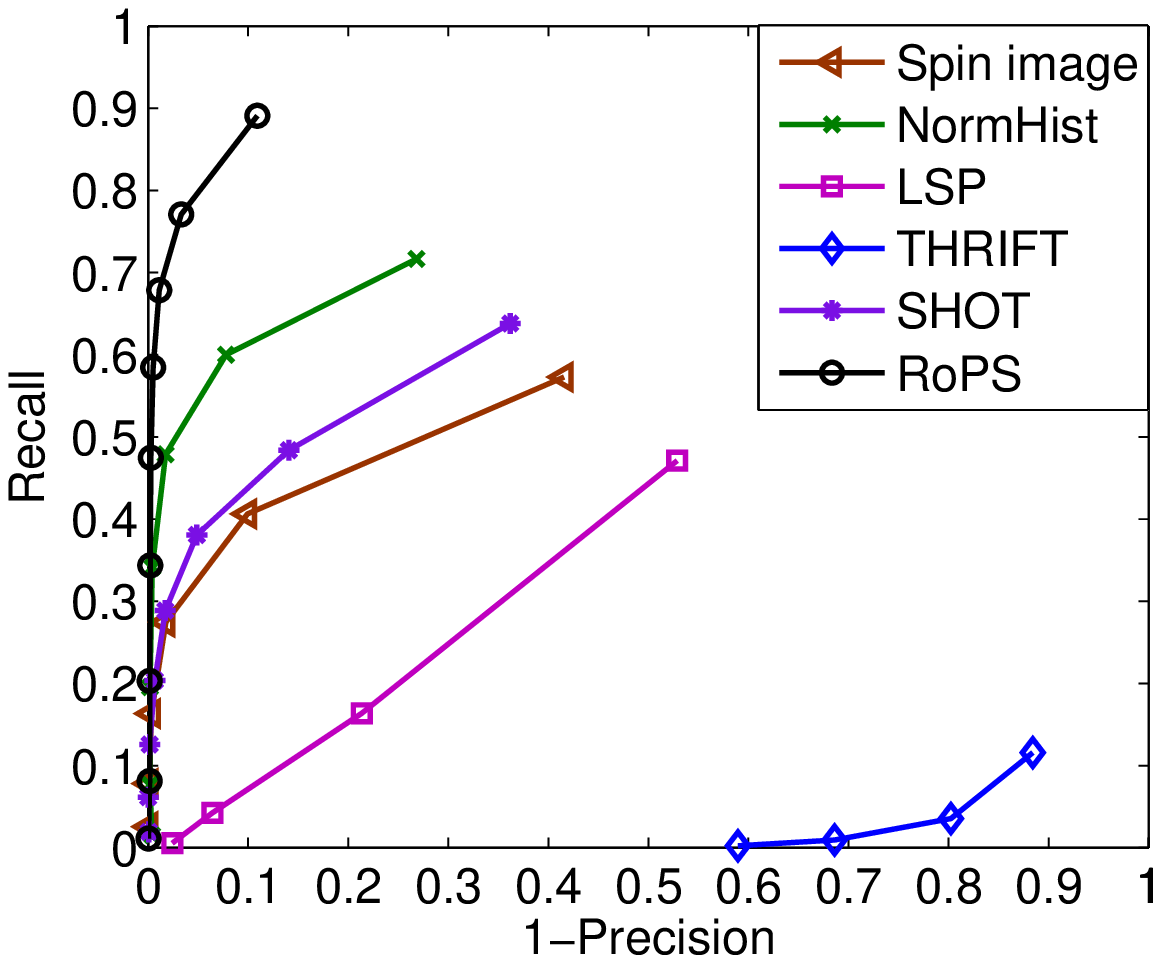}

}
\par\end{centering}

\caption{Recall vs 1-Precision curves with respect to mesh resolution. (Figure best seen in color.) \label{fig:RP-Curve-decimation}}
\end{figure*}

In order to evaluate the robustness of these feature descriptors to
noise, we added a Gaussian noise with increasing standard deviation
of 0.1mr, 0.2mr, 0.3mr, 0.4mr and 0.5mr to the scene data. The RP
Curves under different levels of noise are presented in Fig. \ref{fig:RP-Curve-Noise}.

We made a number of observations. i) These feature descriptors achieved
comparable performance on noise free data, with high recall together
with high precision, as shown in Fig. \ref{fig:RP-Curve-Noise}(a).

ii) With noise, our proposed RoPS feature descriptor achieved the
best performance in most cases, and is followed by SHOT. Specifically,
the performance of RoPS is better than SHOT under a low-level noise
with a standard deviation of 0.1mr, as shown in Fig. \ref{fig:RP-Curve-Noise}(b).
As the standard deviation of the noise increased to 0.2mr and 0.3mr,
SHOT performed slightly better than RoPS, as indicated in Figures
\ref{fig:RP-Curve-Noise}(c) and (d). However, the performance of
our proposed RoPS was significantly better than SHOT under high levels
of noise, e.g., with a noise deviation larger than 0.3mr, as shown
in Figures \ref{fig:RP-Curve-Noise}(e) and (f). It can be inferred
that RoPS is very robust to noise, particularly in the case of scenes
with a high level of noise.

iii) As the noise level increased, the performance of LSP and THRIFT
deteriorated sharply, as shown in Figures \ref{fig:RP-Curve-Noise}(b-e).
THRIFT failed to work even under a low-level of noise with a standard
deviation of 0.1mr. This result is also consistent with the conclusion
given in \citep{flint2008local}. Although NormHist and spin image
worked relatively well under low- and medium-level noise with a standard
deviation less than 0.2mr, they failed completely under noise with
a large standard deviation. The sensitivity of spin image, NormHist,
THR-IFT and LSP to noise is due to the fact that, they rely on surface
normals to generate their feature descriptors. Since the calculation
of surface normal includes a process of differentiation, it is very
susceptible to noise.

iv) The strong robustness of our RoPS feature descriptor to noise
can be explained by at least three facts. First, RoPS encodes the
``complete'' information of the local surface from various viewpoints
through rotation and therefore, encodes more information than the
existing methods. Second, RoPS only uses the low-order moments of
the distribution matrices to form its feature descriptor and is therefore
less affected by noise. Third, our proposed unique, unambiguous and
stable LRF also helps to increase the descriptiveness and robustness
of the RoPS feature descriptor.

\subsubsection{Robustness to Varying Mesh Resolution}

In order to evaluate the robustness of these feature descriptors to
varying mesh resolution, we resampled the noise free scene meshes
to $\nicefrac{1}{2}$, $\nicefrac{1}{4}$ and $\nicefrac{1}{8}$ of
their original mesh resolution. The RP Curves under different levels
of mesh decimation are presented in Figures \ref{fig:RP-Curve-decimation}(a-c).

It was found that our proposed RoPS feature descriptor outperformed
all the other descriptors by a large margin under all levels of mesh
decimation. It is also notable that the performance of our RoPS feature
descriptor with $\nicefrac{1}{8}$ of original mesh resolution was
even comparable to the best results given by the existing feature
descriptors with $\nicefrac{1}{2}$ of original mesh resolution. Specifically,
RoPS obtained a precision more than 0.7 and a recall more than 0.7
with $\nicefrac{1}{8}$ of original mesh resolution, whereas spin
image obtained a precision around 0.8 and a recall around 0.8 with
$\nicefrac{1}{2}$ of original mesh resolution, as shown in Figures
\ref{fig:RP-Curve-decimation}(a) and (c). This indicated that our
RoPS feature descriptor is very robust to varying mesh resolution.

The strong robustness of RoPS to varying mesh resolution is due to
at least two factors. First, the LRF of RoPS is derived by calculating
the scatter matrix of all the points lying on the local surface rather
than just the vertices, which makes RoPS robust to different mesh
resolution. Second, the 2D projection planes are sparsely partitioned
and only the low-order moments are used to form the feature descriptor,
which further improves the robustness of our method to mesh resolution.

\subsubsection{Robustness to Combined Noise and Mesh Decimation}

In order to further test the robustness of these feature descriptors
to combined noise and mesh decimation, we resampled the scene meshes
down to $\nicefrac{1}{2}$ of their original mesh resolution and added
a Gaussian random noise with a standard deviation of 0.1mr to the
scenes. The resulting RP Curves are presented in Fig. \ref{fig:RP-Curve-decimation}(d).

As shown in Fig. \ref{fig:RP-Curve-decimation}(d), RoPS significantly
outperformed the other methods in the scenes with both noise and mesh
decimation, obtaining a high precision of about 0.9 and a high recall
of about 0.9. It is followed by NormHist, SHOT, spin image and LSP,
while THRIFT failed to work.

As summarized in Table \ref{tab:Tuned-parameter-values}, the RoPS
feature descriptor length is 135, while the others such as spin image,
NormHist, LSP and SHOT are 225, 225, 225 and 320, respectively. So
RoPS is more compact and therefore more efficient for feature matching
compared to these methods. Note that, although the length of THRIFT
is smaller than RoPS, THRIFT's performance in terms of recall and
precision results is surpassed by our RoPS feature descriptor by a
large margin.

\subsection{Performance on The PHOTOMESH Dataset \label{sub:Performance-on-PhotoMesh}}

The PHOTOMESH Dataset contains three null shapes.
Two of the null shapes were obtained with multi-view stereo reconstruction
algorithms, and the other one was generated with a modeling program.
13 transformations were applied to each shape. The transformations
include color noise, color shot noise, geometry noise, geometry shot
noise, rotation, scale, local scale, sampling, hole, micro-hole, topology
changes and isometry. Each transformation has five different levels
of strength.

To make a rigorous comparison with \citep{zaharescu2012keypoints},
we set the support radius $r$ to $\sqrt{\nicefrac{\alpha_{r}A_{M}}{\pi}}$,
where $A_{M}$ is the total area of a mesh, and $\alpha_{r}$ is 2\%.
RoPS feature descriptors were calculated at all points of the shapes, without any
feature detection.
We used the average normalized $L_{2}$ distance between the feature
descriptors of corresponding points to measure the quality of a feature
descriptor, as in \citep{zaharescu2012keypoints}.
The experimental results of the RoPS descriptor are shown in Table
\ref{tab:Robustness-of-RoPS-PhotoMesh}. For comparison, the results
of the MeshHOG descriptor (Gaussian curvature) without and with MeshDOG
are also reported in Tables \ref{tab:Robustness-of-MeshHOG-PhotoMesh-without-DoG} and
\ref{tab:Robustness-of-MeshHOG-PhotoMesh}, respectively.

{The RoPS descriptor was clearly invariant to color
noise and color shot noise. Because the geometric information used
in RoPS cannot be affected by color deformations. RoPS was also invariant
to rotation and scale, which means that it was invariant to rigid
transformations.}

{The RoPS descriptor turned out to be very robust
to geometry noise, geometry shot noise, local scale, holes, micro-holes,
topology and isometry with noise. The average normalized $L_{2}$
distances for all these transformations were no more than 0.06, even
under the highest level of transformations. The biggest challenge
for RoPS descriptor was sampling. The average normalized $L_{2}$
distance increased from 0.01 to 0.06 as the strength level changed
from 1 to 5. However, RoPS was more robust to sampling than MeshHOG.
As shown in Tables \ref{tab:Robustness-of-RoPS-PhotoMesh} and \ref{tab:Robustness-of-MeshHOG-PhotoMesh-without-DoG},
the average normalized $L_{2}$ distance of RoPS with a strength level
of 5 was even smaller than that of MeshHOG with a strength level of
1, i.e., 0.02 and 0.04, respectively. Overall, the average normalized
$L_{2}$ distances of RoPS descriptor were much smaller under all
strength levels of all transformations compared to MeshHOG. }

\begin{table}
\caption{Robustness of RoPS descriptor.\label{tab:Robustness-of-RoPS-PhotoMesh}}
\centering{}%
\begin{tabular}{cccccc}
 & \multicolumn{5}{c}{\textbf{\footnotesize Strength}}\tabularnewline
\cline{2-6}
\textbf{\footnotesize Transform.} & {\footnotesize 1} & {\footnotesize $\leq$2} & {\footnotesize $\leq$3} & {\footnotesize $\leq$4} & {\footnotesize $\leq$5}\tabularnewline
\hline
{\footnotesize Color Noise} & {\footnotesize 0.00} & {\footnotesize 0.00} & {\footnotesize 0.00} & {\footnotesize 0.00} & {\footnotesize 0.00}\tabularnewline
{\footnotesize Color Shot Noise} & {\footnotesize 0.00} & {\footnotesize 0.00} & {\footnotesize 0.00} & {\footnotesize 0.00} & {\footnotesize 0.00}\tabularnewline
{\footnotesize Geometry Noise} & {\footnotesize 0.01} & {\footnotesize 0.01} & {\footnotesize 0.01} & {\footnotesize 0.02} & {\footnotesize 0.02}\tabularnewline
{\footnotesize Geometry Shot Noise} & {\footnotesize 0.01} & {\footnotesize 0.01} & {\footnotesize 0.02} & {\footnotesize 0.03} & {\footnotesize 0.05}\tabularnewline
{\footnotesize Rotation} & {\footnotesize 0.00} & {\footnotesize 0.00} & {\footnotesize 0.00} & {\footnotesize 0.00} & {\footnotesize 0.00}\tabularnewline
{\footnotesize Scale} & {\footnotesize 0.00} & {\footnotesize 0.00} & {\footnotesize 0.00} & {\footnotesize 0.00} & {\footnotesize 0.00}\tabularnewline
{\footnotesize Local Scale} & {\footnotesize 0.01} & {\footnotesize 0.01} & {\footnotesize 0.02} & {\footnotesize 0.02} & {\footnotesize 0.02}\tabularnewline
{\footnotesize Sampling} & {\footnotesize 0.01} & {\footnotesize 0.02} & {\footnotesize 0.04} & {\footnotesize 0.05} & {\footnotesize 0.06}\tabularnewline
{\footnotesize Holes} & {\footnotesize 0.01} & {\footnotesize 0.01} & {\footnotesize 0.01} & {\footnotesize 0.01} & {\footnotesize 0.02}\tabularnewline
{\footnotesize Marco-Holes} & {\footnotesize 0.00} & {\footnotesize 0.01} & {\footnotesize 0.01} & {\footnotesize 0.01} & {\footnotesize 0.01}\tabularnewline
{\footnotesize Topology} & {\footnotesize 0.01} & {\footnotesize 0.01} & {\footnotesize 0.02} & {\footnotesize 0.02} & {\footnotesize 0.03}\tabularnewline
{\footnotesize Isometry + Noise} & {\footnotesize 0.02} & {\footnotesize 0.02} & {\footnotesize 0.01} & {\footnotesize 0.02} & {\footnotesize 0.02}\tabularnewline
\hline
\textbf{\footnotesize Average} & {\footnotesize 0.00} & {\footnotesize 0.01} & {\footnotesize 0.01} & {\footnotesize 0.02} & {\footnotesize 0.02}\tabularnewline
\hline
\end{tabular}
\end{table}

\begin{table}
\caption{Robustness of MeshHOG (Gaussian curvature) without MeshDOG detector. \label{tab:Robustness-of-MeshHOG-PhotoMesh-without-DoG}}
\centering{}%
\begin{tabular}{cccccc}
 & \multicolumn{5}{c}{\textbf{\footnotesize Strength}}\tabularnewline
\cline{2-6}
\textbf{\footnotesize Transform.} & {\footnotesize 1} & {\footnotesize $\leq$2} & {\footnotesize $\leq$3} & {\footnotesize $\leq$4} & {\footnotesize $\leq$5}\tabularnewline
\hline
{\footnotesize Color Noise} & {\footnotesize 0.00} & {\footnotesize 0.00} & {\footnotesize 0.00} & {\footnotesize 0.00} & {\footnotesize 0.00}\tabularnewline
{\footnotesize Color Shot Noise} & {\footnotesize 0.00} & {\footnotesize 0.00} & {\footnotesize 0.00} & {\footnotesize 0.00} & {\footnotesize 0.00}\tabularnewline
{\footnotesize Geometry Noise} & {\footnotesize 0.07} & {\footnotesize 0.08} & {\footnotesize 0.09} & {\footnotesize 0.10} & {\footnotesize 0.11}\tabularnewline
{\footnotesize Geometry Shot Noise} & {\footnotesize 0.02} & {\footnotesize 0.03} & {\footnotesize 0.05} & {\footnotesize 0.06} & {\footnotesize 0.09}\tabularnewline
{\footnotesize Rotation} & {\footnotesize 0.00} & {\footnotesize 0.00} & {\footnotesize 0.00} & {\footnotesize 0.00} & {\footnotesize 0.00}\tabularnewline
{\footnotesize Scale} & {\footnotesize 0.00} & {\footnotesize 0.00} & {\footnotesize 0.00} & {\footnotesize 0.00} & {\footnotesize 0.00}\tabularnewline
{\footnotesize Local Scale} & {\footnotesize 0.06} & {\footnotesize 0.07} & {\footnotesize 0.08} & {\footnotesize 0.09} & {\footnotesize 0.10}\tabularnewline
{\footnotesize Sampling} & {\footnotesize 0.10} & {\footnotesize 0.12} & {\footnotesize 0.13} & {\footnotesize 0.13} & {\footnotesize 0.13}\tabularnewline
{\footnotesize Holes} & {\footnotesize 0.01} & {\footnotesize 0.02} & {\footnotesize 0.04} & {\footnotesize 0.03} & {\footnotesize 0.05}\tabularnewline
{\footnotesize Marco-Holes} & {\footnotesize 0.01} & {\footnotesize 0.01} & {\footnotesize 0.03} & {\footnotesize 0.04} & {\footnotesize 0.04}\tabularnewline
{\footnotesize Topology} & {\footnotesize 0.07} & {\footnotesize 0.10} & {\footnotesize 0.11} & {\footnotesize 0.11} & {\footnotesize 0.12}\tabularnewline
{\footnotesize Isometry + Noise} & {\footnotesize 0.08} & {\footnotesize 0.08} & {\footnotesize 0.08} & {\footnotesize 0.09} & {\footnotesize 0.09}\tabularnewline
\hline
\textbf{\footnotesize Average} & {\footnotesize 0.04} & {\footnotesize 0.04} & {\footnotesize 0.05} & {\footnotesize 0.06} & {\footnotesize 0.06}\tabularnewline
\hline
\end{tabular}
\end{table}

\section{3D Object Recognition Algorithm \label{sec:3D-Object-Recognition-Algorithm}}

\begin{figure*}[t]
\centering \includegraphics[scale=0.18]{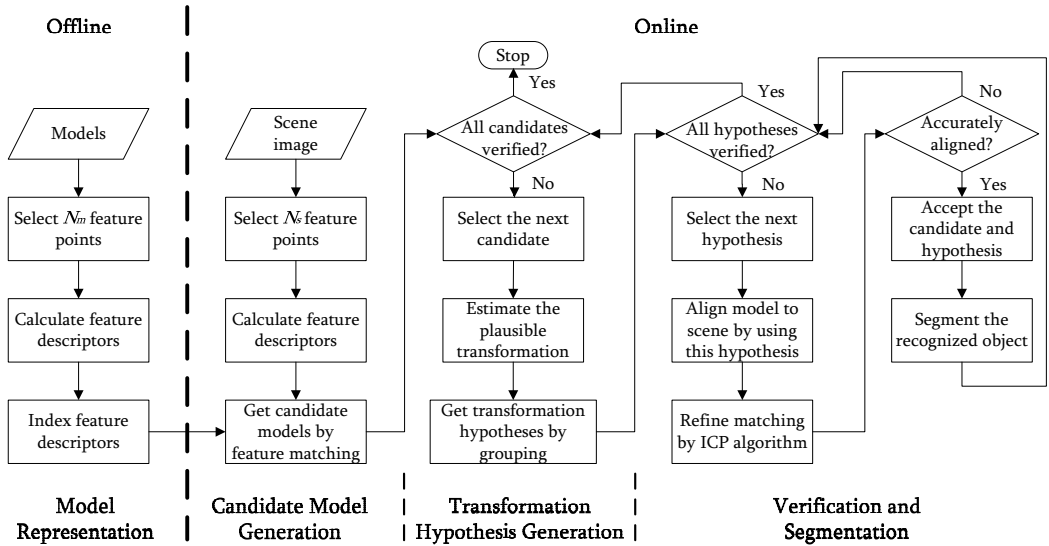}

\caption{Flow chart of the 3D object recognition algorithm. The module of model
representation is performed offline, and the other modules are operated
online. \label{fig:Flow-3D-Object-Recognition}}
\end{figure*}

So far we have developed a novel LRF and a RoPS feature descriptor.
In this section, we propose a new hierarchical 3D object recognition
algorithm based on the LRF and RoPS descriptor. Our 3D object recognition
algorithm consists of four major modules, i.e., model representation,
candidate model generation, transformation hypothesis generation,
verification and segmentation. A flow chart illustration of the algorithm
is given in Fig. \ref{fig:Flow-3D-Object-Recognition}.

\begin{table}
\caption{Robustness of MeshHOG (Gaussian curvature) with MeshDOG detector. \label{tab:Robustness-of-MeshHOG-PhotoMesh}}
\centering{}%
\begin{tabular}{cccccc}
 & \multicolumn{5}{c}{\textbf{\footnotesize Strength}}\tabularnewline
\cline{2-6}
\textbf{\footnotesize Transform.} & {\footnotesize 1} & {\footnotesize $\leq$2} & {\footnotesize $\leq$3} & {\footnotesize $\leq$4} & {\footnotesize $\leq$5}\tabularnewline
\hline
{\footnotesize Color Noise} & {\footnotesize 0.00} & {\footnotesize 0.00} & {\footnotesize 0.00} & {\footnotesize 0.00} & {\footnotesize 0.00}\tabularnewline
{\footnotesize Color Shot Noise} & {\footnotesize 0.00} & {\footnotesize 0.00} & {\footnotesize 0.00} & {\footnotesize 0.00} & {\footnotesize 0.00}\tabularnewline
{\footnotesize Geometry Noise} & {\footnotesize 0.26} & {\footnotesize 0.29} & {\footnotesize 0.31} & {\footnotesize 0.33} & {\footnotesize 0.34}\tabularnewline
{\footnotesize Geometry Shot Noise} & {\footnotesize 0.04} & {\footnotesize 0.09} & {\footnotesize 0.14} & {\footnotesize 0.21} & {\footnotesize 0.29}\tabularnewline
{\footnotesize Rotation} & {\footnotesize 0.01} & {\footnotesize 0.01} & {\footnotesize 0.01} & {\footnotesize 0.01} & {\footnotesize 0.01}\tabularnewline
{\footnotesize Scale} & {\footnotesize 0.01} & {\footnotesize 0.01} & {\footnotesize 0.01} & {\footnotesize 0.01} & {\footnotesize 0.00}\tabularnewline
{\footnotesize Local Scale} & {\footnotesize 0.21} & {\footnotesize 0.25} & {\footnotesize 0.28} & {\footnotesize 0.30} & {\footnotesize 0.31}\tabularnewline
{\footnotesize Sampling} & {\footnotesize 0.31} & {\footnotesize 0.34} & {\footnotesize 0.34} & {\footnotesize 0.36} & {\footnotesize 0.36}\tabularnewline
{\footnotesize Holes} & {\footnotesize 0.02} & {\footnotesize 0.02} & {\footnotesize 0.07} & {\footnotesize 0.07} & {\footnotesize 0.07}\tabularnewline
{\footnotesize Marco-Holes} & {\footnotesize 0.01} & {\footnotesize 0.01} & {\footnotesize 0.07} & {\footnotesize 0.07} & {\footnotesize 0.08}\tabularnewline
{\footnotesize Topology} & {\footnotesize 0.13} & {\footnotesize 0.20} & {\footnotesize 0.22} & {\footnotesize 0.25} & {\footnotesize 0.28}\tabularnewline
{\footnotesize Isometry + Noise} & {\footnotesize 0.23} & {\footnotesize 0.24} & {\footnotesize 0.22} & {\footnotesize 0.25} & {\footnotesize 0.25}\tabularnewline
\hline
\textbf{\footnotesize Average} & {\footnotesize 0.10} & {\footnotesize 0.12} & {\footnotesize 0.14} & {\footnotesize 0.15} & {\footnotesize 0.17}\tabularnewline
\hline
\end{tabular}
\end{table}

\subsection{Model Representation}

We first construct a model library for the 3D objects that we are
interested in.\textcolor{black}{{} Given a model $\mathsf{\mathscr{\mathcal{M}}}$,
$N_{m}$ seed points are evenly selected from the model pointcloud.
}Since the feature descriptors of closely located feature points may
be similar (since they represent more or less the same local surface)\textcolor{black}{,
a resolution control strategy \citep{Zhong2009} is further enforced
on these seed points to extract the final feature points.}\textcolor{magenta}{{}
}For each feature point $\boldsymbol{p}_{m}$, the LRF $\mathbf{F}_{m}$
and the feature descriptor (e.g., our RoPS descriptor) $\boldsymbol{f}_{m}$
are calculated. The point position $\boldsymbol{p}_{m}$, LRF $\mathbf{F}_{m}$
and feature descriptor $\boldsymbol{f}_{m}$ of all the feature points
are then stored in a library for object recognition.

{In order to speed up the process of feature matching
during online recognition, the local feature descriptors from all
models are indexed using a $k$-d tree method \citep{bentley1975multidimensional}.
}Note that, the model feature calculation and indexing can be performed
offline, while the following modules are operated online.

\subsection{Candidate Model Generation \label{sub:Candidate-Model-Generation}}

{The input scene $\mathcal{S}$ is first decimated,
which results in a low resolution mesh $\mathcal{S}'$. The vertices
of $\mathcal{S}$ which are nearest to the vertices of $\mathcal{S}'$
are selected as seed points (following a similar approach of \citep{Mian2006}).
Next, a resolution control strategy \citep{Zhong2009} is enforced
on these seed points to prune out redundant seed points. A boundary
checking strategy \citep{Mian2010} is also applied to the seed points
to eliminate the boundary points of the range image. Further, since
the LRF of a point can be ambiguous when two eigenvalues of the overall
scatter matrix of the underlying  local surface (see Eq. \ref{eq:overall scatter matrix})
are equal, we impose a constraint on the ratios of the eigenvalues
$\nicefrac{\lambda_{1}}{\lambda_{2}}>\tau_{\lambda}$ to exclude seed
points with symmetrical local surfaces, as in \citep{Zhong2009,Mian2010}.
The remaining seed points are considered feature points. It is worth
noting that, the feature point detection and LRF calculation procedures
can be performed simultaneously. Given the LRF $\mathbf{F}_{s}$ of
a feature point $\boldsymbol{p}_{s}$, its feature descriptor $\boldsymbol{f}_{s}$
is subsequently calculated. }

{The scene features are exactly matched against all
model features in the library using the previously constructed $k$-d
tree. }If the ratio between the smallest distance and the second smallest
one is less than a threshold $\tau_{f}$, the scene feature and its
closest model feature are considered a feature correspondence. Each
feature correspondence votes for a model. These models which have
received votes from feature correspondences are considered candidate
models. They are then ranked according to the number of votes received.
With this ranked models, the subsequent steps (Sections \ref{sub:Transformation-Hypotheses-Genera}
and \ref{sub:Verification-and-Segmentation}) can be performed from
the most likely candidate model.

\subsection{Transformation hypothesis Generation\label{sub:Transformation-Hypotheses-Genera}}

For a feature correspondence which votes for the model $\mathsf{\mathscr{\mathcal{M}}}$,
a rigid transformation is calculated by aligning the LRF of the model
feature to the LRF of the scene feature. Specifically, given the LRF
$\mathbf{F}_{s}$ and the point position $\boldsymbol{p}_{s}$ of
a scene feature, the LRF $\mathbf{F}_{m}$ and the point position
$\boldsymbol{p}_{m}$ of a corresponding model feature, the rigid
transformation can be estimated by:

\begin{equation}
\mathbf{R}=\mathbf{F}_{s}^{\mathrm{T}}\mathbf{F}_{m},
\end{equation}

\begin{equation}
\boldsymbol{t}=\boldsymbol{p}_{s}-\boldsymbol{p}_{m}\mathbf{R},
\end{equation}
where $\mathbf{R}$ is the rotation matrix and $\boldsymbol{t}$ is
the translation vector of the rigid transformation. It is worth noting
that a transformation can be estimated from a single feature correspondence
using our RoPS feature descriptor. This is a major advantage of our
algorithm compared with most of the existing algorithms (e.g., splash,
point signatures and spin image based methods) which require at least
three correspondences to calculate a transformation \citep{Johnson1999}.
Our algorithm not only eliminates the combinatorial explosion of feature
correspondences but also improves the reliability of the estimated
transformation.

{As all the plausible transformations $\left(\mathbf{R}_{i},\boldsymbol{t}_{i}\right),i=1,2,\cdots,N_{t}$
between the scene $\mathcal{S}$ and the model $\mathsf{\mathscr{\mathcal{M}}}$
are calculated, these transformations are then grouped into several
clusters. Specifically, for each plausible transformation, its rotation
matrix $\mathbf{R}_{i}$ is first converted into three Euler angles
which form a vector $\boldsymbol{u}_{i}$. In this manner, the difference
between any two rotation matrices can be measured by the Euclidean
distance between their corresponding Euler angles. These transformations
whose Euler angles are around $\boldsymbol{u}_{i}$ (with distances
less than $\tau_{a}$) and translations are around $\boldsymbol{t}_{i}$
(with distances less than $\tau_{t}$) are grouped into a cluster
$\mathcal{C}_{i}$. Therefore, each plausible transformation $\left(\mathbf{R}_{i},\boldsymbol{t}_{i}\right)$
results in a cluster $\mathcal{C}_{i}$. The cluster center $\left(\mathbf{R}_{c},\boldsymbol{t}_{c}\right)$
of $\mathcal{C}_{i}$ is calculated as the average rotation and translation
in that cluster. Next, a confidence score $s_{c}$ for each cluster
is calculated as:}

{
\begin{equation}
s_{c}=\frac{n_{f}}{d},
\end{equation}
where $n_{f}$ is the number of feature correspondences in the cluster,
and $d$ is the average distance between the scene features and their
corresponding model features which fall within the cluster. These
clusters are sorted according to their confidence scores, the ones
with confidence scores smaller than half of the maximum score are
first pruned out. We then select the valid clusters from these remaining
clusters, starting from the highest scored one and discarding the
nearby clusters whose distances to these selected clusters are small
(using $\tau_{a}$ and $\tau_{t}$). $\tau_{a}$ and $\tau_{t}$ are
empirically set to 0.2 and 30mr throughout this paper. These selected
clusters are then allowed to proceed to the final verification and
segmentation stage (Section \ref{sub:Verification-and-Segmentation}).}

\subsection{Verification and Segmentation\label{sub:Verification-and-Segmentation}}

Given a scene $\mathcal{S}$, a candidate model $\mathsf{\mathscr{\mathcal{M}}}$
and a transformation hypothesis $\left(\mathbf{R}_{c},\boldsymbol{t}_{c}\right)$,
the model $\mathsf{\mathscr{\mathcal{M}}}$ is first transformed to
the scene $\mathcal{S}$ by using the transformation hypothesis $\left(\mathbf{R}_{c},\boldsymbol{t}_{c}\right)$.
This transformation is further refined using the ICP algorithm \citep{besl1992method},
resulting in a residual error $\varepsilon$. After ICP refinement,
the visible proportion $\alpha$ is calculated as:

\begin{equation}
\alpha=\frac{n_{c}}{n_{s}},
\end{equation}
where $n_{c}$ is the number of corresponding points between the scene
$\mathcal{S}$ and the model $\mathsf{\mathscr{\mathcal{M}}}$, $n_{s}$
is the total number of points in the scene $\mathcal{S}$. Here, a
scene point and a transformed model point are considered corresponding
if their distance is less than twice the model resolution \citep{Mian2006}.

The candidate model $\mathsf{\mathscr{\mathcal{M}}}$ and the transformation
hypothesis $\left(\mathbf{R}_{c},\boldsymbol{t}_{c}\right)$ are accepted
as being correct only if the residual error $\varepsilon$ is smaller
than a threshold $\tau_{\varepsilon}$ and the proportion $\alpha$
is larger than a threshold $\tau_{\alpha}$. However, it is hard to
determine the thresholds. Because selecting strict thresholds will
reject correct hypotheses which are highly occluded in the scene,
 while selecting loose thresholds will produce many false positives.
In this paper, a flexible thresholding scheme is developed. To deal
with a highly occluded but well aligned object, we select a small
error threshold $\tau_{\varepsilon1}$ together with a small proportion
threshold $\tau_{\alpha1}$. Meanwhile, in order to increase the tolerance
to the residual error which resulted from an inaccurate estimation
of the transformation, we select a relatively larger error threshold
$\tau_{\varepsilon2}$ together with a larger proportion threshold
$\tau_{\alpha2}$. We chose these thresholds empirically and set them
as $\tau_{\varepsilon1}=0.75\textrm{mr}$, $\tau_{\varepsilon2}=1.5\textrm{mr}$,
$\tau_{\alpha1}=0.04$ and $\tau_{\alpha2}=0.2$ throughout the paper.

Therefore, once $\varepsilon<\tau_{\varepsilon1}$ but $\alpha>\tau_{\alpha1}$,
or $\varepsilon<\tau_{\varepsilon2}$ but $\alpha>\tau_{\alpha2}$,
the candidate model $\mathsf{\mathscr{\mathcal{M}}}$ and the transformation
hypothesis $\left(\mathbf{R}_{c},\boldsymbol{t}_{c}\right)$ are accepted,
the scene points which correspond to this model are removed from the
scene. Otherwise, this transformation hypothesis is rejected and the
next transformation hypothesis is verified by turn. If no transformation
hypothesis results in an accurate alignment, we conclude that the
model $\mathsf{\mathscr{\mathcal{M}}}$ is not present in the scene
$\mathcal{S}$. While if more than one transformation hypotheses are
accepted, it means that multiple instances of the model $\mathsf{\mathscr{\mathcal{M}}}$
are present in the scene $\mathcal{S}$.

Once all the transformation hypotheses for a candidate model $\mathsf{\mathscr{\mathcal{M}}}$
are tested, the object recognition algorithm then proceeds to the
next candidate model. This process continues until either all the
candidate models have been verified or there are too few points left
in the scene for recognition.

\section{Performance of 3D Object Recognition \label{sec:Experimental-ObjectRecognition}}

The effectiveness of our proposed RoPS based 3D object recognition
algorithm was evaluated by a set of experiments on four datasets,
including the Bologna Dataset \citep{tombari2010unique}, the UWA
Dataset \citep{Mian2006}, the Queen's Dataset \citep{taati2011local}
and the Ca' Foscari Venezia Dataset \citep{rodola2012scale}. These
four datasets are amongst the most popular datasets publicly available,
containing multiple objects in each scene in the presence of occlusion
and clutter.

\begin{figure*}
\begin{centering}
\subfloat[Recognition rates in the presence of noise]{\includegraphics[scale=0.45]{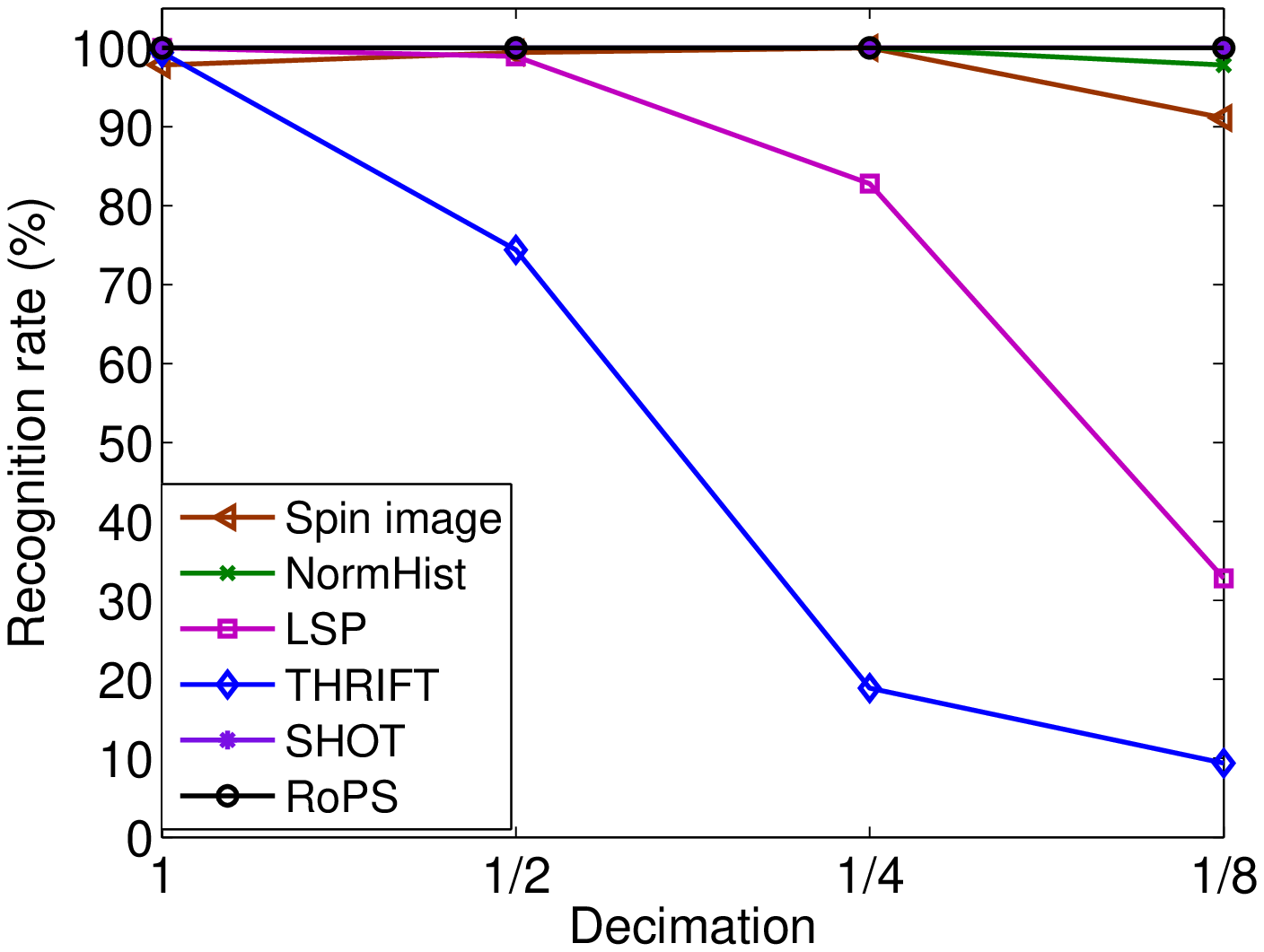}

}\,\subfloat[Recognition rates with respect to varying mesh resolution]{\includegraphics[scale=0.45]{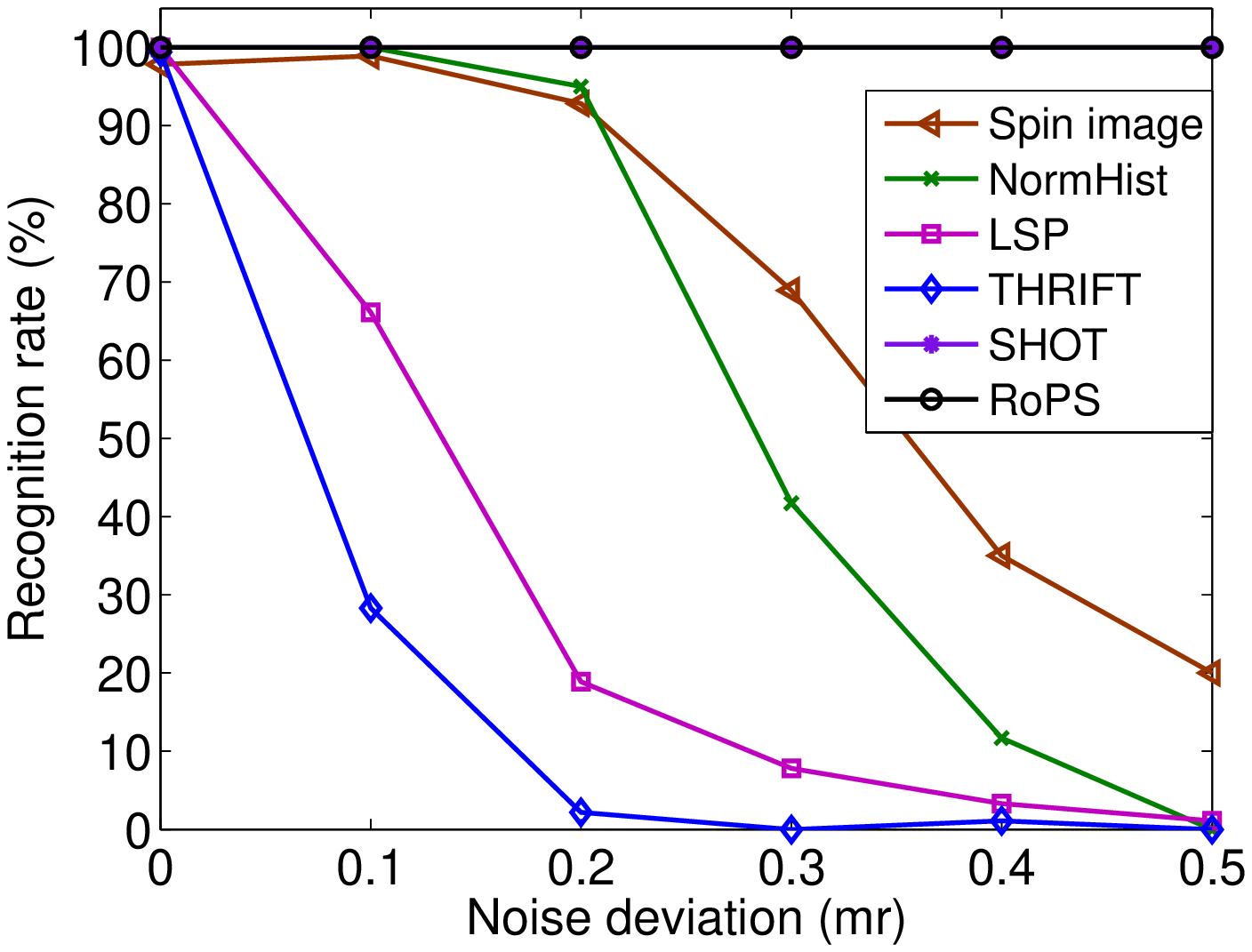}

}
\par\end{centering}

\caption{Recognition rates on the Bologna Dataset. (Figure best seen in color.)\label{fig:Recognition-rates-on-Bologna}}
\end{figure*}

\subsection{Recognition Results on The Bologna Dataset \label{sub:Recognition-Results-Bologna}}

We used the Bologna Dataset to evaluate the effectiveness of our proposed
RoPS based 3D object recognition algorithm. We specifically focused
on the performance with respect to noise and varying mesh resolution.
We also aimed to demonstrate the capability of our 3D object recognition
algorithm to integrate the existing feature descriptors without LRF.

\begin{figure*}
 \begin{centering}
\subfloat[Chef]{\includegraphics[scale=0.23]{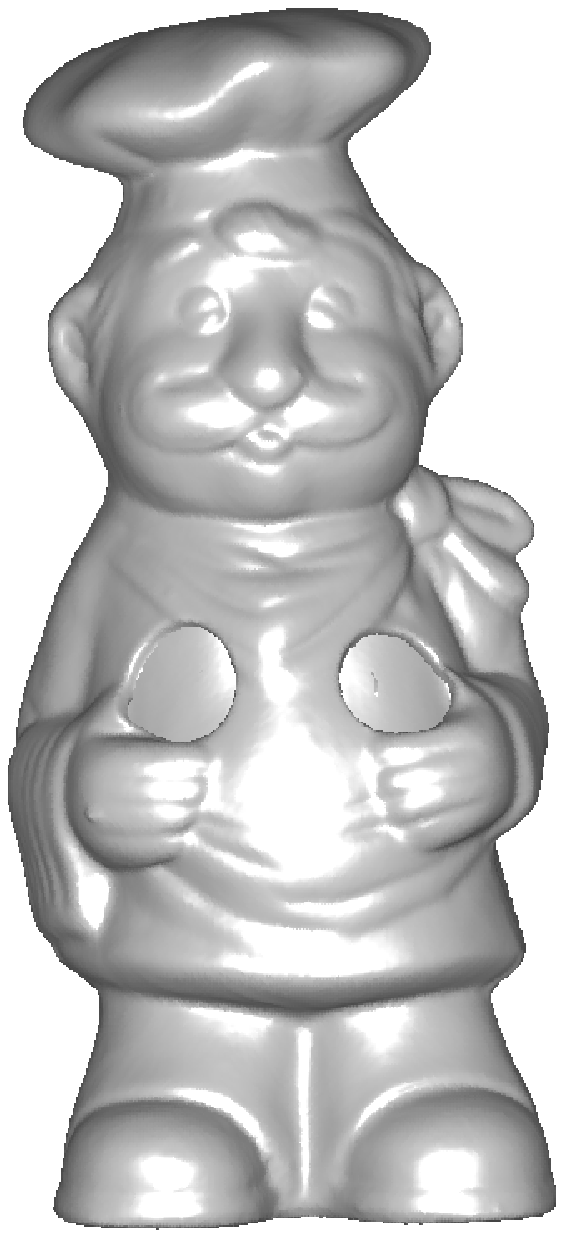}

}\,\subfloat[Chicken]{\includegraphics[scale=0.23]{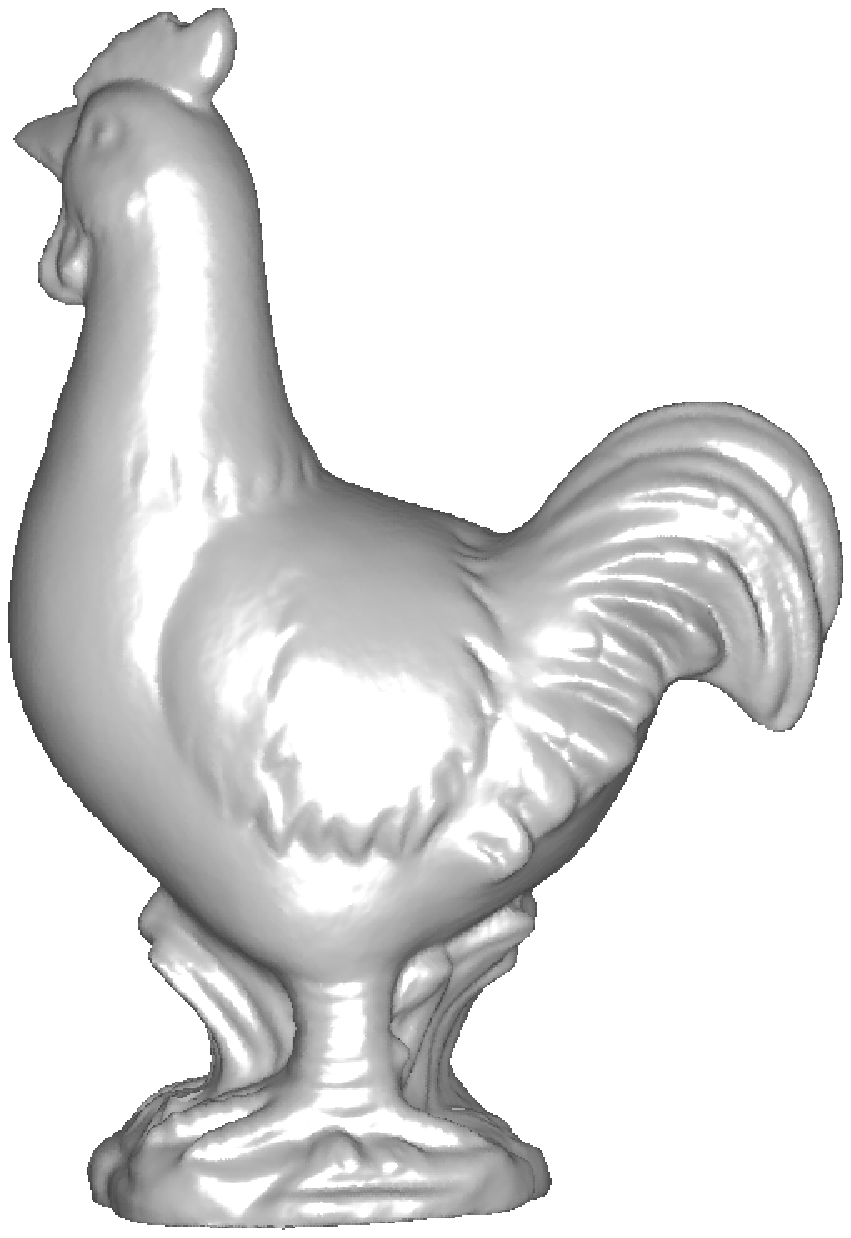}

}\,\subfloat[Parasaurolophus]{\includegraphics[scale=0.23]{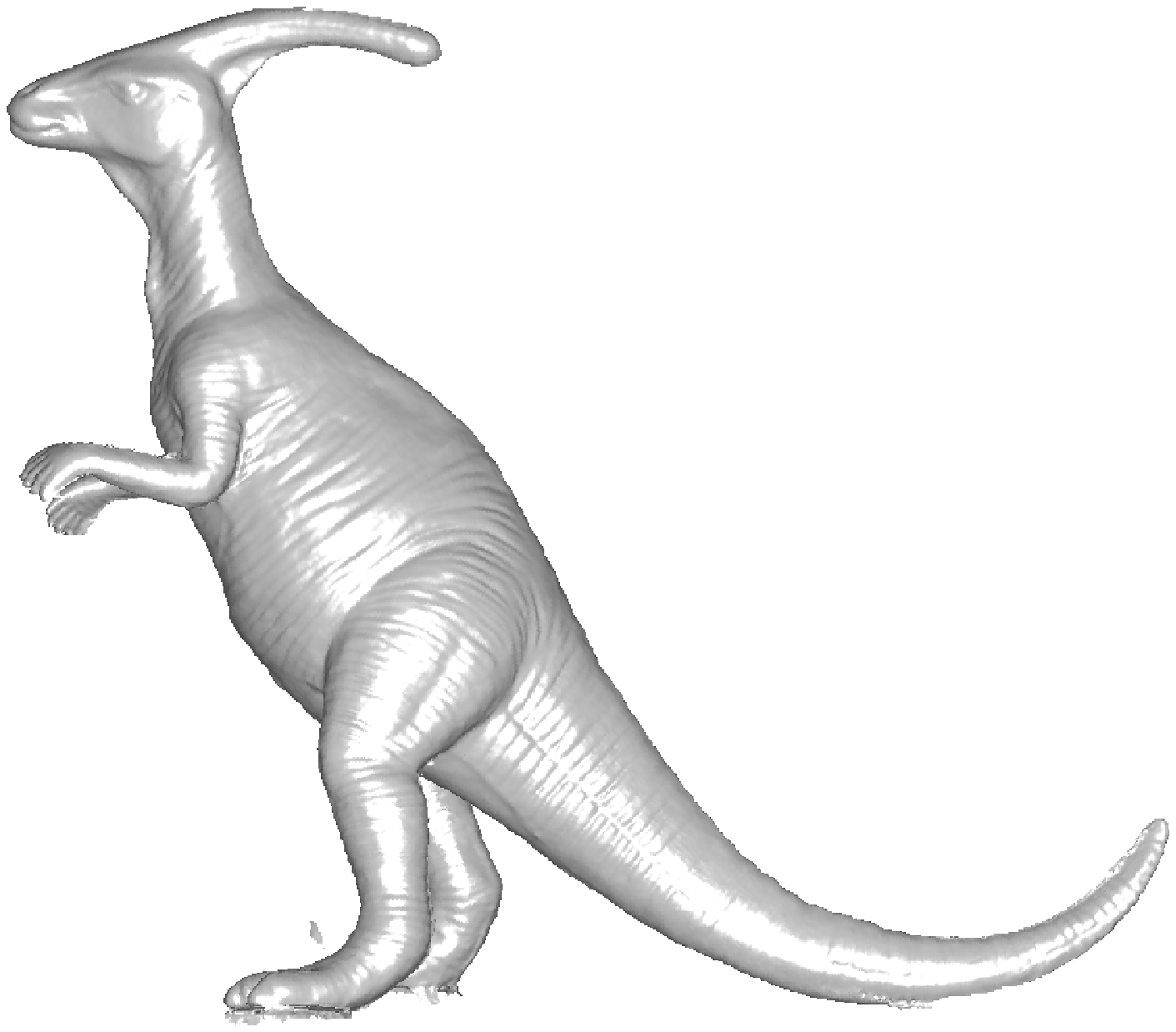}

}\,\subfloat[Rhino]{\includegraphics[scale=0.23]{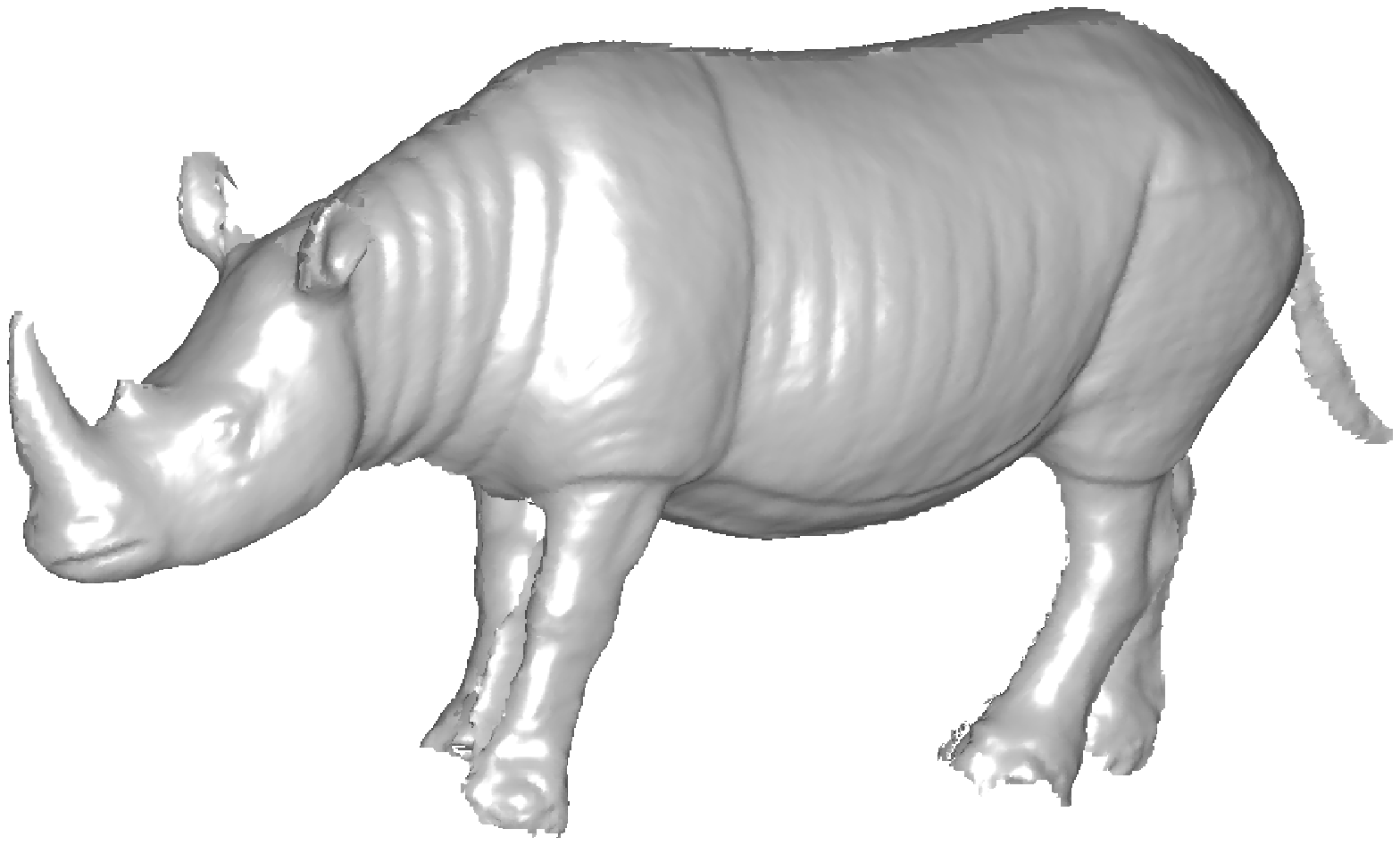}

}\,\subfloat[T-Rex]{\includegraphics[scale=0.23]{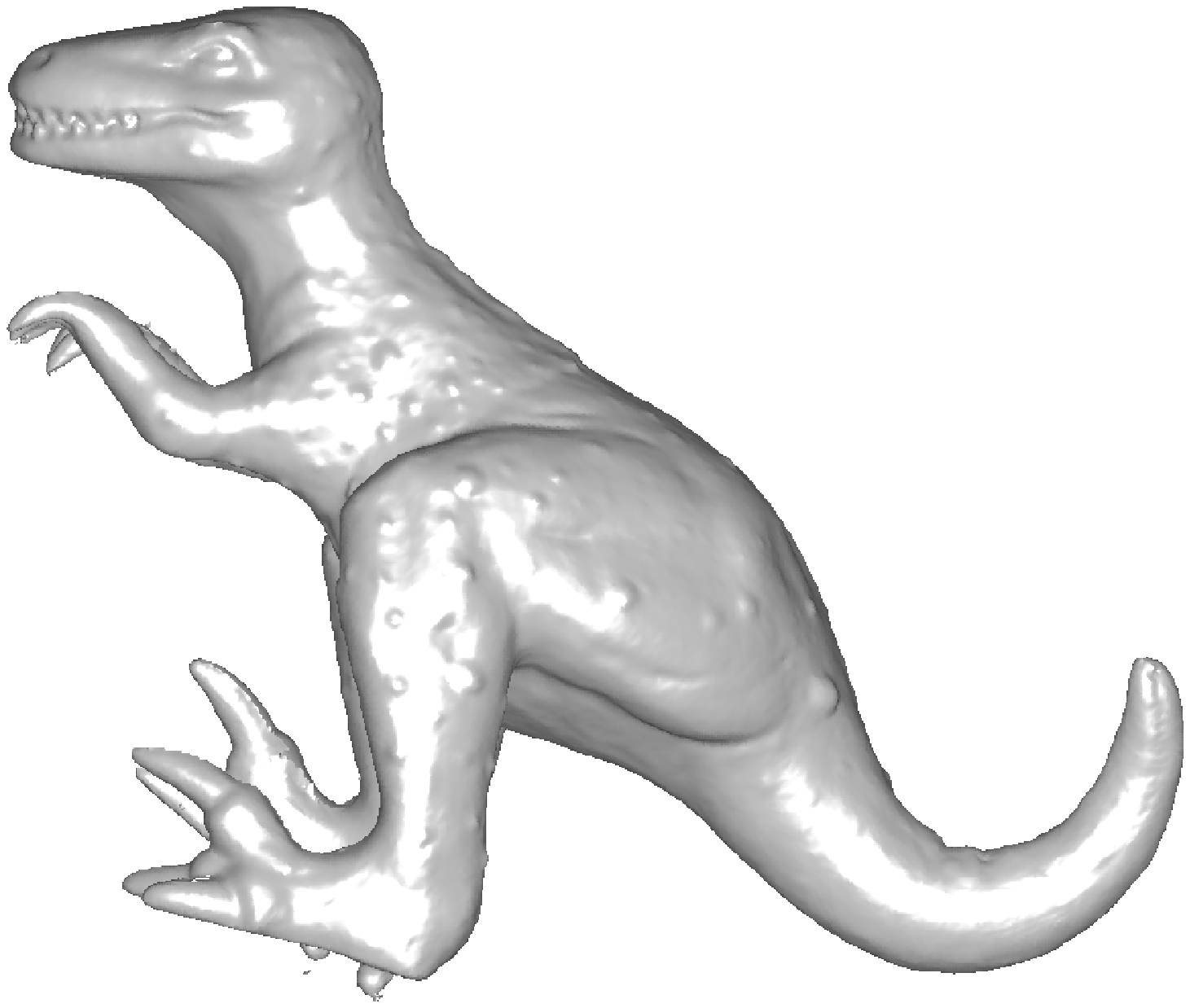}

}
\par\end{centering}

\caption{The five models of the UWA Dataset. \label{fig:The-5-models-UWA}}
\end{figure*}

\begin{figure*}
\begin{centering}
\subfloat[The first sample scene]{\includegraphics[scale=0.32]{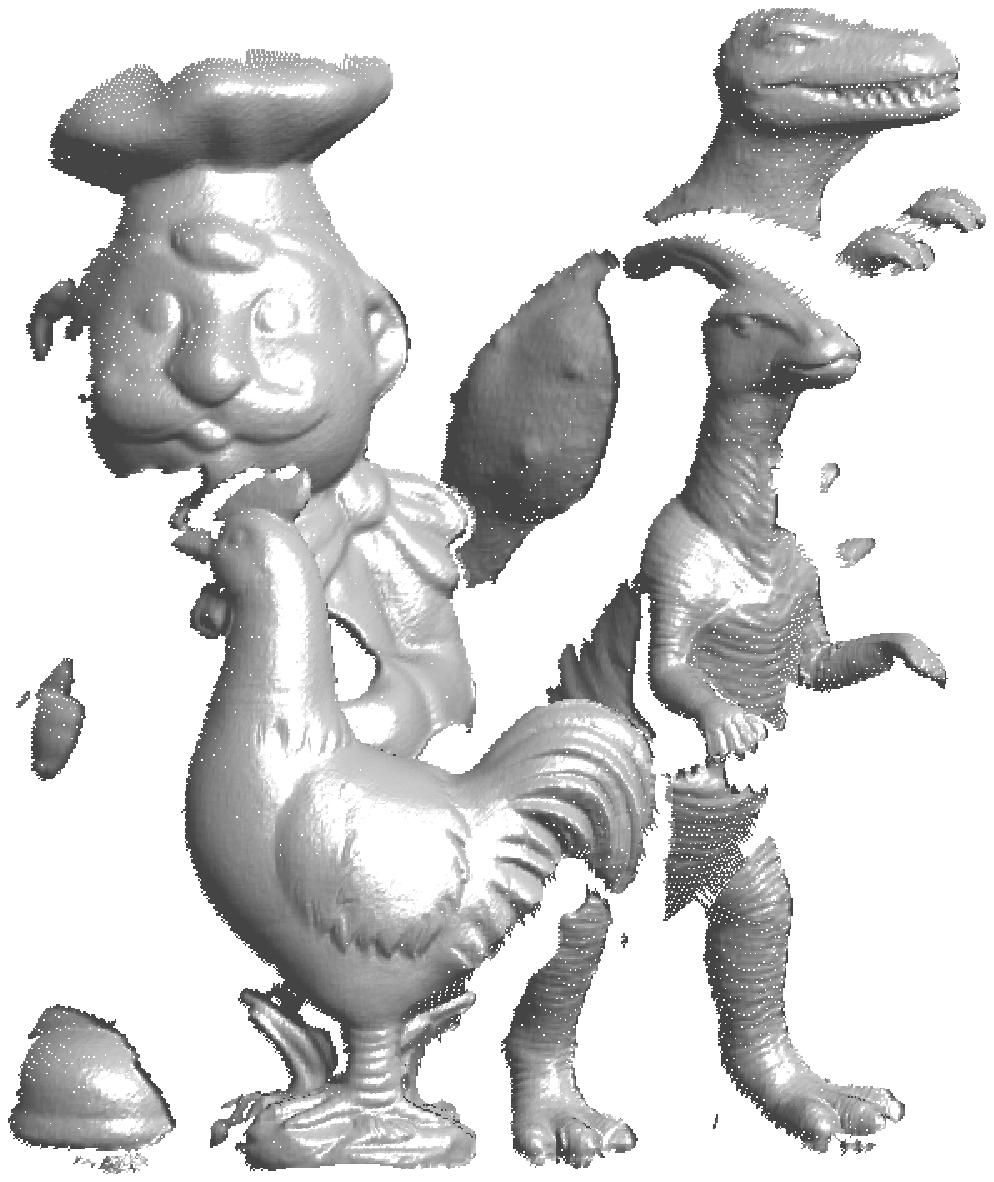}

}\,\subfloat[Our recognition result]{\includegraphics[scale=0.32]{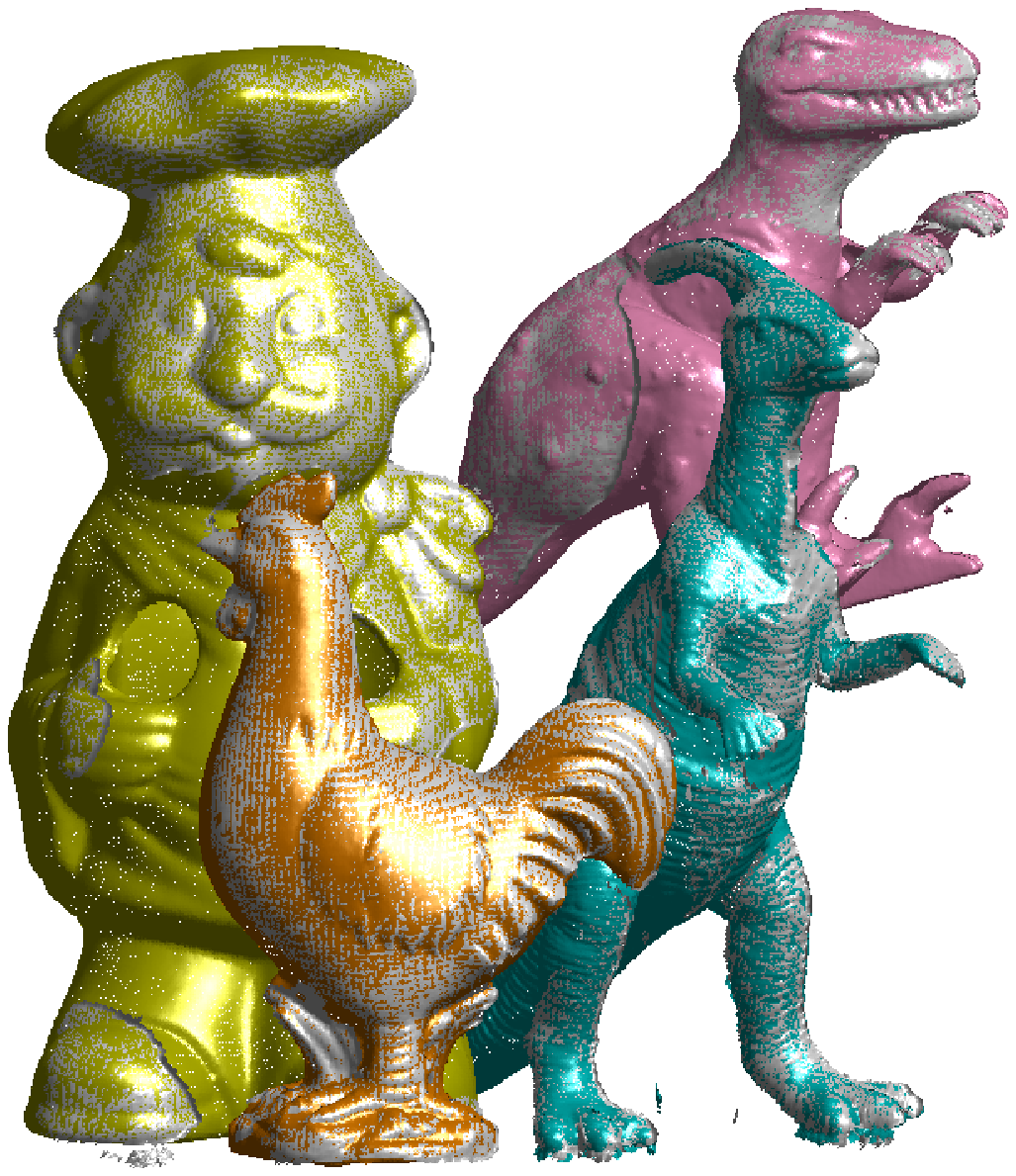}

}\,\subfloat[The second sample scene]{\includegraphics[scale=0.32]{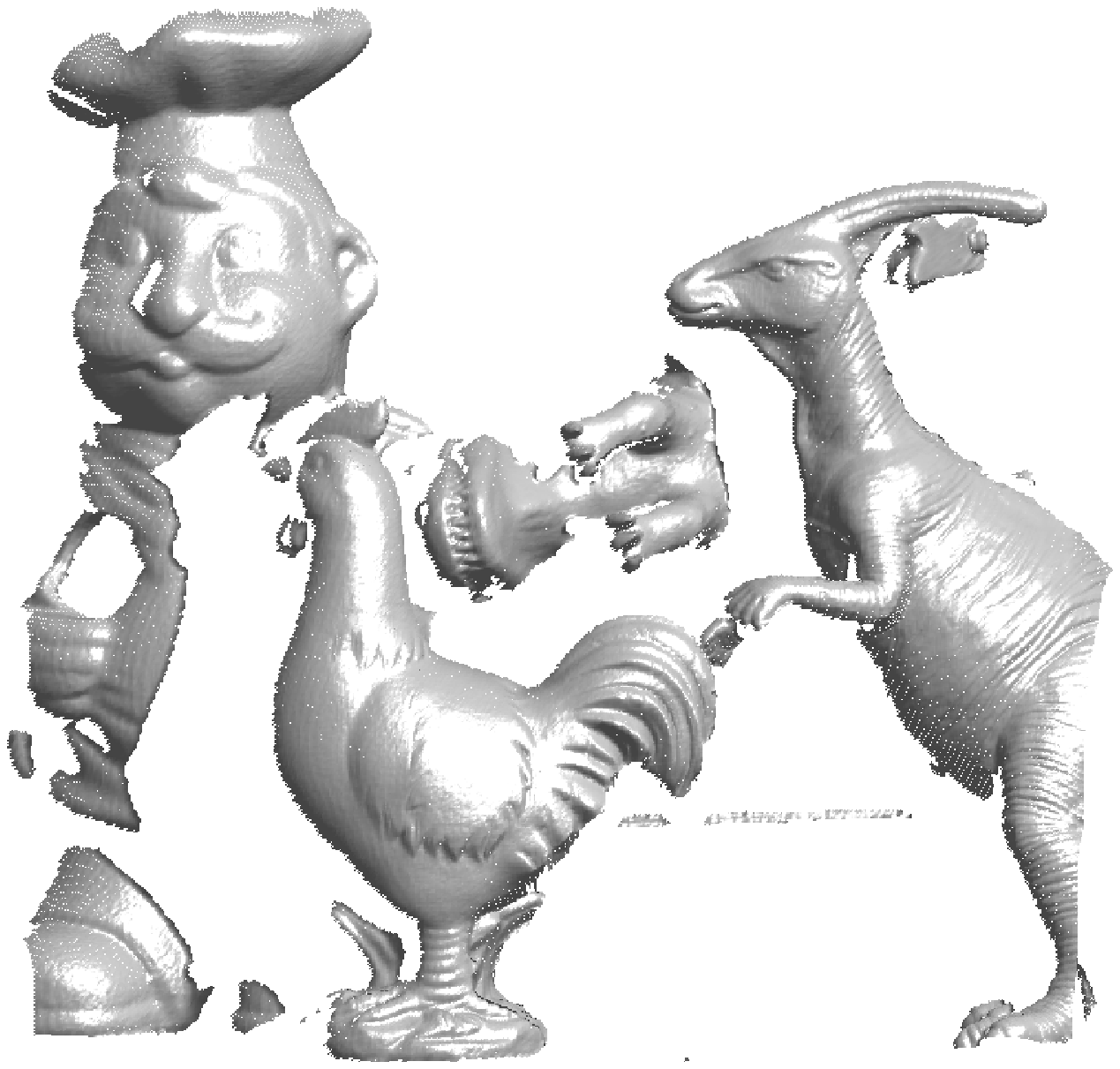}

}\,\subfloat[Our recognition result]{\includegraphics[scale=0.3]{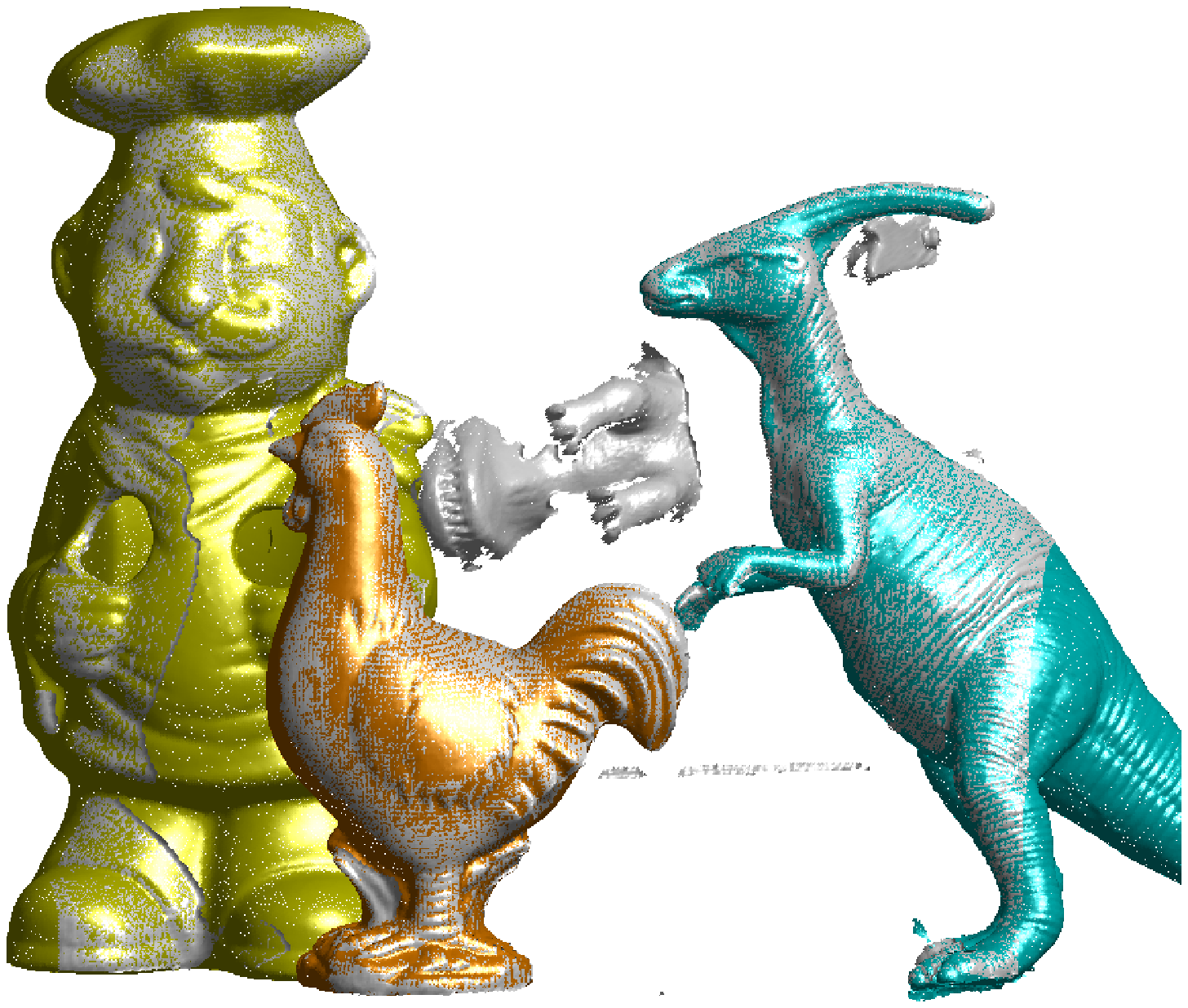}

}
\par\end{centering}

\caption{Two sample scenes and our recognition results on the UWA Dataset.
The correctly recognized objects have been superimposed by their 3D
complete models from the library. All objects were correctly recognized
except for the T-Rex in (d). (Figure best seen in color.) \label{fig:recognition-result-UWA}}
\end{figure*}

We used our RoPS together with the five feature descriptors (as detailed
in Section \ref{sub:Dataset-and-Parameter}) to perform object recognition.
For feature descriptors that do not have a dedicated LRF, e.g., spin
image, NormHist, LSP and THRIFT, the LRFs were defined using our proposed
technique. \textcolor{black}{The average number of detected feature
points in an unsampled scene and a model were 985 and 1000, respectively.}

In order to evaluate the performance of the 3D object recognition
algorithms on noisy data, we added a Gaussian noise with increasing
standard deviation of 0.1mr, 0.2mr, 0.3mr, 0.4mr and 0.5mr to each
scene data, the average recognition rates of the six algorithms on
the 45 scenes are shown in Fig. \ref{fig:Recognition-rates-on-Bologna}(a).
It can be seen that both RoPS and SHOT based algorithms achieved the
best results, with recognition rates of 100\% under all levels of
noise. Spin image and NormHist based algorithms achieved recognition
rates higher than 97\% under low-level noise with deviations less
than 0.1mr. However, their performance deteriorated sharply as the
noise increased. While LSP and THRIFT based algorithms were very sensitive
to noise.

In order to evaluate the effectiveness of the 3D object recognition
algorithms with respect to varying mesh resolution, the 45 noise free
scenes were resampled to $\nicefrac{1}{2}$, $\nicefrac{1}{4}$ and
$\nicefrac{1}{8}$ of their original mesh resolution. The average
recognition rates on the 45 scenes with respect to different mesh
resolutions are given in Fig. \ref{fig:Recognition-rates-on-Bologna}(b).
It is shown that RoPS based algorithm achieved the best performance,
obtaining 100\% recognition rate under all levels of mesh decimation.
It was followed by NormHist and spin image based algorithms. That
is, they obtained recognition rates of 97.8\% and 91.1\% respectively
in scenes with $\nicefrac{1}{8}$ of original mesh resolution.

\subsection{Recognition Results on The UWA Dataset }

The UWA Dataset contains five 3D models and 50 real scenes. The scenes
were generated by randomly placing four or five real objects together
in a scene and scanned from a single viewpoint using a Minolta Vivid
910 scanner. An illustration of the five models is given in Fig. \ref{fig:The-5-models-UWA},
and two sample scenes are shown in Figures \ref{fig:recognition-result-UWA}(a)
and (c).

\begin{figure}[h]
\begin{centering}
\includegraphics[scale=0.5]{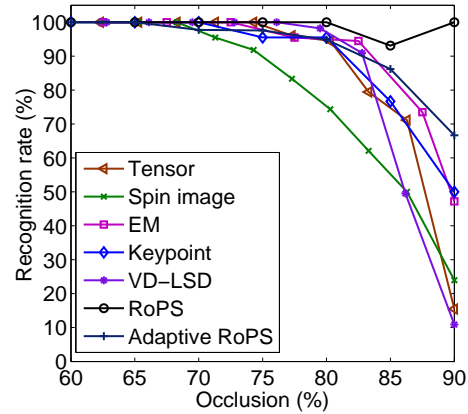}
\par\end{centering}

\caption{Recognition rates on the UWA Dataset. (Figure best seen in color.)\label{fig:Recognition-rate-UWA}}
\end{figure}

For the sake of consistency in comparison, RoPS based 3D object recognition
experiments were performed on the same data as \citet{Mian2006} and
\citet{bariya20123d}. Besides, the Rhino model was excluded from
the recognition results, since it contained large holes and cannot
be recognized by the spin image based algorithm in any of the scenes.
Comparison was performed with a number of state-of-the-art algorithms,
such as tensor \citep{Mian2006}, spin image \citep{Mian2006}, keypoint
\citep{Mian2010}, VD-LSD \citep{taati2011local} and EM based \citep{bariya20123d}
algorithms. Comparison results are shown in Fig. \ref{fig:Recognition-rate-UWA}
with respect to varying levels of occlusion. \textcolor{black}{The
average number of detected feature points in a scene and a model were
2259 and 4247, respectively.}

Occlusion is defined according to \citet{Johnson1999} as:

\begin{equation}
\textrm{occlusion}=\frac{\textrm{model surface patch area in scene}}{\textrm{total model surface area}}.
\end{equation}

The ground truth occlusion values were automatically calculated for
the correctly recognized objects and manually calculated for the objects
which were not correctly recognized. As shown in Fig. \ref{fig:Recognition-rate-UWA},
our RoPS based algorithm outperformed all the existing algorithms.
It achieved a recognition rate of 100\% with up to 80\% occlusion,
and a recognition rate of 93.1\% even under 85\% occlusion. The average
recognition rate of our RoPS based algorithm was \textcolor{black}{98.8\%,}
while the average recognition rate of spin image, tensor and EM based
algorithms were 87.8\%, 96.6\% and 97.5\% respectively, with up to
84\% occlusion. {The overall average recognition rate
of our RoPS based algorithm was 98.9\%. }Moreover, no false positive
occurred in the experiments when using our RoPS based algorithm, and
only two out of the total 188 objects in the 50 scenes was not correctly
recognized. These results confirm that our RoPS based algorithm is
able to recognize objects in complex scenes in the presence of significant
clutter, occlusion and mesh resolution variation.

Two sample scenes and their corresponding recognition results are
shown in Fig. \ref{fig:recognition-result-UWA}. All objects were
correctly recognized and their poses were accurately recovered except
for the T-Rex in Fig. \ref{fig:recognition-result-UWA}(d). The reason
for the failure in Fig. \ref{fig:recognition-result-UWA}(d) relates
to the excessive occlusion of the T-Rex.\textcolor{black}{{} It is highly
occluded }and the visible surface is sparsely distributed in several
parts of the body rather than in a single area. Therefore, almost
no reliable feature could be extracted from the object.

\begin{figure*}
 \begin{centering}
\subfloat[Angle]{\includegraphics[scale=0.20]{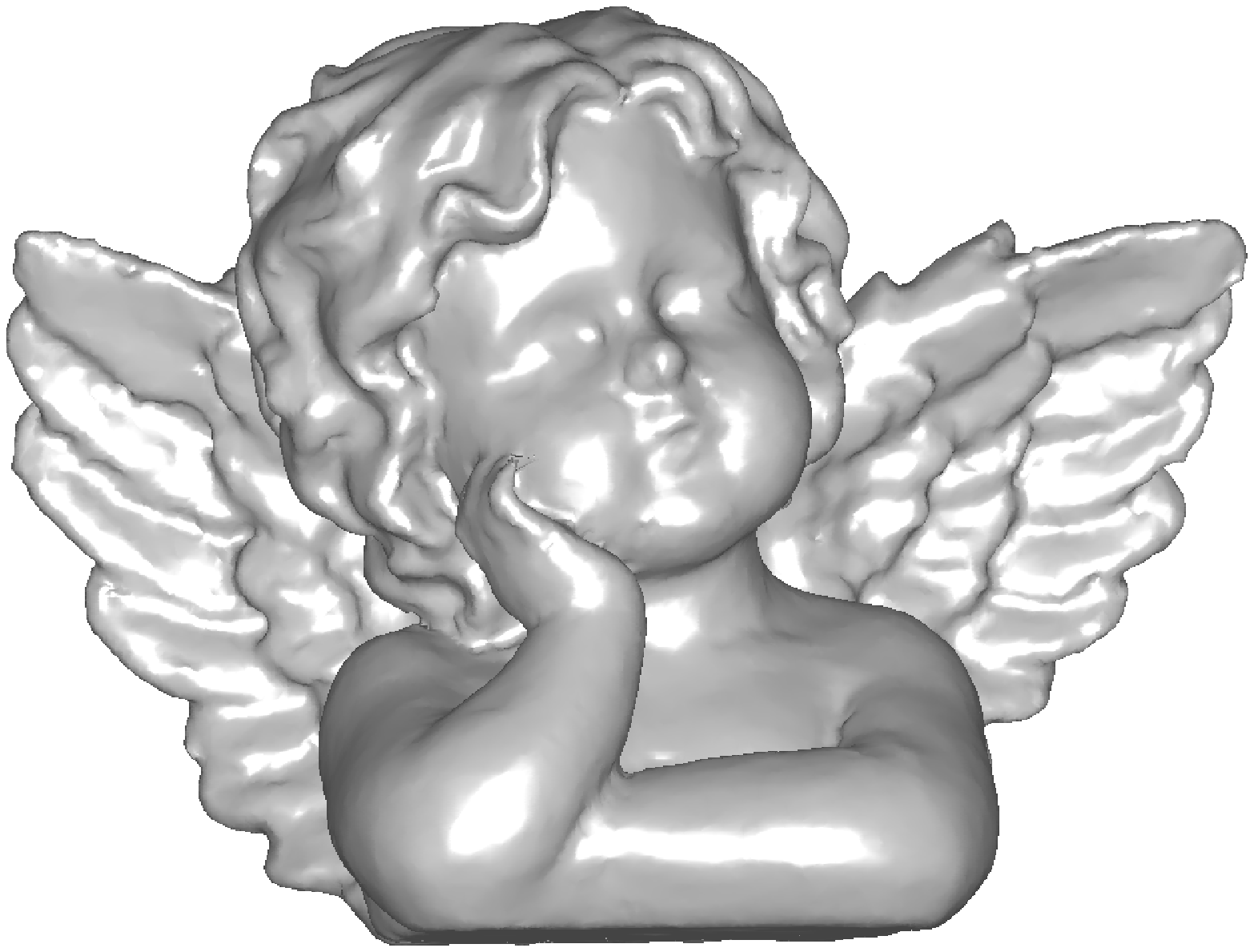}

}\,\subfloat[Big Bird]{\includegraphics[scale=0.23]{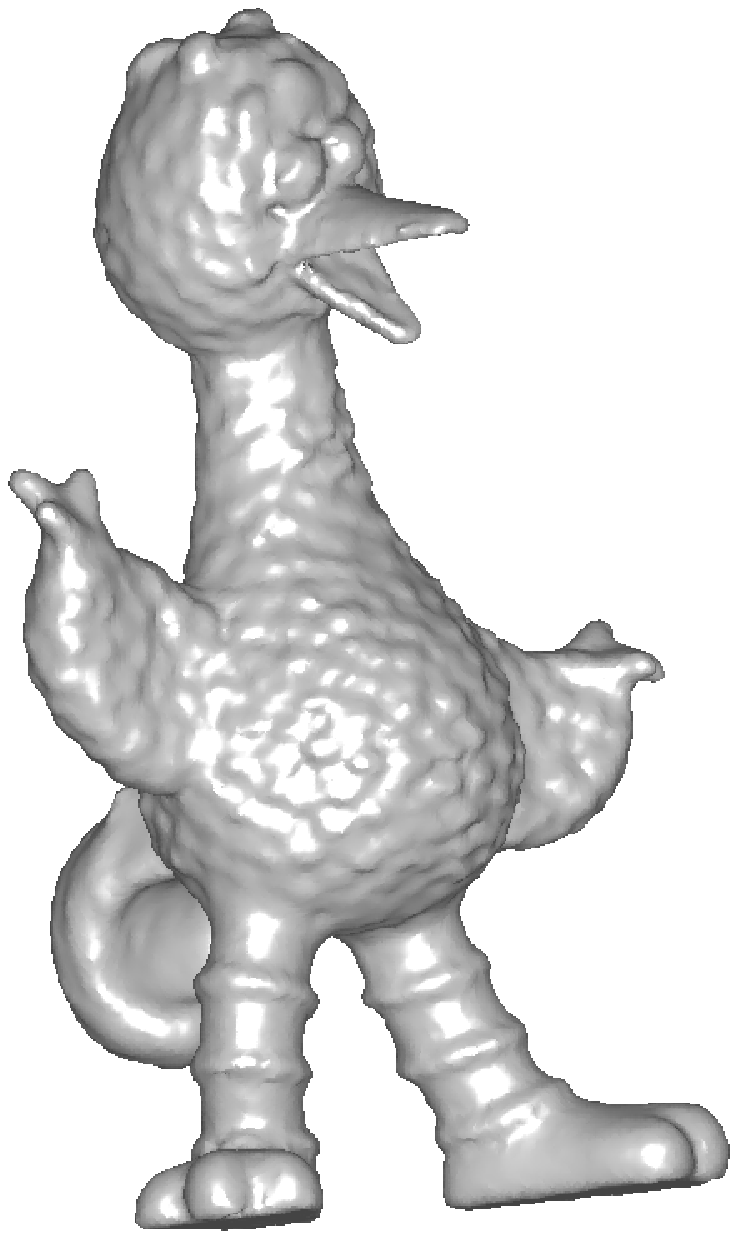}

}\,\subfloat[Gnome]{\includegraphics[scale=0.23]{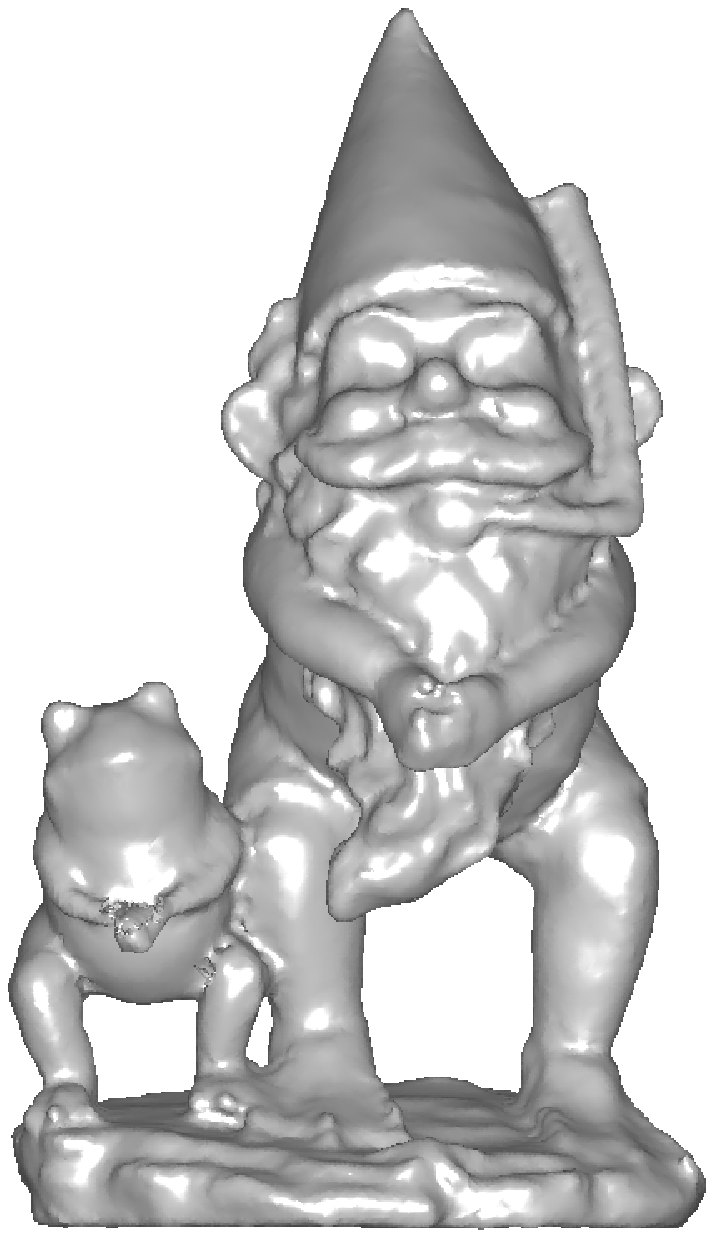}

}\,\subfloat[Kid]{\includegraphics[scale=0.23]{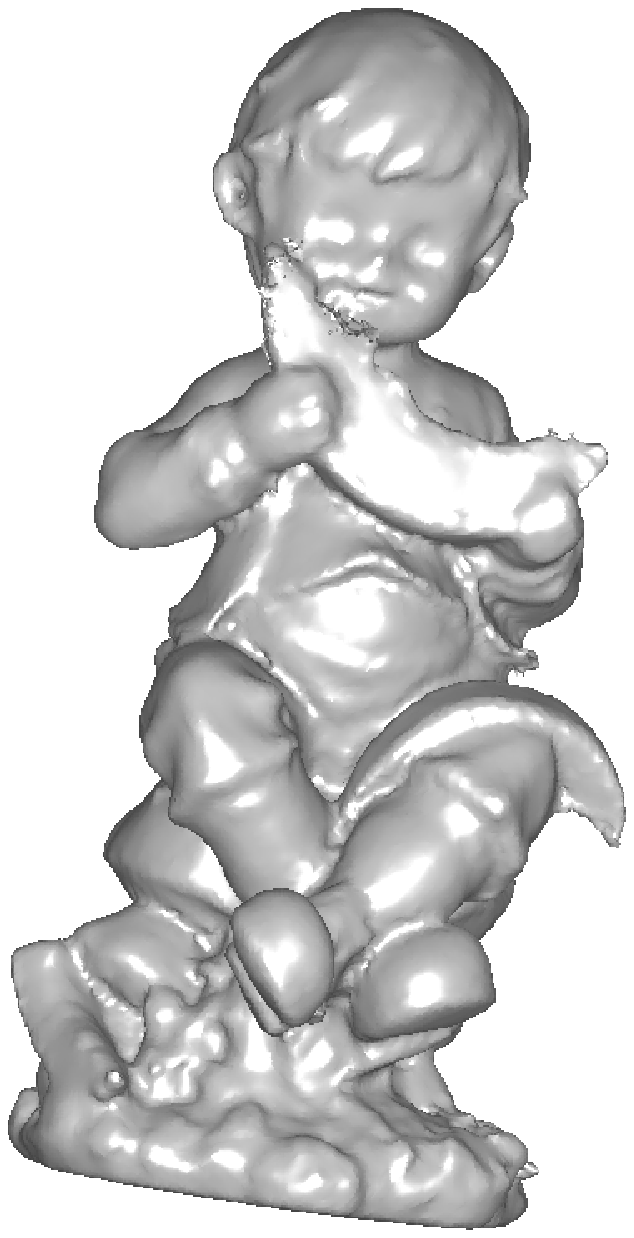}

}\,\subfloat[Zoe]{\includegraphics[scale=0.23]{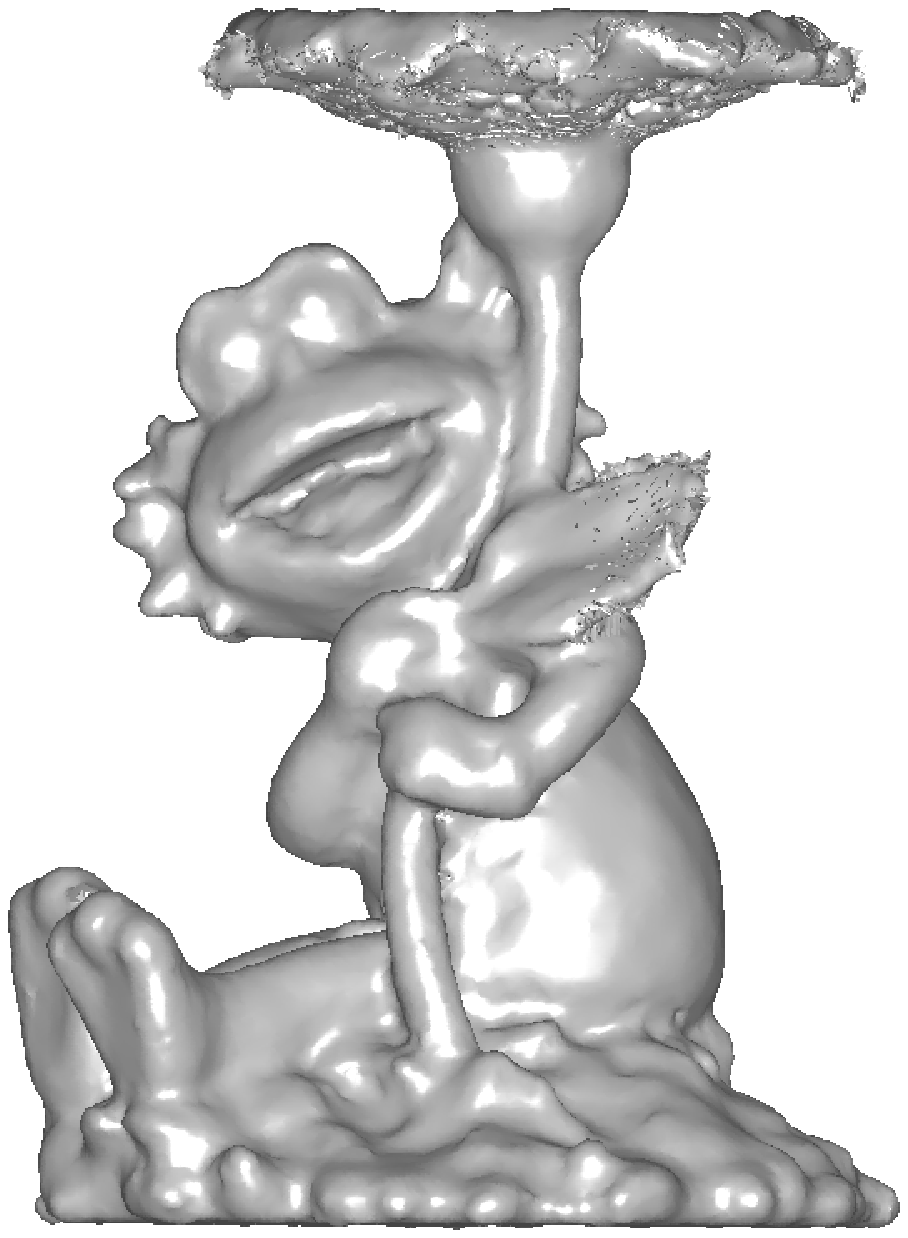}

}
\par\end{centering}

\caption{The five models in the Queen's Dataset. \label{fig:The-5-models-Queen}}
\end{figure*}

\begin{figure*}
\begin{centering}
\subfloat[The first sample scene]{\includegraphics[scale=0.27]{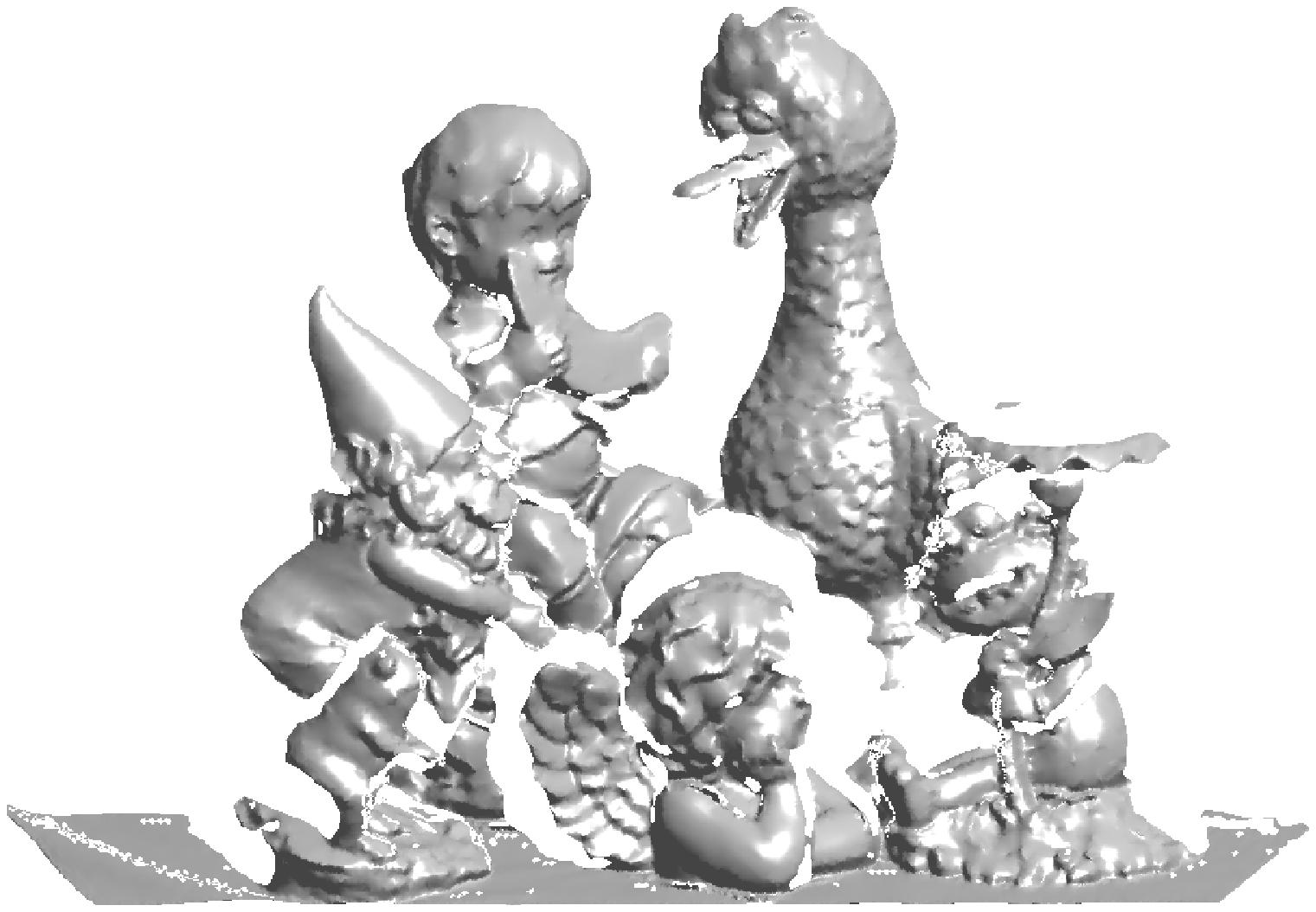}

}\,\subfloat[Our recognition result]{\includegraphics[scale=0.27]{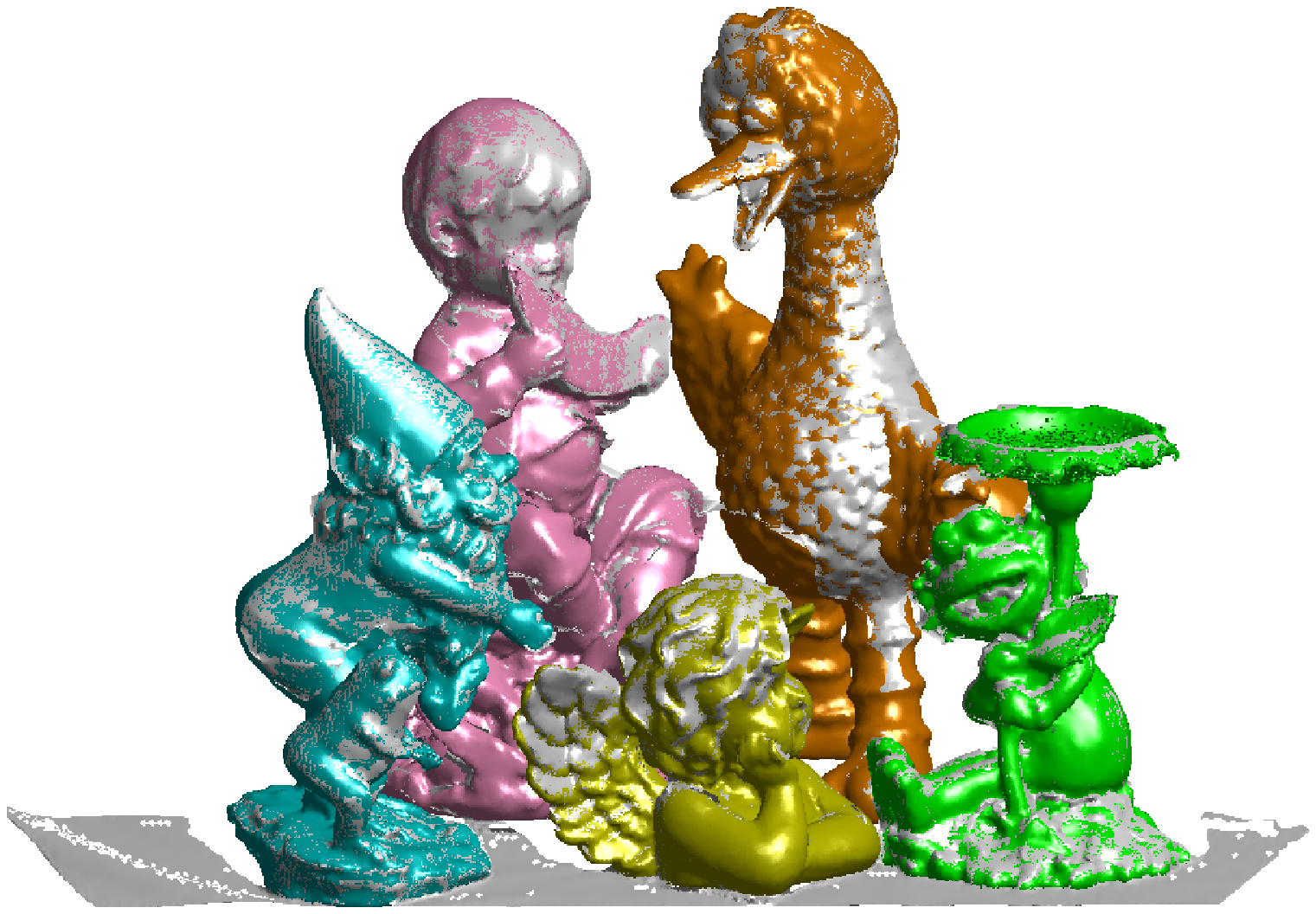}

}\,\subfloat[The second sample scene]{\includegraphics[scale=0.27]{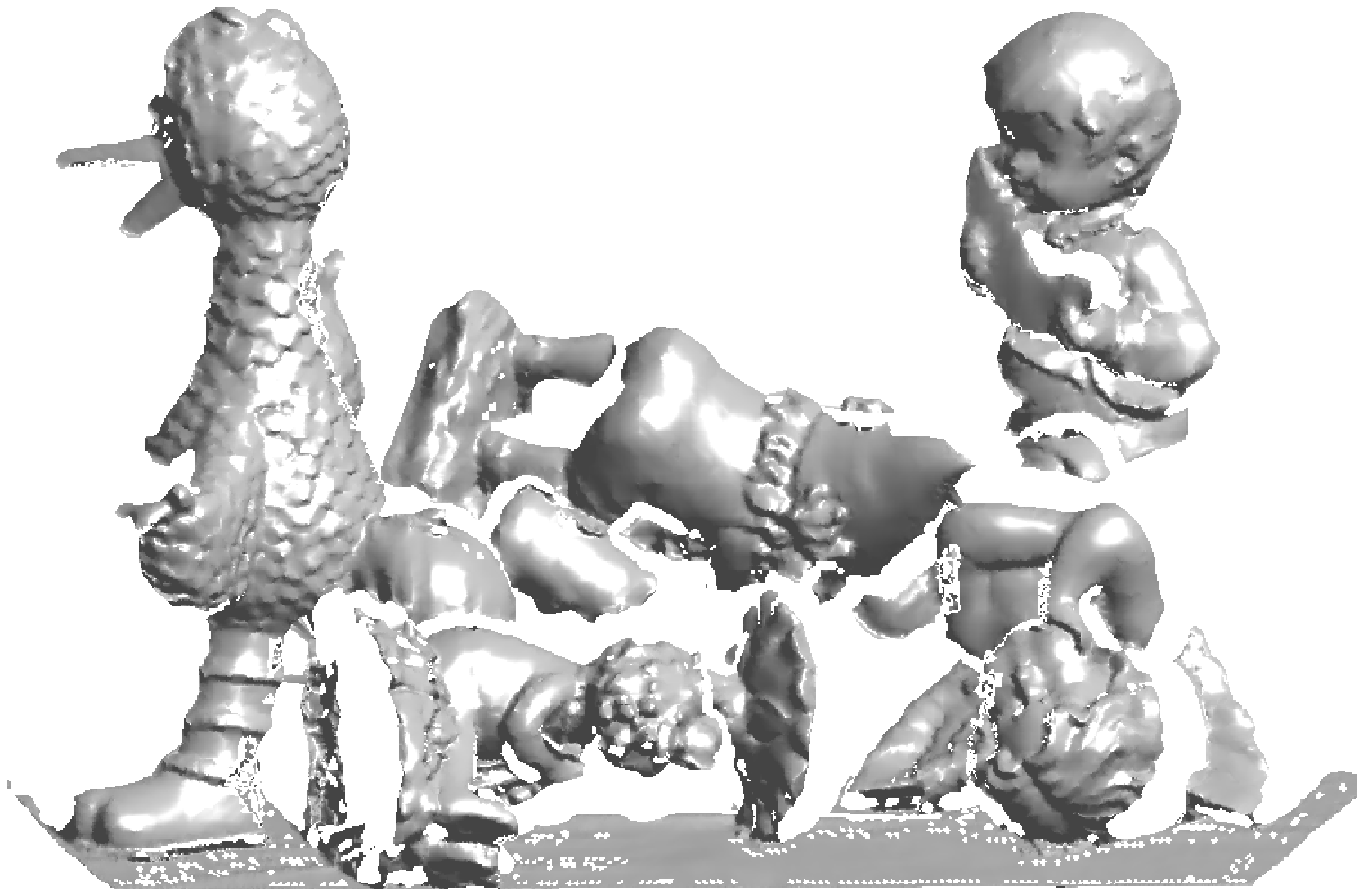}

}\,\subfloat[Our recognition result]{\includegraphics[scale=0.27]{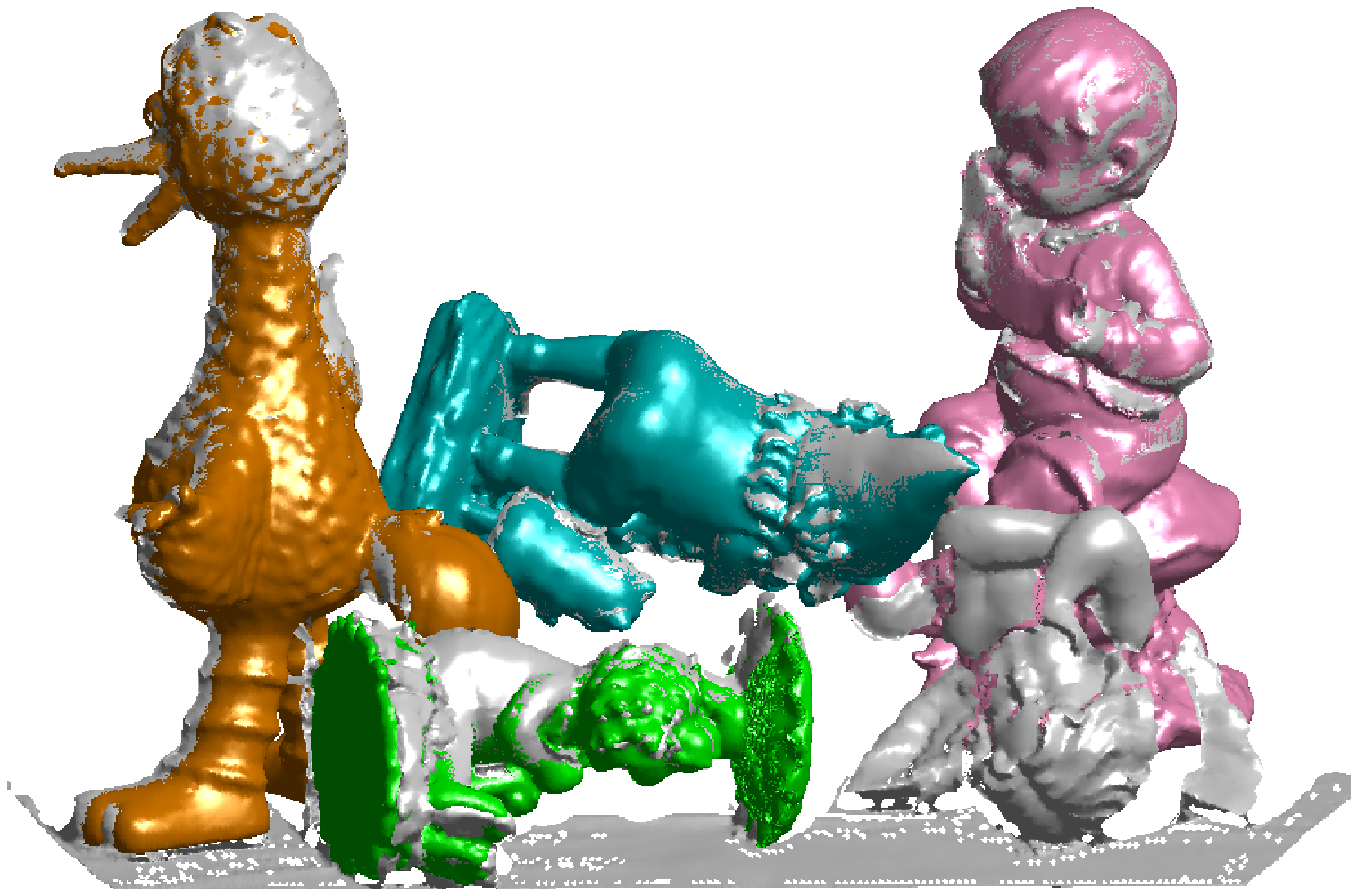}

}
\par\end{centering}

\caption{Two sample scenes and our recognition results on the Queen's dataset.
The correctly recognized objects have been superimposed by their 3D
complete models from the library. All objects were correctly recognized
except for the Angle in (d). (Figure best seen in color.) \label{fig:recognition-result-Queen}}
\end{figure*}

{Note that, although we used a fixed support radius
(i.e., $r$ = 15mr) for feature description throughout this paper,
the proposed algorithm is generic, and different adaptive-scale keypoint
detection methods can be seamlessly integrated within our RoPS descriptor.
In order to further demonstrate the generic nature  of our algorithm,
we generated RoPS descriptors using the support radii estimated by
the adaptive-scale method in \citep{Mian2010}. The recognition result
is shown in Fig. \ref{fig:Recognition-rate-UWA}. The recognition
performance of the adaptive-scale RoPS based algorithm was better
than that reported in \citep{Mian2010}, which means that our RoPS
descriptor was more descriptive than the descriptor used in \citep{Mian2010}.
It is also observed that the performance of adaptive-scale RoPS was
marginally worse than the fixed-scale counterpart. This is because
the errors of scale estimation adversely affected the performance
of feature matching, and ultimately object recognition. That is, the
corresponding points in a scene and model may have different estimated
scales due to the estimation errors. As reported in \citep{tombari2012performance},
the scale repeatability of the adaptive-scale detector in \citep{Mian2010}
were less than 85\% and 60\% on the Retrieval dataset and Random Views
dataset, respectively. }

\subsection{Recognition Results on The Queen's Dataset}

The Queen's Dataset contains five models and 80 real scenes. The 80
scenes were generated by randomly placing one, three, four or five
of the models in a scene and scanned from a single viewpoint using
a LIDAR sensor. The five models were generated by merging several
range images of a single object. Since all scenes and models were
represented in the form of pointclouds, we first converted them into
triangular meshes in order to calculate the LRFs using our proposed
technique. A scene pointcloud was converted by mapping the 3D pointcloud
onto the 2D retina plane of the sensor and performing a 2D Delaunay
triangulation over the mapped points. The 2D points and triangles
were then mapped back to the 3D space, resulting in a triangular mesh.
A model pointcloud was converted into a triangular mesh using the
Marching Cubes algorithm \citep{guennebaud2007algebraic}. An illustration
of the five models is given in Fig. \ref{fig:The-5-models-Queen},
and two sample scenes are shown in Figures \ref{fig:recognition-result-Queen}(a)
and (c).

\begin{table*}
\caption{Recognition rates (\%) on the Queen's Dataset. The results of the
tests on the full dataset containing 80 scenes are shown in parentheses.
The others were tested on a subset dataset which contains 55 scenes.
`NA' indicates that the corresponding item is not available. The
best results are in bold fonts. \label{tab:RecognitionRate-Queen}}

\centering{}%
\begin{tabular}{ccccccc}
\hline
Method & Angel & Big Bird & Gnome & Kid & Zoe & Average\tabularnewline
\hline
{RoPS} & \textbf{{97.4}}{{} (}\textbf{{97.9}}{)} & \textbf{{100.0}}{{} (}\textbf{{100.0}}{)} & \textbf{{97.4}}{{} (}\textbf{{97.9}}{)} & \textbf{{94.9}}{{} (}\textbf{{95.8}}{)} & \textbf{{87.2}}{{} (}\textbf{{85.4}}{)} & \textbf{{95.4}}{{} (}\textbf{{95.4}}{)}\tabularnewline
EM & NA (77.1) & NA (87.5) & NA (87.5) & NA (83.3) & NA (76.6) & 81.9 (82.4)\tabularnewline
VD-LSD(SQ) & 89.7 & 100.0 & 70.5 & 84.6 & 71.8 & 83.8\tabularnewline
VD-LSD(VQ) & 56.4 & 97.4 & 69.2 & 51.3 & 64.1 & 67.7\tabularnewline
3DSC & 53.8 & 84.6 & 61.5 & 53.8 & 56.4 & 62.1\tabularnewline
Spin image (impr.) & 53.8 & 84.6 & 38.5 & 51.3 & 41.0 & 53.8\tabularnewline
Spin image (orig.) & 15.4 & 64.1 & 25.6 & 43.6 & 28.2 & 35.4\tabularnewline
Spin image spherical (impr.) & 53.8 & 74.4 & 38.5 & 61.5 & 43.6 & 54.4\tabularnewline
Spin image spherical (orig.) & 12.8 & 61.5 & 30.8 & 43.6 & 30.8 & 35.9\tabularnewline
\hline
\end{tabular}
\end{table*}

First, we performed object recognition using our RoPS based algorithm
on the full dataset which contains 80 real scenes. \textcolor{black}{The
average number of detected feature points in a scene and a model were
3296 and 4993, respectively. }The results are shown in parentheses
in Table \ref{tab:RecognitionRate-Queen},\textcolor{red}{{} }\textcolor{black}{with
a comparison to the results given by \citet{bariya20123d}. It can
be seen that the average recognition rate of our algorithm is 95.4\%,
in contrast, the average recognition rate of the EM based algorithm
is 82.4\%. }These results indicate that our algorithm is superior
to the EM based algorithm although a complicated keypoint detection
and scale selection strategy has been adopted by the EM based algorithm.

To make a direct comparison with the results given by \citet{taati2011local},
we performed our RoPS based 3D object recognition on the same subset
dataset which contains 55 scenes. The results are given in Table \ref{tab:RecognitionRate-Queen},\textcolor{red}{{}
}with comparisons to the results provided by two variants of VD-LSD,
3DSC and four variants of spin image. As shown in Table. \ref{tab:RecognitionRate-Queen},{{}
our average recognition rate was 95.4\%, while the second best result
achieved by VD-LSD (SQ) was 83.8\%. The RoPS based algorithm achieved
the best recognition rates for all the five models. More than 97\%
of the instances of Angle, Big Bird and Gnome were correctly recognized.
Although RoPS's recognition rate for Zoe was relatively low (i.e.,
87.2\%),}\textcolor{red}{{} }it still outperformed the existing algorithms
by a large margin, since the second best result achieved by VD-LSD
(SQ) was 71.8\%. Fig. \ref{fig:recognition-result-Queen} shows two
sample scenes and our recognition results on the Queen's Dataset.
It can be seen that our RoPS based algorithm was able to recognize
objects with large amounts of occlusion and clutter.

Note that, the Queen's Dataset is more challenging than the UWA Dataset
since the former is more noisy and the points are not uniformly distributed.
That is the reason why the spin image based algorithm had a significant
drop in the recognition performance when tested on the two datasets.
Specifically, the average recognition rate of spin image based algorithm
on the UWA Dataset was 87.8\% while the best result on the Queen's
Dataset was only 54.4\%. Similarly, a notable decrease of performance
can also be found for the EM based algorithm, with 97.5\% recognition
rate for the UWA Dataset and 81.9\% recognition rate for the Queen's
Dataset. However, our RoPS based algorithm was consistently effective
and robust to different kinds of variations (including noise, varying
mesh resolution and occlusion), it outperformed the existing algorithms
and achieved comparable results in both datasets, obtaining a recognition
rate of 98.9\% on the UWA Dataset and 95.4\% on the Queen's Dataset.

{We also performed a timing experiment to measure
the average processing time to recognize each object in the scene.
The experiment was conducted on a computer with a 3.16 GHz Intel Core2
Duo CPU and a 4GB RAM. The code was implemented in MATLAB without
using any program optimization or parallel computing technique. The
average computational time to detect feature points and calculate
LRFs was 42.6s. The average computational time to generate RoPS descriptors
was 7.2s. Feature matching consumed 46.6s, while the computational
time for the transformation hypothesis generation was negligible.
Finally, verification and segmentation cost 57.4s in average. }

\subsection{{Recognition Results on The Ca' Foscari Venezia Dataset}}

{This dataset is composed of 20 models and 150 scenes.
Each scene contains 3 to 5 objects in the presence of occlusion and
clutter. Totally, there are 497 object instances in all scenes. This
dataset has been released just recently. It is the largest available
3D object recognition dataset. It is also more challenging than many
other datasets, containing several models with large flat and featureless
areas, and several models which are very similar in shape \citep{rodola2012scale}.}

{}
\begin{table*}
{\caption{Precision and recall values on the Ca' Foscari Venezia Dataset. The
best results are in bold fonts.  \label{tab:results-on-Foscari}}
}

{}%
\begin{tabular}{ccccccccccc}
\hline
 &  & {\footnotesize Armadillo} & {\footnotesize Bunny} & {\footnotesize Cat1} & {\footnotesize Centaur1} & {\footnotesize Chef} & {\footnotesize Chicken} & {\footnotesize Dog7} & {\footnotesize Dragon} & {\footnotesize Face}\tabularnewline
\hline
\multirow{2}{*}{{\footnotesize Precision}} & {\footnotesize RoPS} & {\footnotesize 97} & \textbf{{\footnotesize 100}} & \textbf{{\footnotesize 100}} & \textbf{{\footnotesize 100}} & \textbf{{\footnotesize 100}} & \textbf{{\footnotesize 97}} & \textbf{{\footnotesize 100}} & \textbf{{\footnotesize 100}} & \textbf{{\footnotesize 100}}\tabularnewline
 & {\footnotesize Game-theoretic} & \textbf{{\footnotesize 100}} & {\footnotesize 100} & {\footnotesize 78} & {\footnotesize 96} & {\footnotesize 93} & {\footnotesize 93} & {\footnotesize 95} & {\footnotesize 100} & {\footnotesize 91}\tabularnewline
\cline{2-11}
\multirow{2}{*}{{\footnotesize Recall}} & {\footnotesize RoPS} & \textbf{{\footnotesize 100}} & \textbf{{\footnotesize 100}} & {\footnotesize 44} & \textbf{{\footnotesize 100}} & \textbf{{\footnotesize 100}} & \textbf{{\footnotesize 100}} & \textbf{{\footnotesize 91}} & \textbf{{\footnotesize 100}} & \textbf{{\footnotesize 100}}\tabularnewline
 & {\footnotesize Game-theoretic} & {\footnotesize 97} & {\footnotesize 97} & \textbf{{\footnotesize 82}} & {\footnotesize 100} & {\footnotesize 100} & {\footnotesize 100} & {\footnotesize 86} & {\footnotesize 89} & {\footnotesize 95}\tabularnewline
\hline
\hline
 &  & {\footnotesize Ganesha} & {\footnotesize Gorilla0} & {\footnotesize Horse7} & {\footnotesize Lioness13} & {\footnotesize Para} & {\footnotesize Rhino} & {\footnotesize T-Rex} & {\footnotesize Victoria3} & {\footnotesize Wolf2}\tabularnewline
\hline
\multirow{2}{*}{{\footnotesize Precision}} & {\footnotesize RoPS} & \textbf{{\footnotesize 100}} & \textbf{{\footnotesize 100}} & \textbf{{\footnotesize 100}} & \textbf{{\footnotesize 100}} & \textbf{{\footnotesize 97}} & \textbf{{\footnotesize 96}} & \textbf{{\footnotesize 100}} & \textbf{{\footnotesize 100}} & \textbf{{\footnotesize 100}}\tabularnewline
 & {\footnotesize Game-theoretic} & {\footnotesize 89} & {\footnotesize 95} & {\footnotesize 97} & {\footnotesize 88} & {\footnotesize 97} & {\footnotesize 91} & {\footnotesize 97} & {\footnotesize 83} & {\footnotesize 82}\tabularnewline
\cline{2-11}
\multirow{2}{*}{{\footnotesize Recall}} & {\footnotesize RoPS} & \textbf{{\footnotesize 100}} & \textbf{{\footnotesize 100}} & \textbf{{\footnotesize 100}} & \textbf{{\footnotesize 100}} & \textbf{{\footnotesize 97}} & \textbf{{\footnotesize 100}} & \textbf{{\footnotesize 100}} & \textbf{{\footnotesize 95}} & \textbf{{\footnotesize 100}}\tabularnewline
 & {\footnotesize Game-theoretic} & {\footnotesize 100} & {\footnotesize 91} & {\footnotesize 100} & {\footnotesize 100} & {\footnotesize 94} & {\footnotesize 91} & {\footnotesize 97} & {\footnotesize 83} & {\footnotesize 95}\tabularnewline
\hline
\end{tabular}
\end{table*}

{The precision and recall values of RoPS based algorithm
on this dataset is shown in Table \ref{tab:results-on-Foscari}, the
results as reported in \citep{rodola2012scale} are also reported
for comparison. As in \citep{rodola2012scale}, two out of the 20
models were left out from the recognition tests and used as clutter.
The average number of detected feature points in a scene and a model
were 2210 and 5000, respectively. The RoPS based algorithm achieved
better precision results compared to \citep{rodola2012scale}. The
average precision of RoPS based algorithm was 99\%, which was higher
than \citep{rodola2012scale} by a margin of 6\%. Besides, the precision
values of 14 individual models were as high as 100\%. }

{The average recall of RoPS based algorithm was 96\%,
in contrast, the average recall of \citep{rodola2012scale} was 95\%.
Moreover, RoPS based algorithm achieved equal or better recall values
on 17 individual models out of the 18 models. Note that, SHOT descriptors
and a game-theoretic framework is used in \citep{rodola2012scale}
for 3D object recognition. It is observed that our RoPS based algorithm
performed better than SHOT based algorithm on this Dataset. }

In summary, the superior performance of our RoPS based 3D object recognition
algorithm is due to several reasons. First, the highly descriptiveness
and strong robustness of our RoPS feature descriptor improve the accuracy
of feature matching and therefore boost the performance of 3D object
recognition. Second, the unique, repeatable and robust LRF enables
the estimation of a rigid transformation from a single feature correspondence,
which therefore reduces the errors of transformation hypotheses. This
is because the probability of selecting only one correct feature correspondence
is much higher than the probability of selecting three correct feature
correspondences. Moreover, our proposed hierarchical object recognition
algorithm enables object recognition to be performed in an effective
and efficient manner.

\section{Conclusion \label{sec:Conclusion}}

In this paper, we proposed a novel RoPS feature descriptor for 3D
local surface description, and a new hierarchical RoPS based algorithm
for 3D object recognition. The RoPS feature descriptor is generated
by rotationally projecting the neighboring points around a feature
point onto three coordinate planes and calculating the statistics
of the distribution of the projected points. We also proposed a novel
LRF by calculating the scatter matrix of all points lying on the local
surface rather than just mesh vertices. The unique and highly repeatable
LRF facilitates the effectiveness and robustness of the RoPS descriptor.

We performed a set of experiments to assess our RoPS feature descriptor{{}
with respect to a set of different nuisances including noise, varying
mesh resolution and holes.} Comparative experimental results show
that our RoPS descriptor outperforms the state-of-the-art methods,
obtaining high descriptiveness and strong robustness to\textcolor{red}{{}
}{noise, varying mesh resolution and other deformations.}

Moreover, we performed extensive experiments for 3D object recognition
in complex scenes in the presence of noise, varying mesh resolution,
clutter and occlusion. Experimental results on the Bologna Dataset
show that our RoPS based algorithm is very effective and robust to
noise and mesh resolution variation. Experimental results on the UWA
Dataset show that RoPS based algorithm is very robust to occlusion
and outperforms existing algorithms. {The recognition
results achieved on the Queen's Dataset show that our algorithm outperforms
the state-of-the-art algorithms by a large margin. The RoPS based
algorithm was further tested on the largest available 3D object recognition
dataset (i.e., the Ca' Foscari Venezia Dataset), reporting superior
results.} Overall, our algorithm has achieved significant improvements
over the existing 3D object recognition algorithms when tested on
the same dataset.

{Interesting future research directions include the
extension of the proposed RoPS feature to encode both geometric and
photometric information. Integrating geometric and photometric cues
would be beneficial for the recognition of 3D objects with poor geometric
but rich photometric features (e.g., a flat or spherical surface).
Another direction is to adopt our RoPS descriptors to perform 3D shape
retrieval on a large scale 3D shape corpus, e.g., the SHREC Datasets
\citep{bronstein2010shrecRe}. }


\begin{acknowledgements}
The authors would like to acknowledge the following institutions.
Stanford University for providing the 3D models; Bologna University for providing the 3D scenes; INRIA
for providing the PHOTOMESH Dataset; Queen's University for
providing the 3D models and scenes; Universit\`a Ca\textquoteright{}
Foscari Venezia for providing the 3D models and scenes. The authors also
acknowledge A. Zaharescu from Aimetis Corporation for the results on the PHOTOMESH Dataset
shown in Tables \ref{tab:Robustness-of-RoPS-PhotoMesh} and \ref{tab:Robustness-of-MeshHOG-PhotoMesh-without-DoG}.

 \end{acknowledgements}
%
%


\bibliographystyle{apa}
\bibliography{3D_Object_IJCV} 

%
%

\end{document}